\documentclass{article} % For LaTeX2e
\usepackage{iclr2027_conference,times}

\usepackage{amsmath,amsfonts,bm}

\def\eqref#1{equation~\ref{#1}}
\def\1{\bm{1}}

\DeclareMathAlphabet{\mathsfit}{\encodingdefault}{\sfdefault}{m}{sl}
\SetMathAlphabet{\mathsfit}{bold}{\encodingdefault}{\sfdefault}{bx}{n}
\newcommand{\tens}[1]{\bm{\mathsfit{#1}}}

\def\tX{{\tens{X}}}

\DeclareMathOperator*{\argmin}{arg\,min}

\usepackage{hyperref}
\usepackage{url}

\usepackage{bm, bbm, amsmath, algorithm, algpseudocode, amssymb, amsthm} 
\usepackage[pdftex]{graphicx}
\usepackage{bigints}

\newtheorem{Pro}{Proposition}
\newcommand{\pkg}[1]{\texttt{#1}}

\usepackage{booktabs} 

\DeclareMathAlphabet{\mathsfit}{\encodingdefault}{\sfdefault}{m}{sl}
\SetMathAlphabet{\mathsfit}{bold}{\encodingdefault}{\sfdefault}{bx}{n}
\def\tX{{\tens{X}}}

\usepackage{wrapfig}

\usepackage{tcolorbox}
\tcbuselibrary{listings, breakable}

\title{Geometric Feature Learning for Functional Data Valued on the Symmetric Positive Definite Manifold}

\author{Samuel ~V.~ Singh \& Mimi Zhang  \\
School of Computer Science and Statistics\\
Trinity College Dublin\\
Dublin, Ireland \\
\texttt{\{ssingh9, mimi.zhang\}@tcd.ie}
}

\iclrfinalcopy % Uncomment for camera-ready version, but NOT for submission.
\begin{document}

\maketitle

\lhead{Preprint. Under review.}

\begin{abstract}
We here develop a functional neural network, termed \pkg{MatFAE}, for learning trajectories on the Riemannian manifold of symmetric positive definite (SPD) matrices.  \pkg{MatFAE} features intrinsic layers that map manifold-valued functions to Euclidean vector-valued functions, followed by a functional layer that projects them into a finite-dimensional Euclidean space. Unlike most neural networks for discrete-time sequences, \pkg{MatFAE} treats each sequence as a continuous function and can therefore encode trajectory dynamics (e.g., first-order derivatives) in its latent representations. Additionally, the morphology of the functional weights in the functional layer offers interpretability by revealing the regions of the input functional data that contribute most to the latent representations. We justify the design principles and properties of each intrinsic layer and detail how matrix factorization is handled during backpropagation. We apply \pkg{MatFAE} to a range of fMRI datasets, demonstrating its ability to efficiently learn informative representations from high-dimensional SPD trajectories and its practical value for real-world neuroimaging analysis.
\end{abstract}

\section{Introduction}
Functional data taking values in the Riemannian manifold of symmetric positive-definite (SPD) matrices are increasingly common in scientific and engineering applications. In neuroimaging, time-varying functional connectivity from fMRI is typically represented as trajectories of SPD covariance or correlation matrices  \citep{Lurie202030}. In DT-MRI, each voxel is associated with a \(3\times3\) SPD diffusion tensor that, when sampled along white matter tracts, yields SPD-valued functions over a one-dimensional spatial domain  \citep{Yuan2013102}. In subsurface hydrology and geomechanics, anisotropic hydraulic conductivity (or permeability) is commonly modelled by a \(3\times3\) SPD tensor describing directional flow of groundwater or heat; profiles measured along boreholes or evolving in time during injection, pumping, or thermal loading likewise form SPD-valued functional data \citep{https://doi.org/10.1029/2005RG000169}.  Other application fields include dynamic texture/action recognition, where each frame or temporal segment is encoded by an SPD descriptor \cite{6479703}, and structural health monitoring, where covariance matrices of vibration or strain signals from sensor networks are tracked over time.

A natural representation-learning strategy for SPD-matrix trajectories is to interpret each matrix as a graph and model each trajectory as a sequence of graphs, to which one can apply temporal attention mechanisms or recurrent architectures to capture dependencies over time \citep{NEURIPS2018_7070f908, NEURIPS2021_22785dd2}. Such sequence-based approaches, however, treat the functional data as discrete-time snapshots and may fail to capture structure in the underlying continuous-time dynamics. We hypothesise that first-order manifold behaviour, which captures how trajectories evolve in direction and speed, encodes additional discriminative and mechanistic information. A functional data perspective on SPD-valued trajectories provides a principled way to capture both the trajectories and their derivatives on the manifold, and to embed this richer geometric and dynamical information into latent representations for downstream tasks such as clustering. A detailed review of related work is given in Appendix~\ref{Related_work}.

Leveraging recent advances in Riemannian neural networks \citep{10.5555/3298483.3298534, 9122448, NEURIPS2019_6e69ebbf} and functional data analysis \citep{pmlr-v235-akeweje24a, singh2025NeurIPS}, we introduce \pkg{MatFAE}, a neural architecture for representation learning with functional data taking values on the manifold of SPD matrices. To the best of our knowledge, \pkg{MatFAE} is the first neural network explicitly designed for this setting. Its geometric layers are intrinsically Riemannian, respecting the non-Euclidean structure of SPD-valued trajectories and inducing a principled geometric bias that yields more efficient and structured representations.

The \pkg{MatFAE} architecture is modular. In Section \ref{Dimension_reduction}, we explain the dimension-reduction module, which maps high-dimensional SPD-valued functional data to lower-dimensional Euclidean functional data. In Section \ref{Functional_autoencoder}, we present the autoencoder module, which learns latent representations of the vector-valued Euclidean functional data and reconstructs the original SPD-valued functional data. Section \ref{Network_training} outlines the computational challenges in training the network, and Section \ref{Experiments} reports extensive numerical studies. All proofs are collected in the appendix.

\section{Mapping manifold-valued functions to Euclidean functions}\label{Dimension_reduction}
Let \(\mathcal{S}^m_+\) denote the manifold of \(m\times m\) SPD matrices: \(\mathcal{S}^m_+=\{\mathbf{X}\in\mathbb{R}^{m\times m}: \mathbf{X}=\mathbf{X}^{\top}, \mathbf{X}\succ0\}\). Given a matrix-valued trajectory \(\mathbf{X}(t)\in\mathcal{S}^m_+\) for \(t\in [t_0, t_1]\), we below develop a geometric network $\mathcal{F}$ that maps \(\mathbf{X}(t)\) to a \(p\)-dimensional Euclidean function $\pmb{y}(t)=\mathcal{F}(\mathbf{X}(t))\in \mathbb{R}^p$. We employ the Log-Euclidean (LE) metric \citep{doi:10.1137/050637996} and the Log-Cholesky (LC) metric \citep{doi:10.1137/18M1221084}, both having the bounded-determinant property: for a collection of SPD matrices, the determinant of the Fr\'{e}chet mean under the metric lies within the range of the input determinants. This property is crucial for the average-pooling layers in our network. Although the affine-invariant metric \citep{Pennec200641} also satisfies the bounded-determinant property, there is no general closed-form expression for the Fr\'{e}chet mean, making its use substantially more costly and complicating backpropagation. To avoid confusion, we write exp(\(\mathbf{U}\)) for the matrix exponential and log(\(\mathbf{U}\))  for the principal matrix logarithm. By contrast, \(\text{Exp}_{\mathbf{X}}(\mathbf{U})\) denotes the Riemannian exponential at \(\mathbf{X}\) and \(\text{Log}_{\mathbf{X}}(\mathbf{Z})\)  denotes the Riemannian logarithm at  \(\mathbf{X}\).

All trainable weights in $\mathcal{F}$ are time-invariant; thus, when no ambiguity arises, we expound the architecture at a fixed \(t\) and write  \(\mathbf{X}\) for \(\mathbf{X}(t)\). The manifold \(\mathcal{S}^m_+\) is an open cone in \(\text{Sym}(m)=\{\mathbf{U}\in\mathbb{R}^{m\times m}: \mathbf{U}=\mathbf{U}^{\top}\}\), and hence the tangent space \(\mathcal{T}_{\mathbf{X}}\mathcal{S}^m_+\)  at any point \(\mathbf{X}\in\mathcal{S}^m_+\) identifies with \(\text{Sym}(m)\): \(\mathcal{T}_{\mathbf{X}}\mathcal{S}^m_+=\text{Sym}(m)\), independently of \(\mathbf{X}\). Let  \(\mathfrak{L}(\mathbf{X})\) be the lower triangular matrix of the Cholesky decomposition of \(\mathbf{X}\): \(\mathbf{X}=\mathfrak{L}(\mathbf{X})\mathfrak{L}(\mathbf{X})^{\top}\), and \(\lfloor\mathfrak{L}(\mathbf{X})\rfloor\) the strictly lower triangular matrix of \(\mathfrak{L}(\mathbf{X})\). Define the log-Cholesky coordinate map \(\Phi: \mathcal{S}^m_+\rightarrow\mathbb{R}^{m(m+1)/2}\), \(\Phi(\mathbf{X})=(\text{vnz}(\mathfrak{L}(\mathbf{X})), \log(\text{vd}(\mathfrak{L}(\mathbf{X}))))^{\top}\), where \(\text{vnz}(\mathfrak{L}(\mathbf{X}))\) is the vector of the non-zero entries of \(\lfloor\mathfrak{L}(\mathbf{X})\rfloor\), and \(\text{vd}(\mathfrak{L}(\mathbf{X}))\) is the vector of the  diagonal entries of \(\mathfrak{L}(\mathbf{X})\). The LC metric makes \(\Phi\) a global Riemannian isometry: equipping \(\Phi(\mathbf{X})\) with the standard Euclidean inner product and pulling it back by \(\Phi\) defines the LC metric: 
\[\langle \mathbf{U}, \mathbf{V}\rangle _{\mathbf{X}}^{\text{LC}}=\langle (D_{\mathbf{X}}\Phi)(\mathbf{U}), ~(D_{\mathbf{X}}\Phi)(\mathbf{V})\rangle _2,~~~\mathbf{U}, \mathbf{V}\in\mathcal{T}_{\mathbf{X}}\mathcal{S}^m_+=\text{Sym}(m),\]
where \((D_{\mathbf{X}}\Phi)(\mathbf{U})=\frac{d}{ds}|_{s=0}\Phi(\mathbf{X}+s\mathbf{U})\) is the Fr\'{e}chet derivative of \(\Phi\) at the point \(\mathbf{X}\) to the tangent direction \(\mathbf{U}\).
Likewise, the LE metric is the pullback of the Frobenius metric on \(\text{Sym}(m)\) via the principal matrix logarithm \(\log: \mathcal{S}^m_+\rightarrow \text{Sym}(m)\); for \(\mathbf{X}\in\mathcal{S}^m_+\) and \(\mathbf{U}, \mathbf{V}\in\mathcal{T}_{\mathbf{X}}\mathcal{S}^m_+\), the LE inner product at \(\mathbf{X}\) is 
\[\langle \mathbf{U}, \mathbf{V}\rangle _{\mathbf{X}}^{\text{LE}}=\langle (D_{\mathbf{X}}\log)(\mathbf{U}), ~(D_{\mathbf{X}}\log)(\mathbf{V})\rangle _{F},\]
where \((D_{\mathbf{X}}\log)(\mathbf{U})=\frac{d}{ds}|_{s=0}\log(\mathbf{X}+s\mathbf{U})\) is the Fr\'{e}chet derivative of \(\log\) at the point \(\mathbf{X}\) to the tangent direction \(\mathbf{U}\).
Both Riemannian manifolds \((\mathcal{S}^m_+, \langle\cdot,\cdot\rangle ^{\text{LC}})\) and  \((\mathcal{S}^m_+, \langle\cdot,\cdot\rangle ^{\text{LE}})\) are flat (zero sectional curvature), geodesically complete, and simply connected; hence each is a Cartan-Hadamard manifold. Further properties of the LC and LE metrics are provided in Appendix \ref{Properties_of_the_two_metrics}.

According to Theorem 1 in \citet{singh2025NeurIPS}, on Cartan-Hadamard manifolds we can linearize the manifold-valued functions by mapping them into Euclidean space via the (globally defined) Riemannian logarithm map. The linearized functions can then be processed by the FAEclust functional network, which guarantees universal approximation. However, a direct mapping of \(\mathbf{X}(t)\) onto \(\text{Sym}(m)\) produces \(\pmb{y}(t)\) with dimension \(m(m+1)/2\), which imposes a heavy training cost for FAEclust. Therefore, we here develop a modular geometric network that compresses the trajectory \(\mathbf{X}(t)\) to a lower-order SPD trajectory \(\mathbf{X}_4(t)\in\mathcal{S}^{m_2}_+\) with \(m_2<m\), and then applies the Riemannian logarithm to map the trajectory \(\mathbf{X}_4(t)\) to the Euclidean space \(\text{Sym}(m_2)\) for subsequent processing by any functional neural network. This dimensionality reduction substantially decreases computational burden while preserving the relevant geometry. 

\begin{figure}[!htp]
    \centering
    \includegraphics[width=0.8\linewidth]{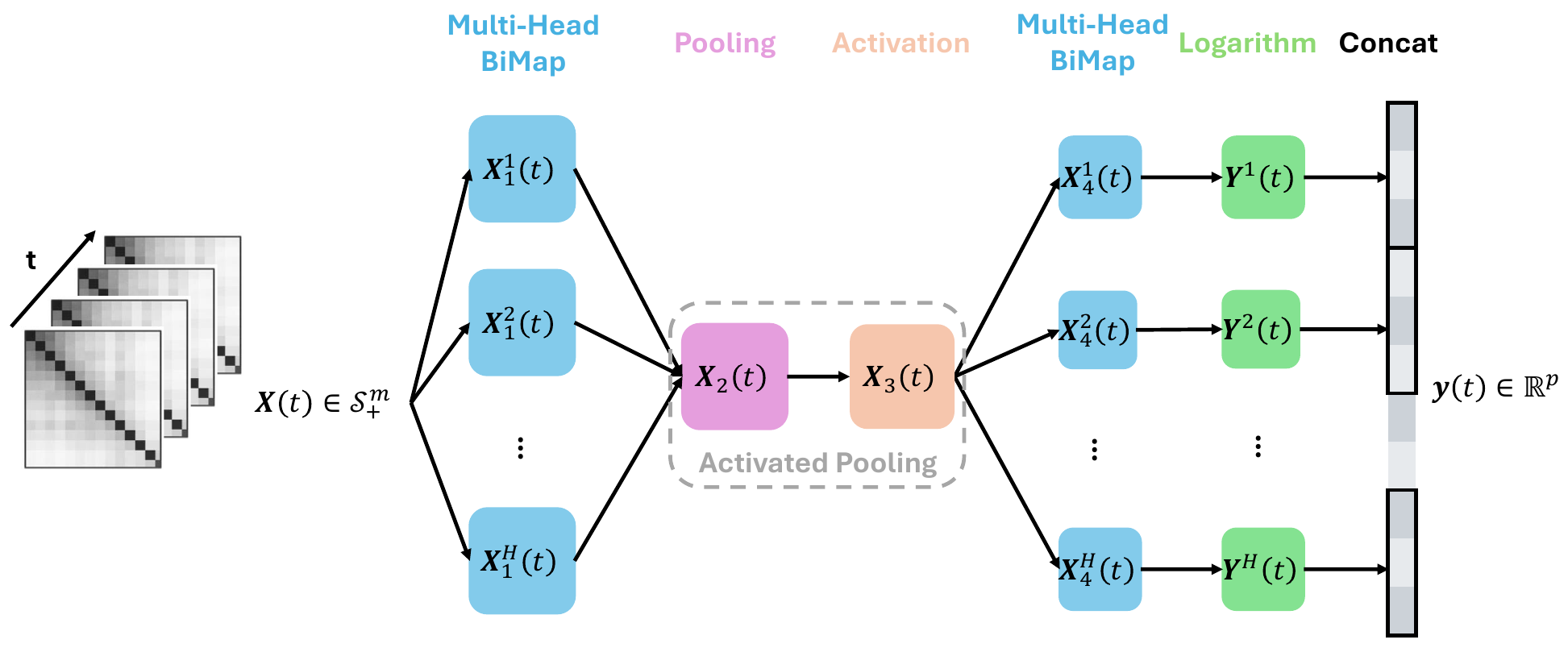}
    \caption{The two BiMap layers reduce the matrix dimension from \(m\) to \(m_1 (<m)\), and further from  \(m_1\) to  \(m_2 (<m_1)\); that is, \(\mathbf{X}_1^h(t)\in\mathcal{S}^{m_1}_+\) and \(\mathbf{X}_4^h(t)\in\mathcal{S}^{m_2}_+\) for \(1\leq h\leq H\). The Pooling, Activation and Logarithm layers are geometric layers that depend on the Riemannian metric on \(\mathcal{S}^m_+\). The Pooling layer performs manifold-average pooling of the SPD matrices \(\{\mathbf{X}_1^h(t): 1\leq h\leq H\}\); the Activation layer applies an activation on \(\mathbf{X}_2(t)\), and the Logarithm layer maps each \(\mathbf{X}_4^h(t)\) to the tangent space via the Riemannian logarithmic map \(\text{Log}^{\text{LC}}_{\mathbf{I}}(\cdot)\) or  \(\text{Log}^{\text{LE}}_{\mathbf{I}}(\cdot)\). Finally, we vectorize the lower-triangular entries of \(\mathbf{Y}^h(t)\) and concatenate them across heads to obtain a vector-valued function \(\pmb{y}(t)\).}
    \label{fig:fSPDnet}
\end{figure}

An overview of the network architecture is provided in \autoref{fig:fSPDnet}. The two BiMap layers perform dimensionality reduction, the Pooling layer aggregates SPD outputs across heads, and the Activation layer appropriately regularize the SPD matrix. The Logarithm layer performs Riemannian logarithm, mapping each head \(\mathbf{X}_4^h(t)\) to \(\mathbf{Y}^h(t)\): \(\mathbf{Y}^h(t)=\text{Log}^{\text{LC}}_{\mathbf{I}}(\mathbf{X}_4^h(t))\) or \(\mathbf{Y}^h(t)=\text{Log}^{\text{LE}}_{\mathbf{I}}(\mathbf{X}_4^h(t))\), where \(\mathbf{I}\) is the identity matrix. The explicit expressions for \(\text{Log}^{\text{LC}}_{\mathbf{I}}(\cdot)\) and \(\text{Log}^{\text{LE}}_{\mathbf{I}}(\cdot)\) are provided in Eqs. (\ref{LC_Log_Iden}) and (\ref{LE_Log}), respectively.  In the Concat layer, we vectorize the lower-triangular entries of \(\mathbf{Y}^h(t)\) and concatenate them across heads to obtain a vector-valued Euclidean function \(\pmb{y}(t)\). We now provide a rigorous justification for the BiMap, Pooling, and Activation layers and detail their properties.

\subsection{Multi-head BiMap}\label{Multi-head BiMap}
\citet{10.5555/3298483.3298534} introduced SPDNet, a deep architecture that operates directly on SPD matrices and preserves positive-definiteness throughout. Its core linear block (i.e., the BiMap layer) applies a congruence transformation:  \(\mathbf{X}_k=\mathbf{W}_k\mathbf{X}_{k-1}\mathbf{W}_k^{\top}\), where \(\mathbf{X}_{k-1}\in\mathcal{S}^{ d_{k-1}}_+\), and \(\mathbf{W}_k\in\mathbb{R}^{d_k\times d_{k-1}}~(d_k< d_{k-1})\) is a full row rank weight matrix with orthonormal rows. The bilinear congruence ensures that the output remains SPD, i.e., \(\mathbf{X}_k\in\mathcal{S}^{ d_k}_+\), while reducing dimension from \(d_{k-1}\) to \(d_k\). Stacking multiple BiMap layers yields progressively more compact and task-discriminative SPD representations.

For the \(\mathcal{S}^m_+\)-valued function \(\mathbf{X}(t)\), a natural extension of the bilinear mapping is  \(\mathbf{X}_1(t)=\mathbf{W}_1\mathbf{X}(t)\mathbf{W}_1^{\top}\). However, time-varying SPD signals almost never commute across time (i.e., \(\mathbf{X}(t_1)\mathbf{X}(t_2)\neq\mathbf{X}(t_2)\mathbf{X}(t_1)\)), and therefore no single global projection \(\mathbf{W}_1\) can align the covariance geometry uniformly over \(t\). When regimes switch and eigenvalues cross, causing the spectral subspaces to drift, any fixed \(\mathbf{W}_1\)  will inevitably overfit some intervals and underfit others. We therefore introduce a multi-head BiMap layer with \(H\) parallel bilinear maps: \(\mathbf{X}_1^h(t)=\mathbf{W}_1^h\mathbf{X}(t)(\mathbf{W}_1^h)^{\top}\), \(h=1, \ldots, H\), each with its own full row rank \(\mathbf{W}_1^h\). 

When \(\mathbf{X}(t)\) for different \(t\) values do not share a common eigenbasis, different heads approximate different (temporally) local joint structures, improving conditioning and reducing bias compared with a single global projection. Different heads naturally specialize to complementary, time-dependent relationships (e.g., distinct temporal regimes, frequency bands, or sensor groups). Their SPD outputs are then fused by an SPD-preserving pooling operator in the Pooling layer. This design also mirrors 1-D CNNs along the temporal axis: each \(\mathbf{W}_1^h\) serves as a shared temporal filter, and multiple heads enhance expressive power by capturing diverse temporal patterns while retaining weight sharing.

\subsection{Pooling}
The Pooling layer aggregates the SPD inputs \(\{\mathbf{X}_1^h: 1\leq h\leq H\}\). A natural choice is the Fr\'{e}chet mean on \(\mathcal{S}^{m_1}_+\) under either the LC or LE geometry:
\[
\mathbf{X}_2=\argmin_{\mathbf{X}\in\mathcal{S}^{m_1}_+}\sum_{h=1}^H d_g(\mathbf{X}, \mathbf{X}_1^h)^2,
\]
where \(d_g\) is the geodesic distance. Under either the LC metric or the LE metric, the Fr\'{e}chet mean has a closed form (Eqs. (\ref{LC_Frechet_mean}) and (\ref{LE_Frechet_mean})). Conceptually, this operation is the manifold analogue of global average pooling: it compresses \(H\) inputs into one SPD representative while strictly preserving positive-definiteness. A caveat is that the congruence mapping  \(\mathbf{W}_1^h\mathbf{X}(\mathbf{W}_1^h)^{\top}\) does not commute with the matrix logarithm/exponential in the LE metric: in general,
\[
\log(\mathbf{W}_1^h\mathbf{X}(\mathbf{W}_1^h)^{\top})\neq \mathbf{W}_1^h\log(\mathbf{X})(\mathbf{W}_1^h)^{\top} ~\text{ and }~ \exp(\mathbf{W}_1^h\mathbf{X}(\mathbf{W}_1^h)^{\top})\neq \mathbf{W}_1^h\exp(\mathbf{X})(\mathbf{W}_1^h)^{\top}.
\]
Likewise, for the LC metric, the Cholesky factor \(\mathfrak{L}(\mathbf{W}_1^h\mathbf{X}(\mathbf{W}_1^h)^{\top})\) depends nonlinearly on both \(\mathbf{W}_1^h\) and \(\mathbf{X}\). Consequently, the Fr\'{e}chet mean computed in the Pooling layer is neither invariant nor equivariant to dimension-reducing congruence actions applied to \(\mathbf{X}(t)\). We detail this lack of invariance/equivariance below.

A canonical family of symmetries of \(\mathcal{S}^m_+\) is given by the congruence action of GL($m$): any invertible matrix \(\mathbf{U}\in\text{GL}(m)\) defines the symmetry \(\mathbf{X}\mapsto\mathbf{U}\mathbf{X}\mathbf{U}^{\top}\), acting on all points \(\mathbf{X}\in\mathcal{S}^m_+\) \citep{10.5555/3540261.3541664}. (1) When \(\mathbf{U}\) is an element of \(\mathcal{S}^m_+\), the symmetry \(\mathbf{X}\mapsto\mathbf{U}\mathbf{X}\mathbf{U}^{\top}\) is a generalization of the Euclidean \textit{translation}, fixing no points of \(\mathcal{S}^m_+\). (2) When \(\mathbf{U}\in\text{O}(m)\) is an orthogonal matrix, the symmetry \(\mathbf{X}\mapsto\mathbf{U}\mathbf{X}\mathbf{U}^{\top}\) is conjugation by \(\mathbf{U}\), and hence fixes the base point \(\mathbf{I}=\mathbf{U}\mathbf{I}\mathbf{U}^{\top}=\mathbf{U}\mathbf{U}^{\top}=\mathbf{I}\). The stabilizer of \(\mathbf{I}\) is precisely \(\text{O}(m)\): elements with \(\text{det}(\mathbf{U})=1\) are \(\mathcal{S}^m_+\)-\textit{rotations}, and elements with \(\text{det}(\mathbf{U})=-1\) are \(\mathcal{S}^m_+\)-\textit{reflections}. 
\begin{Pro}\label{not-invariant-equvariant}
    The LE Fr\'{e}chet mean \(\mathbf{X}_2\) is invariant or equivariant under orthogonal congruence only under pathological conditions.  The LC Fr\'{e}chet mean is neither invariant nor equivariant under orthogonal congruence. Moreover, under the LC and LE geometries, the Fr\'{e}chet mean \(\mathbf{X}_2\) is generally neither invariant nor equivariant under actions of the isometry group on  \(\mathcal{S}^m_+\).
\end{Pro}

\paragraph{Max pooling on \(\mathcal{S}^{m_1}_+\)}  Unlike Euclidean space, the manifold \(\mathcal{S}^{m_1}_+\) admits neither canonical global coordinates nor a natural total order. Therefore, a ``max'' operation is not intrinsically defined: any elementwise maximum defined in a particular chart (e.g., \(\log\) or \(\Phi\)) is chart-dependent and generally fails to be equivariant under manifold isometries. An intrinsic alternative is to define a scalar score \(s: \mathcal{S}^{m_1}_+\rightarrow\mathbb{R}\) and select the head \(\mathbf{X}_1^h\) with the highest score \(s(\mathbf{X}_1^h)\). However, because features pass through the congruence map  \(\mathbf{X}\mapsto\mathbf{W}_1^h\mathbf{X}(\mathbf{W}_1^h)^{\top}\), no non-trivial choice of \(s\) yields invariance/equivariance to general matrix congruence; this follows from the same obstruction discussed in Appendix \ref{Properties_BiMap_layer}. Equivariance is attainable only in pathological cases (e.g., orthogonal conjugation with spectral scores, or specific metric-based selections with matched transformations of any reference). Moreover, the \(\max\) operation is non-smooth and therefore less amenable to optimization. For these reasons, \pkg{MatFAE} adopts average (Fr\'{e}chet) pooling.

Although the BiMap’s rectangular congruence deprives the Pooling layer of invariance or equivariance guarantees, it remains a highly efficient dimension-reduction operator: it preserves SPD by construction and needs only two matrix-matrix products, completely avoiding eigendecomposition. 

%All three Fr\'{e}chet means (LE, LC, AI) are not invariant to simultaneous orthogonal transformations of the weight matrices \(\{\mathbf{W}_1^h: h=1, \ldots, H\}\). If we left-multiply every \(\mathbf{W}_1^h\) by the same orthogonal matrix \(\mathbf{Q}\): \(\mathbf{W}_1^h\mapsto\mathbf{Q}\mathbf{W}_1^h\), then each input to the Pooling layer transforms by congruence: \(\mathbf{X}_1^h=\mathbf{Q}\mathbf{X}_1^h\mathbf{Q}^{\top}\). Under this change, LE and AI Fr\'{e}chet means are equivariant (not invariant): \(\mathbf{X}_2\mapsto\mathbf{Q}\mathbf{X}_2\mathbf{Q}^{\top}\). For the LC Fr\'{e}chet mean, there is no simple closed-form relation; in general, the new mean differs from both \(\mathbf{X}_2\) and \(\mathbf{Q}\mathbf{X}_2\mathbf{Q}^{\top}\).  Consequently, orthogonal reparameterizations of \(\{\mathbf{W}_1^h: h=1, \ldots, H\}\) do not leave the pooled output unchanged. Thus back-propagation proceeds without ambiguity in \(\{\mathbf{W}_1^h: h=1, \ldots, H\}\).

\subsection{Activation}\label{Activation}
Conventional deep architectures interleave pointwise nonlinearities \(a: \mathbb{R}\rightarrow\mathbb{R}\) between linear layers (e.g., ReLU). Such activations are non-linear and, in most cases, non-expansive with respect to the Euclidean norm (i.e., \(\|a(\pmb{x})-a(\pmb{y})\|_2\leq L\|\pmb{x}-\pmb{y}\|_2\) with \(L\leq1\)); certain choices (e.g., the logistic sigmoid) are strictly contractive. Non-expansiveness helps control the network’s overall Lipschitz constant, improving numerical stability, robustness to perturbations, and generalization by biasing toward smoother mappings. The non-linearities between layers prevent the deep neural networks from collapsing to a single fully connected layer. Below we prove that the Fr\'{e}chet-mean pooling operator is, in general, neither contractive nor non-expansive (Proposition \ref{mapping_Lipschitz_LE} and Proposition \ref{mapping_non-expansive_LC}). However, the Fr\'{e}chet-mean pooling operator itself is non-linear  (Proposition \ref{without_activation}). Therefore, a repeated block ``BiMap \(\rightarrow\) Pooling \(\rightarrow\) BiMap \(\rightarrow\) Pooling''  will not collapse to a single  ``BiMap \(\rightarrow\) Pooling'' block. 

\begin{Pro}\label{mapping_Lipschitz_LE}
    Fix any spectral band \(0<\alpha\leq \beta<\infty\) and consider the subset \(\mathcal{X}_{\alpha, \beta}=\{\mathbf{X}\in\mathcal{S}^m_+: \alpha\mathbf{I}\preceq \mathbf{X}\preceq \beta\mathbf{I} \}\). Under the LE metric, the mapping \(F: \mathcal{S}^m_+\mapsto\mathcal{S}^{m_1}_+\),
    \[
    F(\mathbf{X}; \{\mathbf{W}_1^h\}_{h=1}^H)=\mathbb{E}_{\text{LE}}(\mathbf{W}_1^1\mathbf{X}(\mathbf{W}_1^1)^{\top}, \ldots, \mathbf{W}_1^H\mathbf{X}(\mathbf{W}_1^H)^{\top})=\exp(\frac{1}{H}\sum_{h=1}^H \log(\mathbf{W}_1^h\mathbf{X}(\mathbf{W}_1^h)^{\top}),
    \]
    is Lipschitz on \(\mathcal{X}_{\alpha, \beta}\) with constant \(L\leq \beta/\alpha\). In particular, \(F\) is neither contractive nor non-expansive.
\end{Pro}
\begin{Pro}\label{mapping_non-expansive_LC}
Fix any spectral band \(0<\alpha\leq \beta<\infty\) and consider the subset \(\mathcal{X}_{\alpha, \beta}=\{\mathbf{X}\in\mathcal{S}^m_+: \alpha\mathbf{I}\preceq \mathbf{X}\preceq \beta\mathbf{I} \}\). Under the LC metric, the mapping \(F: \mathcal{S}^m_+\mapsto\mathcal{S}^{m_1}_+\),
\[
F(\mathbf{X}; \{\mathbf{W}_1^h\}_{h=1}^H)=\mathbb{E}_{\text{LC}}(\mathbf{W}_1^1\mathbf{X}(\mathbf{W}_1^1)^{\top}, \ldots, \mathbf{W}_1^H\mathbf{X}(\mathbf{W}_1^H)^{\top})=\Phi^{-1}(\frac{1}{H}\sum_{h=1}^H \Phi(\mathbf{W}_1^h\mathbf{X}(\mathbf{W}_1^h)^{\top}),
\]
is Lipschitz with constant \(L\leq \frac{2\beta^{3/2}}{\alpha}\max\{1, \alpha^{-1}\}\). In particular, \(F\) is neither contractive nor non-expansive.
\end{Pro}

\begin{Pro}\label{without_activation}
    The pooling map is itself nonlinear. Without any non-linear activation after the Pooling layer, the repeated block ``BiMap \(\rightarrow\) Pooling \(\rightarrow\) BiMap \(\rightarrow\) Pooling''  will not collapse to a single  ``BiMap \(\rightarrow\) Pooling'' block.
\end{Pro}

Proposition \ref{without_activation} implies that, if contractivity is not required, the Activation layer can be omitted. The architecture in Figure \ref{fig:fSPDnet} then reduces to ``BiMap \(\rightarrow\) Pooling \(\rightarrow\) BiMap \(\rightarrow\) Logarithm \(\rightarrow\) Concat''. However, the Fr\'{e}chet-mean pooling is only Lipschitz, neither contractive nor non-expensive (Propositions \ref{mapping_Lipschitz_LE} and \ref{mapping_non-expansive_LC}). Therefore, when stability or spectral control is desired, inserting a contractive activation after pooling will enforce a uniform Lipschitz bound, curb eigenvalue inflation, and improve numerical stability. 

The ReEig layer introduced in \citet{10.5555/3298483.3298534} performs a spectral rectification by eigendecomposing \(\mathbf{X}_2=\mathbf{Q}\mathbf{\Lambda}\mathbf{Q}^{\top}\), replacing each eigenvalue \(\lambda_i\) with \(\max\{\lambda_i, \epsilon\}\), and reconstructing \(\mathbf{X}_3=\mathbf{Q}\max\{\mathbf{\Lambda}, \epsilon\mathbf{I}\}\mathbf{Q}^{\top}\). However, the ReEig activation operator is not scale-invariant (positively homogeneous); for \(c>0\), we will have \(c\mathbf{X}_3\neq\mathbf{Q}\max\{c\mathbf{\Lambda}, \epsilon\mathbf{I}\}\mathbf{Q}^{\top}\), whenever a scaled eigenvalue crosses the threshold. We can modify the ReEig operation to be scale-invariant:
\begin{equation*}\label{LE_spectral_rectification}
   \mathbf{X}_3= s(\mathbf{X}_2)\mathbf{Q}\exp(a(\log(\mathbf{\Lambda})-\log(s(\mathbf{X}_2))\mathbf{I}))\mathbf{Q}^{\top},
\end{equation*}
where \(a(\cdot)\) is elementwise activation, and \(s(\mathbf{X}_2)\) is any 1-homogeneous scale functional, e.g., \(s(\mathbf{X}_2)=\frac{1}{m_1}\text{trace}(\mathbf{X}_2)\). The map is positively homogeneous: \(\log(c\mathbf{X}_2)=\mathbf{Q}[\log(\mathbf{\Lambda})+\log(c)\mathbf{I}]\mathbf{Q}^{\top}\) and \(s(c\mathbf{X}_2)=c\times s(\mathbf{X}_2)\). However, embedding \(s(\mathbf{X}_2)\) inside the spectral nonlinearity couples the scale and eigenvalue paths in backpropagation, requiring extra spectral Fr\'{e}chet operations and raising the computational burden (\(\approx\mathcal{O}(m_1^3)\)). Another type of activation, introduced by \citet{8451626}, directly applies the scalar function \(\exp(\cdot)\), \(\sinh(\cdot)\) or \(\cosh(\cdot)\) elementwise to \(\mathbf{X}_2\). However, \(\exp(\cdot)\) and \(\cosh(\cdot)\) grow rapidly; their derivatives  can trigger exploding activations/gradients and inflate condition numbers. Moreover, elementwise maps provide no direct spectral control and can mix scales, obscuring links to the underlying process. %\citet{9122448} developed a matrix-learning architecture and proved that non-linear activation is not needed for their network.

Motivated by the need to control the Lipschitz bound and by the fact that our pooling operator is already non-linear, we develop an intrinsic geodesic-shrinkage activation on \(\mathcal{S}^{m_1}_+\), which preserves scale behavior while yielding tractable derivatives.

\paragraph{Geodesic shrinkage} Let \(\phi\) denote either the principal matrix logarithm \(\log\) or the log-Cholesky coordinate map \(\Phi\). The Pooling layer computes the  Fr\'{e}chet mean in \(\phi\)-coordinates via \[F(\mathbf{X}; \{\mathbf{W}_1^h\}_{h=1}^H)=\phi^{-1}(\frac{1}{H}\sum_{h=1}^H \phi(\mathbf{W}_1^h\mathbf{X}(\mathbf{W}_1^h)^{\top})).\] 
We want to regularize the Fr\'{e}chet mean \(\mathbf{X}_2=F(\mathbf{X}; \{\mathbf{W}_1^h\}_{h=1}^H)\) by moving \(\mathbf{X}_2\) back toward the identity matrix \(\mathbf{I}\) along the unique geodesic joining \(\mathbf{I}\) to \(\mathbf{X}_2\), stopping at a proportion \(0<\alpha< 1\) of the total geodesic length. Because \(\phi\) maps  \(\mathbf{I}\) to the origin of the coordinate space: \(\phi(\mathbf{I})=\mathbf{0}\), geodesic distances from \(\mathbf{I}\) reduce to Euclidean norms: \(d_{\text{LE}}(\mathbf{I}, \mathbf{X}_2)=\|\log(\mathbf{X}_2)\|_F\) under the LE metric, and \(d_{\text{LC}}(\mathbf{I}, \mathbf{X}_2)=\|\Phi(\mathbf{X}_2)\|_2\) under the LC metric. Define the activation output  \(\mathbf{X}_3\) as the point at distance \(\alpha d_g(\mathbf{I}, \mathbf{X}_2)\) from \(\mathbf{I}\) along that geodesic. Since geodesics are straight lines in \(\phi\)-coordinates, this construction has a particularly simple coordinate description:
\[
\phi(\mathbf{X}_3)=(1-\alpha)\phi(\mathbf{I})+\alpha\phi(\mathbf{X}_2)=\alpha\phi(\mathbf{X}_2),
\]
or, equivalently, \(\mathbf{X}_3=\phi^{-1}(\alpha\phi(\mathbf{X}_2)\)). In words: we shrink the coordinate vector toward zero (the image of \(\mathbf{I}\)) by a factor \(\alpha\), and then map back to the manifold. This yields \(d_g(\mathbf{I}, \mathbf{X}_3)=\alpha d_g(\mathbf{I}, \mathbf{X}_2)\). Hence the operation contracts distances from the identity exactly by \(\alpha\). The effect is directly analogous to weight decay in Euclidean networks, but carried out in intrinsic coordinates of the manifold.

To keep overhead negligible, rather than ``pool, then activate,'' we combine the Pooling layer and the Activation layer into one layer, namely, the ``Activated Pooling'' layer in Figure \ref{fig:fSPDnet}. The activated-pooling computes the average in the \(\phi\)-chart,  shrinks the average by \(\alpha\), and maps the shrinked average back to the manifold:
\[\mathbf{X}_3=\phi^{-1}(\frac{\alpha}{H}\sum_{h=1}^H \phi(\mathbf{W}_1^h\mathbf{X}(\mathbf{W}_1^h)^{\top})).\] 
This eliminates the intermediate \(\mathbf{X}_2\). The only extra computation relative to plain pooling is a scalar multiplication by \(\alpha\) in the \(\phi\)-domain.

\subsection{Concatenation}
Substituting the intermediate layers, the composite mapping from the input \(\mathbf{X}(t)\) to \(\mathbf{Y}^h(t)\) is:
\begin{equation}
\mathbf{Y}^h(t)=\text{Log}^g_{\mathbf{I}}\left(\mathbf{W}_4^h\phi^{-1}(\frac{\alpha}{H}\sum_{k=1}^H \phi\left(\mathbf{W}_1^k\mathbf{X}(t)(\mathbf{W}_1^k)^{\top})\right)(\mathbf{W}_4^h)^{\top}\right).    
\end{equation}
For each head \(h=1, \ldots, H\) and time \(t\), the Concatenation layer maps a symmetric matrix  \(\mathbf{Y}^h(t)\in\text{Sym}(m_2)\) to a vector \(\text{vl}(\mathbf{Y}^h(t))\in\mathbb{R}^{m_2(m_2+1)/2}\), and then concatenates the vectors across heads. We  equip \(\text{Sym}(m_2)\) with the Frobenius inner product and \(\mathbb{R}^{m_2(m_2+1)/2}\) with the Euclidean inner product and require $\text{vl}(\cdot)$ to be an isometry: \(\langle\mathbf{Y}_1, \mathbf{Y}_2\rangle_F=\langle\text{vl}(\mathbf{Y}_1), \text{vl}(\mathbf{Y}_2)\rangle _2\), for all \(\mathbf{Y}_1, \mathbf{Y}_2\). Isometry guarantees norm preservation and, crucially for backpropagation, it makes the adjoint equal the inverse. Fix any indexing \(\kappa: \{(i, j): 1\leq j\leq i\leq m_2\}\mapsto \{1, \ldots, m_2(m_2+1)/2\}\) (e.g., lexicographic). Define
\[
[\text{vl}(\mathbf{Y})]_{\kappa(i,i)}=Y_{i,i}, ~~~[\text{vl}(\mathbf{Y})]_{\kappa(i,j)}=\sqrt{2}Y_{i,j}, \text{ for } i>j.
\]
It is easy to prove that $\text{vl}(\cdot)$ is an isometry, and the adjoint equals the inverse. Finally, the output of the network is \(\pmb{y}(t)^{\top}=[\text{vl}(\mathbf{Y}^1(t))^{\top}, \ldots, \text{vl}(\mathbf{Y}^H(t))^{\top}]\).

\section{Functional autoencoder with geometry-aware layers}\label{Functional_autoencoder}
\begin{figure}[!htp]
  \centering
  \includegraphics[width=0.65\textwidth]{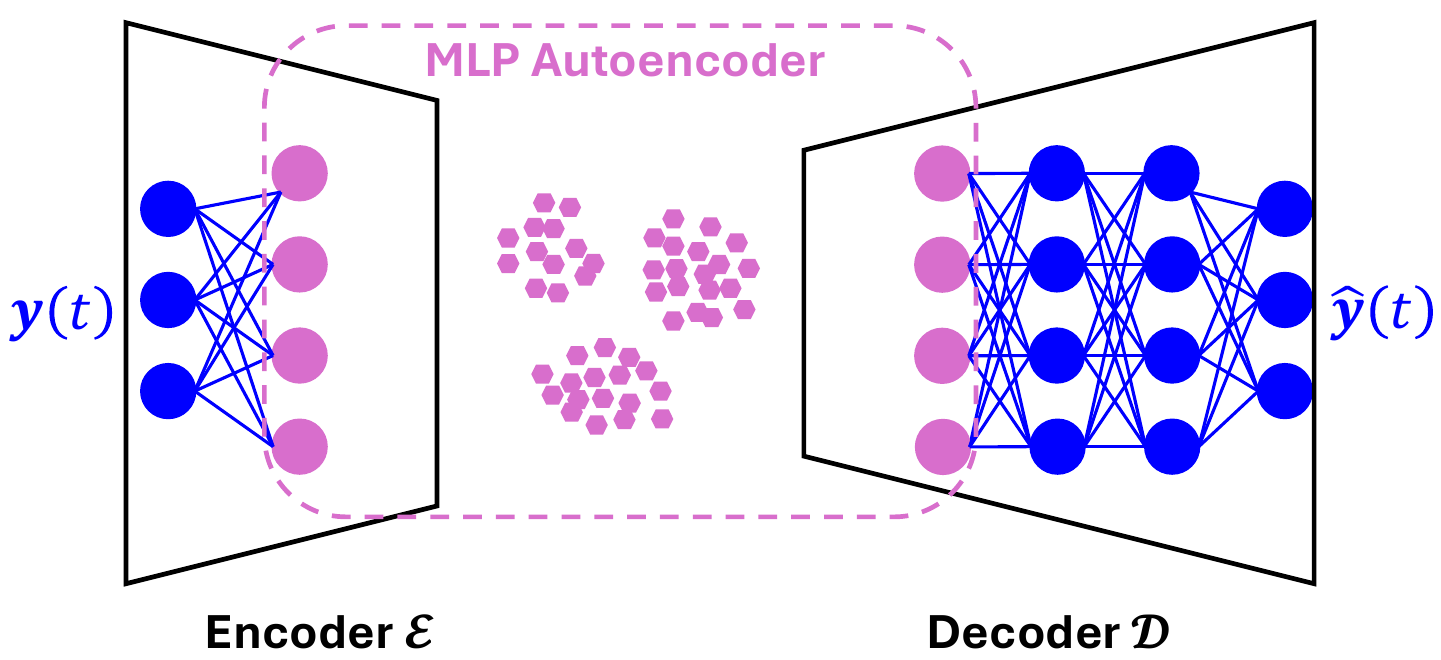}
  \vspace{-10pt}
  \caption{The FAEclust architecture generalizes the multilayer perceptron (MLP) autoencoder by appending functional layers at the encoder input and the decoder output. In the above figure, the plum block denotes the MLP autoencoder, and the blue components denote functional nodes/edges.}
  \label{fig:FAEclust}
\end{figure}

    The matrix-to-vector network in Figure \ref{fig:fSPDnet} is modular: it can be paired with any network that accept vector-valued functions.  We here adapt the FAEclust architecture \citep{singh2025NeurIPS} to learn a latent embedding of the Euclidean vector-valued function \(\pmb{y}(t)\) and, via a geometry-aware decoder, to reconstruct the original SPD trajectory  \(\mathbf{X}(t)\). FAEclust comprises an encoder $\mathcal{E}$ and a decoder $\mathcal{D}$ (see Figure \ref{fig:FAEclust}). Let \( \mathcal{H}([t_0, t_1]; \mathbb{R}^p)\) denote the separable Hilbert space of square-integrable functions that are defined on \([t_0, t_1]\) and taking values in \(\mathbb{R}^p\). Given a \(p\)-variate function $\pmb{y}\in \mathcal{H}([t_0, t_1], \mathbb{R}^p)$, the encoder maps the entire trajectory to a latent multivariate data point $\pmb{x}=\mathcal{E}(\pmb{y})\in\mathbb{R}^s$; the decoder then reconstructs a functional trajectory $\hat{\pmb{y}}=\mathcal{D}(\pmb{x})\in \mathcal{H}([t_0, t_1], \mathbb{R}^p)$. Cluster analysis is performed on the embedded data in the latent space. Unlike a conventional MLP autoencoder, FAEclust introduces four functional layers: one placed at the encoder entrance and three at the decoder exit, which act on functions rather than vectors. The feedforward equation for the functional layer in the encoder is given by 
\begin{equation}
   \pmb{x}^{(1)}=a(\int_{t_0}^{t_1}\pmb{W}^{(1)}(t) \pmb{y}(t)dt+\pmb{b}), \label{functional_layer_encoder}
\end{equation}
where \(\pmb{W}^{(1)}(t)\) is a functional weight matrix, $\pmb{b}$ is a bias vector, and $a$ is an activation function. \(\pmb{x}^{(1)}\) is then fed into the MLP autoencoder. The equations for the three functional layers in the decoder are:
\begin{align*}
    \hat{\pmb{y}}^{(1)}(t)&=a(\pmb{\mathcal{W}}^{(1)}(t) \hat{\pmb{x}}+\pmb{b}_1(t)),\\
\hat{\pmb{y}}^{(2)}(t)&=a(\pmb{\mathcal{W}}^{(2)}(t) \hat{\pmb{y}}^{(1)}(t)+\pmb{b}_2(t)),\\
\hat{\pmb{y}}(t)&=\pmb{\mathcal{W}}^{(3)}(t) \hat{\pmb{y}}^{(2)}(t),
\end{align*}
where \(\hat{\pmb{x}}\) is the output of the MLP autoencoder, \(\{\pmb{\mathcal{W}}^{(l)}(t), l=1, 2, 3\}\) are functional weight matrices, and  \(\{\pmb{b}_1(t), \pmb{b}_2(t)\}\) are functional biases.  For the output layer, the activation function is linear, and there is no functional bias. 

For a matrix \(\mathbf{U}\in\text{Sym}(m)\), let \(\mathcal{V}(\mathbf{U})=(\text{vec}(\lfloor\mathbf{U}\rfloor), \text{vec}(\text{diag}(\mathbf{U})))\in\mathbb{R}^{m(m+1)/2}\) denote the vector of the lower triangular entries of \(\mathbf{U}\). Note that \(\mathcal{V}\) is different from the log-Cholesky coordinate map \(\Phi\). The inverse \(\mathcal{V}^{-1}\) maps a vector \(\pmb{u}\in\mathbb{R}^{m(m+1)/2}\) back to an \(m\times m\) matrix in \(\text{Sym}(m)\): \(\mathbf{U}=\mathcal{V}^{-1}(\mathcal{V}(\mathbf{U}))\). We can directly apply FAEclust and train the network to output \(\hat{\pmb{y}}(t)\in\mathbb{R}^{m(m+1)/2}\) that approximates \(\mathcal{V}(\text{Log}^{\text{LC}}_{\mathbf{I}}(\mathbf{X}(t)))\) or \(\mathcal{V}(\text{Log}^{\text{LE}}_{\mathbf{I}}(\mathbf{X}(t)))\). In practice, this is computationally prohibitive: each functional weight matrix  \(\pmb{\mathcal{W}}^{(l)}(t)\) will have size about \(m^2\times m^2\), yielding \(\mathcal{O}(m^4)\) parameters per layer and a heavy training burden. We cut computation by decoding in a low-dimensional tangent space, then lifting via conjugation with thin, column-orthonormal matrices, giving an injective isometry for the Frobenius inner product.  Let \(\hat{\pmb{y}}^{(1)}(t)\in\mathbb{R}^{p_1(p_1+1)/2}\), where \(p_1<m\), and \(\mathcal{V}^{-1}(\hat{\pmb{y}}^{(1)}(t))\in\text{Sym}(p_1)\) is a trajectory in the tangent space of \(\mathcal{S}^{p_1}_+\).   Let \(\mathbf{M}_1\in\mathbb{R}^{p_2\times p_1}\) and \(\mathbf{M}_2\in\mathbb{R}^{m\times p_2}\) be two thin matrices of orthonormal columns. The modified functional layers are
\begin{eqnarray}
    \hat{\pmb{y}}^{(1)}(t)&=&a(\pmb{\mathcal{W}}^{(1)}(t) \hat{\pmb{x}}+\pmb{b}_1(t)),\nonumber\\
\hat{\pmb{y}}^{(2)}(t)&=&a(\pmb{\mathcal{W}}^{(2)}(t) {\color{gray} \mathcal{V}(\mathbf{M}_1 \mathcal{V}^{-1}(\hat{\pmb{y}}^{(1)}(t))\mathbf{M}_1^{\top})}+\pmb{b}_2(t)),\label{expandanding1}\\
\hat{\mathbf{X}}(t)&=&\text{Exp}^{g}_{\mathbf{I}}({\color{gray}\mathbf{M}_2 \mathcal{V}^{-1}(\hat{\pmb{y}}^{(2)}(t))\mathbf{M}_2^{\top}}),\label{expandanding2}
\end{eqnarray}
where \(\hat{\mathbf{X}}(t)\in\mathcal{S}^m_+\) is the reconstructed SPD trajectory, and \(\text{Exp}^{g}_{\mathbf{I}}\) is the Riemannian exponential at \(\mathbf{I}\) under the chosen metric \(g\) (i.e., LC or LE). In Eq. \eqref{expandanding1}, we lift from \(\text{Sym}(p_1)\) to \(\text{Sym}(p_2)\) by conjugation before applying the nonlinearity in vectorized form, and in Eq. \eqref{expandanding2} we lift again to \(\text{Sym}(m)\), and map back to the manifold.  In line with the output layer of FAEclust, Eq. \eqref{expandanding2} only involves a bilinear mapping (in the tangent space).

\section{Network training}\label{Network_training}
For a Riemannian metric \(g\) on \(\mathcal{S}^m_+\) (i.e., the LC or LE metric), define \(\mathcal{H}_g([t_0, t_1]; \mathcal{S}^m_+)=\{\tX: [t_0, t_1]\rightarrow \mathcal{S}^m_+ \text{ measurable } |\int_{t_0}^{t_1} d_g(\tX(t), \mathbf{I})^2 dt< \infty \}\). The matrix-to-vector module is the mapping \(\mathcal{F}\): \(\tX\in \mathcal{H}_g([t_0, t_1]; \mathcal{S}^m_+)\mapsto \pmb{y}=\mathcal{F}(\tX)\in \mathcal{H}([t_0, t_1]; \mathbb{R}^p)\), and the \pkg{MatFAE} network is the mapping \(\mathcal{D}\circ\mathcal{E}\circ\mathcal{F}\): \(\tX\in \mathcal{H}_g([t_0, t_1]; \mathcal{S}^m_+)\mapsto \hat{\tX}=\mathcal{D}(\mathcal{E}(\mathcal{F}(\tX)))\in \mathcal{H}_g([t_0, t_1]; \mathcal{S}^m_+)\). Given a set of \(\mathcal{S}^m_+\)-valued trajectories \(\{\tX_i\in\mathcal{H}_g([t_0, t_1]; \mathcal{S}^m_+)\}_{i=1}^n\), \pkg{MatFAE} produces reconstructions \(\{\hat{\tX}_i\in\mathcal{H}_g([t_0, t_1]; \mathcal{S}^m_+)\}_{i=1}^n\). Under the LE metric, the reconstruction loss is 
\[\mathcal{L}_r=\sum_{i=1}^n\int_{t_0}^{t_1}d_{\text{LE}}(\tX_i(t), \hat{\tX}_i(t))^2dt=\sum_{i=1}^n\int_{t_0}^{t_1}\|\log(\tX_i(t))- \mathbf{M}_2 \mathcal{V}^{-1}(\hat{\pmb{y}}^{(2)}_i(t))\mathbf{M}_2^{\top}\|_F^2dt.
\] Under the LC metric, the reconstruction loss \(\mathcal{L}_r\) is 
\begin{align*}
    \sum_{i=1}^n\int_{t_0}^{t_1}d_{\text{LC}}(\tX_i(t), \hat{\tX}_i(t))^2dt&=\sum_{i=1}^n\int_{t_0}^{t_1}\|\lfloor\mathfrak{L}(\tX_i(t))\rfloor- \lfloor\mathbf{M}_2 \mathcal{V}^{-1}(\hat{\pmb{y}}_i^{(2)}(t))\mathbf{M}_2^{\top}\rfloor\|_F^2dt\\
    &+\sum_{i=1}^n\int_{t_0}^{t_1}\|\log(\text{diag}(\mathfrak{L}(\tX_i(t))))- \frac{1}{2}\text{diag}(\mathbf{M}_2 \mathcal{V}^{-1}(\hat{\pmb{y}}_i^{(2)}(t))\mathbf{M}_2^{\top})\|_F^2dt.
\end{align*}

The training objective function of \pkg{MatFAE} comprises three components: (1) \(\mathcal{L}_r\), the reconstruction loss; (2) \(\mathcal{L}_w\), a regularization term on the functional weights and functional biases \(\{\pmb{W}^{(1)}(t)\), \(\pmb{\mathcal{W}}^{(1)}(t), \pmb{\mathcal{W}}^{(2)}(t), \pmb{\mathcal{W}}^{(3)}(t), \pmb{b}_1(t), \pmb{b}_2(t)\}\); and (3) \(\mathcal{L}_{\mathrm{div}}\), a row-space decorrelation penalty on the congruence weight matrices \(\{\mathbf{W}_1^h\}_{h=1}^H\) and \(\{\mathbf{W}_4^h\}_{h=1}^H\), The detailed formulation of \(\mathcal{L}_w\) is given in \citep{singh2025NeurIPS}, while the formulation of \(\mathcal{L}_{\mathrm{div}}\) is provided in Appendix \ref{Congruence matrix regularization}.

\paragraph{Riemannian gradient} \pkg{MatFAE} involves a few matrix parameters that are constrained to Stiefel manifolds. We here briefly outline the gradient-descent procedure for optimizing them. Let \(\text{St}(m, p)=\{\mathbf{M}\in\mathbb{R}^{m\times p}: \mathbf{M}^{\top}\mathbf{M}=\mathbf{I}\}\) denote the Stiefel manifold of \(m\times p\) matrices with orthonormal columns (i.e., $m>p$). For any smooth mapping \(f: \text{St}(m, p)\rightarrow\mathbb{R}\), let \(\mathbf{D}=\nabla_{\mathbf{M}}f(\mathbf{M})\) denote the normal Euclidean gradient. Then project \(\mathbf{D}\) onto the tangent space to get the Riemannian gradient: \(\Delta=\mathbf{D}-\frac{1}{2}\mathbf{M}(\mathbf{M}^{\top}\mathbf{D}+\mathbf{D}^{\top}\mathbf{M})\). For a given step size \(\eta>0\), compute a thin QR factorization of \(\mathbf{M}+\eta(-\Delta)\):  \(\mathbf{M}-\eta\Delta=\mathbf{Q}\mathbf{R}\); then the updated \(\mathbf{M}\) is \(\mathbf{Q}\text{diag}(\text{sign}(\text{diag}(\mathbf{R})))\), where \(\text{sign}(\cdot)\) is the sign function, and \(\text{sign}(0)=1\).

\paragraph{Matrix factorization in backpropagation} The matrix-to-vector mapping \(\mathcal{F}\) involves eigen-decomposition or Cholesky factorization (e.g.,  \(\mathbf{X}_3(t)=\phi^{-1}(\frac{\alpha}{H}\sum_{h=1}^H \phi(\mathbf{W}_1^h\mathbf{X}(t)(\mathbf{W}_1^h)^{\top}))\) for geodesic-shrinkage activated pooling). Computing the partial derivatives required for backpropagation is therefore quite demanding. Appendix \ref{matrix_partial_derivative} provides the detailed derivations, as well as the functional gradients required for backpropagation from the encoder \(\mathcal{E}\) to the mapping \(\mathcal{F}\).

\section{Experiments}\label{Experiments}
\subsection{Cluster analysis}
\pkg{MatFAE} is available as a fully documented Python package. We here compare \pkg{MatFAE} with TSRVF \citep{8786184}, GeoAtt \citep{9761822}, SPDNet \citep{10.5555/3298483.3298534}, PGA \citep{1318725}, MLP autoencoder (DC), and k-means applied directly to the pairwise distance matrix (PDM) induced by the LE metric.  Appendix \ref{Supplementary_algorithm} briefly explains each benchmark method and reports the configuration settings used for all algorithms. For methods that produce vector representations, clustering is performed using k-means. The number of clusters is selected by maximizing the silhouette score over \(k\in\{2, 3, 4, 5\}\). All simulated and real datasets are available in the \pkg{MatFAE} GitHub repository.

\paragraph{Simulation study} We consider ten simulation scenarios, and the data-generating mechanism for each is provided in detail in Appendix \ref{Supplementary_synthetic}. For each scenario, we generate 100 independent datasets and apply all benchmark methods to each dataset. Clustering performance is then summarized by the mean and standard deviation of the AMI and ARI across the 100 replications, reported in Tables \ref{data_synthetic_AMI} and \ref{data_synthetic_ARI}, respectively. The benchmark is structured as an ablation ladder, with each synthetic dataset designed to isolate a single type of structural signal. Accordingly, failure on a given rung indicates that the method lacks the feature-extraction mechanism needed to capture that particular structure.

Table~\ref{data_synthetic_AMI} shows that MatFAE performs best overall across the ten simulation scenarios. It achieves the highest mean AMI in six scenarios (A, C, D, E, F, and I) and ties for the best in H and J. Its main advantage appears in the more difficult settings, especially dwell time (C), transition frequency (D), and smoothness (E), where competing methods show only modest or near-chance performance. The gain in E is particularly pronounced: MatFAE reaches 0.725, whereas the next-best method, PGA, achieves only 0.163. Scenario I is the most challenging for all methods, with all AMI values below 0.16. Even in this case, however, MatFAE achieves the highest mean score, suggesting limited but detectable sensitivity to the multi-scale structure.  %By contrast, TSRVF and DC perform particularly well in the temporally driven scenarios A and H, where they achieve perfect or near-perfect recovery. PGA performs best in B, G and J, indicating that it is especially effective when the cluster structure is dominated by static or low-dimensional geometric differences. GeoAtt also performs exceptionally well in F, although MatFAE still attains the highest score in that scenario.

\begin{table}[!ht]
\centering
\caption{AMI scores for the ten simulation scenarios.  The table reports the mean (top row) and standard deviation (bottom row) of the scores over 100 repetitions.}
\label{data_synthetic_AMI}
\begin{tabular}{l|ccccccc}
\toprule
Scenario (\textit{rung}) & TSRVF& GeoAtt & SPDNet & PGA & DC & PDM & MatFAE \\
\midrule
A (\textit{temporal ordering}) &  0.998&  -0.001&  -0.000&  -0.000&  0.989&  -0.000& \textbf{1.000} \\
  &  0.016&  0.008&  0.010&  0.008&  0.100&  0.009&  0.000 \\
B (\textit{static identity}) &  0.009&  0.021&  0.033&  \textbf{0.998}&  0.006&  0.949& 0.971 \\
  &  0.025&  0.046&  0.055&  0.012&  0.022&  0.135&  0.013 \\
C (\textit{dwell time}) &  0.319&  0.447&  0.438&  0.451&  0.444&  0.449& \textbf{0.593} \\
  &  0.064&  0.076&  0.059&  0.063&  0.077&  0.062&  0.040 \\
D (\textit{transition frequency}) &  0.350&  -0.001&  0.000&  -0.000&  0.336&  0.002& \textbf{0.368} \\
  &  0.069&  0.008&  0.010&  0.010&  0.126&  0.015&  0.044 \\
E (\textit{smoothness}) &  0.123&  0.115&  0.009&  0.163&  0.062&  -0.000& \textbf{0.725} \\
  &  0.077&  0.073&  0.024&  0.064&  0.044&  0.000&  0.054 \\
F (\textit{Wishart concentration}) &  0.125&  0.999&  0.469&  0.753&  0.182&  0.648& \textbf{1.000} \\
  &  0.055&  0.007&  0.067&  0.042&  0.074&  0.100&  0.000 \\
G (\textit{real HRF}) &  -0.001&  0.000&  0.003& \textbf{1.000}&  0.979&  0.151&  0.929 \\
  &  0.010&  0.011&  0.015&  0.000&  0.119&  0.105&  0.025 \\
H (\textit{trajectory direction}) & \textbf{1.000}&  0.149&  0.029&  0.010& \textbf{1.000}&  0.002& \textbf{1.000} \\
  &  0.000&  0.075&  0.032&  0.022&  0.000&  0.017&  0.000 \\
I (\textit{multi-scale dynamics}) &  0.141&  0.032&  0.033&  0.031&  0.048&  0.032& \textbf{0.155} \\
  &  0.090&  0.044&  0.043&  0.042&  0.059&  0.042&  0.085 \\
J (\textit{Neurosynth}) &  0.000&  0.235&  0.714&  \textbf{1.000}&  0.766&  \textbf{1.000}&  \textbf{1.000}\\
  &  0.010&  0.242&  0.134&  0.000&  0.352&  0.000&  0.000\\
\bottomrule
\end{tabular}
\end{table}

\paragraph{Real datasets} We used six real resting-state fMRI datasets from two public neuroimaging repositories: OpenNeuro and the International Neuroimaging Data-sharing Initiative. Detailed dataset profiles, preprocessing procedures, and additional analyses are provided in Appendix \ref{Supplementary_real}. Using the COBRE dataset, Appendix \ref{COBRE_interpretability} provides a step-by-step demonstration of MatMAE's intrinsic interpretability. Specifically, it shows how the temporal profiles of the functional weights \(\pmb{W}^{(1)}(t)\) in the encoder's functional layer identify the portions of the SPD trajectory that contribute most strongly to the latent representations.

\begin{table}[!ht]
\centering
\caption{AMI scores for the six real resting-state fMRI datasets.}
\label{data_real_AMI}
\begin{tabular}{l|ccccccc}
\toprule
Dataset & TSRVF & GeoAtt & SPDNet & PGA & DC & PDM & MatFAE \\
\midrule
CNP & 0.051 & 0.015 & 0.013 & 0.035 & 0.015 & 0.001 & \textbf{0.109} \\
COBRE & 0.064 & 0.052 & 0.052 & 0.041 & 0.013 & 0.041 & \textbf{0.137} \\
ADHD-200 & 0.026 & 0.003 & 0.004 & 0.027 & 0.053 & 0.055 & \textbf{0.078} \\
ABIDE-I & 0.010 & 0.001 & 0.001 & 0.000 & 0.004 & 0.001 & \textbf{0.030} \\
TCP & 0.026 & 0.002 & 0.010 & 0.020 & 0.004 & 0.018 & \textbf{0.042} \\
CAT-D & 0.019 & 0.022 & 0.009 & 0.041 & 0.003 & 0.028 & \textbf{0.062} \\
\bottomrule
\end{tabular}
\end{table}
Clustering performance is reported using AMI and ARI in Tables~\ref{data_real_AMI} and~\ref{data_real_ARI}, respectively. \pkg{MatFAE}  achieves the best performance across all six cohorts, demonstrating its consistent advantage over the competing methods. At the same time, the absolute AMI values reflect the intrinsic difficulty of unsupervised diagnostic discovery from resting-state fMRI. ADHD-200 and ABIDE-I are heterogeneous multi-site cohorts, CNP contains small patient subgroups across four DSM categories, and resting-state cognition is itself an unconstrained latent state. Even supervised models on these datasets typically improve only moderately over chance, with reported balanced accuracies of approximately 0.60-0.70 for ABIDE-I \citep{Heinsfeld2018ABIDE,Saponaro2022harmonization}, 0.72-0.81 for COBRE \citep{Zeng2018ebiomed}, and 0.55-0.66 for ADHD-200 \citep{Qiu2024ASTNet}. These results are about $0.10$-$0.25$ above chance, indicating the intrinsic difficulty of prediction from resting-state fMRI. 

\subsection{Classification analysis} 
We further evaluate the encoder blocks of MatFAE in a supervised setting, denoted MatFAE-C, by removing the decoder and appending a classification head to the latent representation. We compare MatFAE-C against three recent classifiers for temporal SPD/graph-structured data on the same six real datasets: SPD-SRU \citep{NEURIPS2018_7070f908}, which extends recurrent neural networks to operate directly on sequences of SPD matrices; STAGIN \citep{NEURIPS2021_22785dd2}, which learns dynamic brain-connectome representations via spatio-temporal graph attention; and BW-Norm \citep{Wang_2025_CVPR}, a Riemannian batch-normalization method under the Bures-Wasserstein metric that processes SPD trajectories frame by frame. Full configuration details for all methods are reported in Appendix~\ref{classfication_method_config}. Classification results are reported in Table \ref{data_real_Classification}. The folds are not site-stratified and no ComBat harmonisation is applied; therefore, the reported results should be interpreted as within-distribution classification performance rather than evidence of out-of-site generalisation. 

\begin{table}[!ht]
\centering
\caption{Balanced classification accuracy (mean $\pm$ std over five-fold CV) for MatFAE-C and three classification baselines. Chance-level accuracy is shown in parentheses next to each dataset name.}
\label{data_real_Classification}
\begin{tabular}{l|cccc}
\toprule
Dataset & SPD-SRU & STAGIN & BW-norm & MatFAE-C\\
\midrule
CNP (0.25)  &0.245 $\pm$ 0.032 &0.287 $\pm$ 0.028 & 0.338 $\pm$ 0.032 & 0.453 $\pm$ 0.048\\
COBRE (0.5) &0.541 $\pm$ 0.036 &0.638 $\pm$ 0.036 & 0.604 $\pm$ 0.033 & 0.741 $\pm$ 0.047\\
ADHD-200 (0.5)  &0.519 $\pm$ 0.069 &0.528 $\pm$ 0.042 & 0.571 $\pm$ 0.041 & 0.630$\pm$ 0.072\\
ABIDE-I (0.5)   &0.495 $\pm$ 0.024 &0.587 $\pm$ 0.030 & 0.525 $\pm$ 0.034& 0.628 $\pm$ 0.063\\
TCP (0.5)   &0.505 $\pm$ 0030	&0.498 $\pm$ 0.034 & 0.498 $\pm$ 0.029 & 0.609 $\pm$ 0.067\\
CAT-D (0.5) &0.521 $\pm$ 0.040	&0.542 $\pm$ 0.033 & 0.493 $\pm$ 0.050 & 0.642 $\pm$ 0.056\\
\bottomrule
\end{tabular}
\end{table}

The supervised variant \pkg{MatFAE-C} achieves the highest balanced classification accuracy on all six cohorts. Although the CNP accuracy appears lower in absolute terms, CNP is a four-class problem, for which chance-level balanced accuracy is 0.25 rather than 0.50; the CNP result therefore remains substantially above chance, and in fact represents the largest relative margin over the best baseline of any cohort (0.453 vs.\ 0.338 for BW-norm, a $34\%$ relative improvement). Statistical significance analyses of the reported improvements are provided in Appendix \ref{Statistical significance tests and ablation study}.  Overall, the performance of \pkg{MatFAE-C} is comparable to conservative resting-state fMRI classification results reported on multisite psychiatric datasets such as ABIDE-I and ADHD-200, where accuracies around 0.60-0.70 are commonly observed under realistic validation settings. However, it is below some highly optimized supervised studies on single binary datasets, such as COBRE schizophrenia classification, where substantially higher accuracies have been reported using task-specific feature engineering, feature selection, multimodal inputs, or deep supervised architectures. These comparisons should therefore be interpreted cautiously, as published results differ substantially in diagnostic task, number of classes, preprocessing, site handling, validation protocol, and whether external generalisation is evaluated. The purpose of \pkg{MatFAE-C} is not to claim state-of-the-art supervised diagnostic classification, but to demonstrate that the representation learned by the \pkg{MatFAE} encoder contains diagnostically relevant information. In neuroscience, many neurological and psychiatric conditions are heterogeneous and may contain clinically meaningful subtypes that are not known in advance. \pkg{MatFAE} can therefore be used to identify latent subgroups and support symptom-driven, transdiagnostic characterizations of psychiatric conditions.

\section{Conclusion}
We introduced \pkg{MatFAE}, a geometry-aware neural architecture for representation learning from functional data valued on the SPD manifold. Beyond combining SPD-aware dimensionality reduction with functional feature extraction, \pkg{MatFAE} contributes three theoretically grounded mechanisms: a regularized multi-head BiMap layer that captures heterogeneous, time-varying joint structure inaccessible to a single global projection; a geodesic-shrinkage activation, derived from our analysis of the non-contractive, nonlinear properties of Fr\'{e}chet-mean pooling, which generalizes to Riemannian manifolds beyond the SPD setting; and an intrinsic interpretability procedure that traces latent representations back to the original SPD-valued trajectories. Together, these components allow \pkg{MatFAE} to capture both the geometric structure and the temporal morphology of SPD-valued functional data within a single, lightweight, non-recurrent architecture.

Across ten simulation scenarios constructed as an ablation ladder, \pkg{MatFAE} achieved the best or tied-best performance on eight of ten rungs, with especially large gains on rungs requiring sensitivity to trajectory smoothness, dwell time, and transition frequency. On six real resting-state fMRI cohorts, \pkg{MatFAE} achieved the best AMI and ARI on all six, and its supervised variant, MatFAE-C, achieved balanced classification accuracies of 0.61-0.74, comparable to or exceeding published results for diagnosis prediction on these cohorts. The consistency of \pkg{MatFAE}'s advantage across both idealized synthetic settings and heterogeneous real data, despite the latter's inherently lower achievable signal, suggests that the architecture captures generalizable structure in SPD-valued dynamics rather than overfitting to a particular data-generating process.

\section*{Acknowledgments}
This work was conducted with the financial support of the Research Ireland Centre for Research Training in Digitally-Enhanced Reality (d-real) under Grant No. 18/CRT/6224.

%%%%%%%%%%%%%%%%%%%%%%%%%%%%%%%%%%%%%%%%%%%%%%%%%%%%%%%%%%%%

\bibliography{Ref_nLRL}

@inproceedings{singh2025NeurIPS,
    title={Shape-Informed Clustering of Multi-Dimensional Functional Data via Deep Functional Autoencoders}, 
    author={Samuel Singh and Shirley Coyle and Mimi Zhang},
    year={2025},
 booktitle = {Advances in Neural Information Processing Systems},
 publisher = {Curran Associates, Inc.},
}

@inproceedings{10.5555/3298483.3298534,
author = {Huang, Zhiwu and Gool, Luc Van},
title = {A {Riemannian} network for {SPD} matrix learning},
year = {2017},
publisher = {AAAI Press},
booktitle = {Proceedings of the Thirty-First AAAI Conference on Artificial Intelligence},
pages = {2036–2042},
numpages = {7},
}

@article{doi:10.1073/pnas.1705120114,
author = {Diego Vidaurre  and Stephen M. Smith  and Mark W. Woolrich },
title = {Brain network dynamics are hierarchically organized in time},
journal = {Proceedings of the National Academy of Sciences},
volume = {114},
number = {48},
pages = {12827-12832},
year = {2017},
doi = {10.1073/pnas.1705120114},
}

@ARTICLE{Fine199841,
	author = {Fine, Shai and Singer, Yoram and Tishby, Naftali},
	title = {The hierarchical hidden Markov model: Analysis and applications},
	year = {1998},
	journal = {Machine Learning},
	volume = {32},
	number = {1},
	pages = {41 – 62},
	doi = {10.1023/A:1007469218079},
}

@article{doi:10.1137/18M1221084,
author = {Lin, Zhenhua},
title = {Riemannian Geometry of Symmetric Positive Definite Matrices via {Cholesky} Decomposition},
journal = {SIAM Journal on Matrix Analysis and Applications},
volume = {40},
number = {4},
pages = {1353-1370},
year = {2019},
doi = {10.1137/18M1221084},
}

@article{doi:10.1137/050637996,
author = { Arsigny, Vincent and  Fillard, Pierre and  Pennec, Xavier and  Ayache, Nicholas},
title = {Geometric Means in a Novel Vector Space Structure on Symmetric Positive‐Definite Matrices},
journal = {SIAM Journal on Matrix Analysis and Applications},
volume = {29},
number = {1},
pages = {328-347},
year = {2007},
doi = {10.1137/050637996},
}

@article{10.1093/cercor/bhs352,
    author = {Allen, Elena A. and Damaraju, Eswar and Plis, Sergey M. and Erhardt, Erik B. and Eichele, Tom and Calhoun, Vince D.},
    title = {Tracking Whole-Brain Connectivity Dynamics in the Resting State},
    journal = {Cerebral Cortex},
    volume = {24},
    number = {3},
    pages = {663-676},
    year = {2014},
    month = {03},
    doi = {10.1093/cercor/bhs352},
}

@article{GLOVER1999416,
title = {Deconvolution of Impulse Response in Event-Related BOLD fMRI1},
journal = {NeuroImage},
volume = {9},
number = {4},
pages = {416-429},
year = {1999},
doi = {https://doi.org/10.1006/nimg.1998.0419},
author = {Gary H. Glover},
}

@ARTICLE{Pennec200641,
	author = {Pennec, Xavier and Fillard, Pierre and Ayache, Nicholas},
	title = {A {Riemannian} framework for tensor computing},
	year = {2006},
	journal = {International Journal of Computer Vision},
	volume = {66},
	number = {1},
	pages = {41 – 66},
	doi = {10.1007/s11263-005-3222-z},
}

@inproceedings{10.5555/3540261.3541664,
author = {L\'{o}pez, Federico and Pozzetti, Beatrice and Trettel, Steve and Strube, Michael and Wienhard, Anna},
title = {Vector-valued distance and gyrocalculus on the space of symmetric positive definite matrices},
year = {2021},
publisher = {Curran Associates Inc.},
booktitle = {Proceedings of the 35th International Conference on Neural Information Processing Systems},
articleno = {1403},
numpages = {17},
series = {NIPS '21}
}

@article{Fan1957ImbeddingCF,
  title={Imbedding Conditions for {Hermitian} and Normal Matrices},
  author={Ky Fan and Gordon Pall},
  journal={Canadian Journal of Mathematics},
  year={1957},
  volume={9},
  pages={298 - 304},
  doi={10.4153/CJM-1957-036-1}
}

@INPROCEEDINGS{8451626,
  author={Zhang, Tong and Zheng, Wenming and Cui, Zhen and Li, Chaolong},
  booktitle={2018 25th IEEE International Conference on Image Processing (ICIP)}, 
  title={Deep Manifold-to-Manifold Transforming Network}, 
  year={2018},
  volume={},
  number={},
  pages={4098-4102},
  doi={10.1109/ICIP.2018.8451626}
}

@ARTICLE{9122448,
  author={Chakraborty, Rudrasis and Bouza, Jose and Manton, Jonathan H. and Vemuri, Baba C.},
  journal={IEEE Transactions on Pattern Analysis and Machine Intelligence}, 
  title={{ManifoldNet}: A Deep Neural Network for Manifold-Valued Data With Applications}, 
  year={2022},
  volume={44},
  number={2},
  pages={799-810},
  doi={10.1109/TPAMI.2020.3003846}}

@ARTICLE{8786184,
  author={Dai, Mengyu and Zhang, Zhengwu and Srivastava, Anuj},
  journal={IEEE Transactions on Medical Imaging}, 
  title={Analyzing Dynamical Brain Functional Connectivity as Trajectories on Space of Covariance Matrices}, 
  year={2020},
  volume={39},
  number={3},
  pages={611-620},
  doi={10.1109/TMI.2019.2931708}}

@ARTICLE{9761822,
  author={Dan, Tingting and Huang, Zhuobin and Cai, Hongmin and Laurienti, Paul J. and Wu, Guorong},
  journal={IEEE Transactions on Medical Imaging}, 
  title={Learning Brain Dynamics of Evolving Manifold Functional MRI Data Using Geometric-Attention Neural Network}, 
  year={2022},
  volume={41},
  number={10},
  pages={2752-2763},
  doi={10.1109/TMI.2022.3169640}}

@ARTICLE{1318725,
  author={Fletcher, P.T. and Conglin Lu and Pizer, S.M. and Sarang Joshi},
  journal={IEEE Transactions on Medical Imaging}, 
  title={Principal geodesic analysis for the study of nonlinear statistics of shape}, 
  year={2004},
  volume={23},
  number={8},
  pages={995-1005},
  doi={10.1109/TMI.2004.831793}}

@inproceedings{NEURIPS2021_22785dd2,
 author = {Kim, Byung-Hoon and Ye, Jong Chul and Kim, Jae-Jin},
 booktitle = {Advances in Neural Information Processing Systems},
 editor = {M. Ranzato and A. Beygelzimer and Y. Dauphin and P.S. Liang and J. Wortman Vaughan},
 pages = {4314--4327},
 publisher = {Curran Associates, Inc.},
 title = {Learning Dynamic Graph Representation of Brain Connectome with Spatio-Temporal Attention},
volume = {34},
 year = {2021}
}

@ARTICLE{Yuan2013102,
	author = {Yuan, Ying and Zhu, Hongtu and Styner, Martin and Gilmore, John H. and Marron, J.S.},
	title = {Varying coefficient model for modeling diffusion tensors along white matter tracts},
	year = {2013},
	journal = {Annals of Applied Statistics},
	volume = {7},
	number = {1},
	pages = {102 – 125},
	doi = {10.1214/12-AOAS574},
}

@ARTICLE{Lurie202030,
	author = {Lurie, Daniel J. and Kessler, Daniel and Bassett, Danielle S. and Betzel, Richard F. and Breakspear, Michael and Keilholz, Shella and Kucyi, Aaron and Liégeois, Raphaël and Lindquist, Martin A. and McIntosh, Anthony Randal and Poldrack, Russell A. and Shine, James M. and Thompson, William Hedley and Bielczyk, Natalia Z. and Douw, Linda and Kraft, Dominik and Miller, Robyn L. and Muthuraman, Muthuraman and Pasquini, Lorenzo and Razi, Adeel and Vidaurre, Diego and Xie, Hua and Calhoun, Vince D.},
	title = {Questions and controversies in the study of time-varying functional connectivity in resting {fMRI}},
	year = {2020},
	journal = {Network Neuroscience},
	volume = {4},
	number = {1},
	pages = {30 – 69},
	doi = {10.1162/netn_a_00116},
}

@article{https://doi.org/10.1029/2005RG000169,
author = {Sanchez-Vila, Xavier and Guadagnini, Alberto and Carrera, Jesus},
title = {Representative hydraulic conductivities in saturated groundwater flow},
journal = {Reviews of Geophysics},
volume = {44},
number = {3},
pages = {},
doi = {10.1029/2005RG000169},
year = {2006}
}

@inproceedings{NEURIPS2018_7070f908,
 author = {Chakraborty, Rudrasis and Yang, Chun-Hao and Zhen, Xingjian and Banerjee, Monami and Archer, Derek and Vaillancourt, David and Singh, Vikas and Vemuri, Baba},
 booktitle = {Advances in Neural Information Processing Systems},
 pages = {},
 publisher = {Curran Associates, Inc.},
 title = {A Statistical Recurrent Model on the Manifold of Symmetric Positive Definite Matrices},
 volume = {31},
 year = {2018}
}

@inproceedings{NEURIPS2019_6e69ebbf,
 author = {Brooks, Daniel and Schwander, Olivier and Barbaresco, Frederic and Schneider, Jean-Yves and Cord, Matthieu},
 booktitle = {Advances in Neural Information Processing Systems},
 pages = {},
 publisher = {Curran Associates, Inc.},
 title = {Riemannian batch normalization for {SPD} neural networks},
 volume = {32},
 year = {2019}
}

@InProceedings{pmlr-v235-akeweje24a,
  title = 	 {Learning Mixtures of {G}aussian Processes through Random Projection},
  author =       {Akeweje, Emmanuel and Zhang, Mimi},
  booktitle = 	 {Proceedings of the 41st International Conference on Machine Learning},
  pages = 	 {720--739},
  year = 	 {2024},
  volume = 	 {235},
  series = 	 {Proceedings of Machine Learning Research},
  month = 	 {21--27 Jul},
  publisher =    {PMLR},
}

@ARTICLE{Hall200770,
	author = {Hall, Peter and Horowitz, Joel L.},
	title = {Methodology and convergence rates for functional linear regression},
	year = {2007},
	journal = {Annals of Statistics},
	volume = {35},
	number = {1},
	pages = {70 – 91},
	doi = {10.1214/009053606000000957},
}

@ARTICLE{Dette2024304,
	author = {Dette, Holger and Tang, Jiajun},
	title = {Statistical inference for function-on-function linear regression},
	year = {2024},
	journal = {Bernoulli},
	volume = {30},
	number = {1},
	pages = {304 – 331},
	doi = {10.3150/23-BEJ1598},
}

@ARTICLE{Balasubramanian2025973,
	author = {Balasubramanian, Krishnakumar and Müller, Hans-Georg and Sriperumbudur, Bharath K.},
	title = {Functional linear and single-index models: A unified approach via Gaussian Stein identity},
	year = {2025},
	journal = {Bernoulli},
	volume = {31},
	number = {2},
	pages = {973 – 1006},
	doi = {10.3150/24-BEJ1755},
}

@ARTICLE{Yeon20232599,
	author = {Yeon, Hyemin and Dai, Xiongtao and Nordman, Daniel J.},
	title = {Bootstrap inference in functional linear regression models with scalar response},
	year = {2023},
	journal = {Bernoulli},
	volume = {29},
	number = {4},
	pages = {2599 – 2626},
	doi = {10.3150/22-BEJ1554},
}

@ARTICLE{Cai20221435,
	author = {Cai, Xiong and Xue, Liugen and Cao, Jiguo},
	title = {VARIABLE SELECTION FOR MULTIPLE FUNCTION-ON-FUNCTION LINEAR REGRESSION},
	year = {2022},
	journal = {Statistica Sinica},
	volume = {32},
	number = {3},
	pages = {1435 – 1465},
	doi = {10.5705/ss.202020.0473},
}

@article{Patilea01102016,
author = {Valentin Patilea and César Sánchez-Sellero and Matthieu Saumard},
title = {Testing the Predictor Effect on a Functional Response},
journal = {Journal of the American Statistical Association},
volume = {111},
number = {516},
pages = {1684--1695},
year = {2016},
doi = {10.1080/01621459.2015.1110031},
}

@article{annurev-statistics-041715-033624,
   author = "Wang, Jane-Ling and Chiou, Jeng-Min and Müller, Hans-Georg",
   title = "Functional Data Analysis",
   journal= "Annual Review of Statistics and Its Application",
   year = "2016",
   volume = "3",
   number = "Volume 3, 2016",
   pages = "257-295",
   doi = "10.1146/annurev-statistics-041715-033624",
  }

@ARTICLE{Zhang2023TKDD,
  author =       {Mimi Zhang and Andrew Parnell},
  title =        {Review of clustering methods for functional data},
  journal =      {ACM Transactions on Knowledge Discovery from Data},
  year =         {2023},
  volume =       {17},
  number =       {7},
  pages =        {34},
  doi =        {10.1145/3581789},
}

@article{83c3d63f-3b1e-372a,
 author = {Jeng-Min Chiou and Yu-Ting Chen and Ya-Fang Yang},
 journal = {Statistica Sinica},
 number = {4},
 pages = {1571--1596},
 publisher = {Institute of Statistical Science, Academia Sinica},
 title = {MULTIVARIATE FUNCTIONAL PRINCIPAL COMPONENT ANALYSIS: A NORMALIZATION APPROACH},
 urldate = {2026-01-21},
 volume = {24},
 year = {2014}
}

@inproceedings{1007599,
  author={Rossi, F. and Conan-Guez, B. and Fleuret, F.},
  booktitle={ 2002 International Joint Conference on Neural Networks (IJCNN)},
  title={Functional data analysis with multi layer perceptrons},
  year={2002},
  volume={3},
  number={},
  pages={2843-2848 vol.3},
  doi={10.1109/IJCNN.2002.1007599}}

@article{10618600.2022.2097914,
author = {Barinder Thind and Kevin Multani and Jiguo Cao},
title = {Deep Learning With Functional Inputs},
journal = {Journal of Computational and Graphical Statistics},
volume = {0},
number = {0},
pages = {1-10},
year  = {2022},
publisher = {Taylor & Francis},
doi = {10.1080/10618600.2022.2097914},
}

@inproceedings{pmlr-v202-heinrichs23a,
  title = 	 {Functional Neural Networks: Shift invariant models for functional data with applications to {EEG} classification},
  author =       {Heinrichs, Florian and Heim, Mavin and Weber, Corinna},
  booktitle = 	 {Proceedings of the 40th International Conference on Machine Learning},
  pages = 	 {12866--12881},
  year = 	 {2023},
  volume = 	 {202},
  series = 	 {Proceedings of Machine Learning Research},
  month = 	 {23--29 Jul},
  publisher =    {PMLR},
}

@article{10.1002/mrm.20965,
author = {Arsigny, Vincent and Fillard, Pierre and Pennec, Xavier and Ayache, Nicholas},
title = {Log-Euclidean metrics for fast and simple calculus on diffusion tensors},
journal = {Magnetic Resonance in Medicine},
volume = {56},
number = {2},
pages = {411-421},
doi = {10.1002/mrm.20965},
year = {2006}
}

@ARTICLE{4479482,
  author={Tuzel, Oncel and Porikli, Fatih and Meer, Peter},
  journal={IEEE Transactions on Pattern Analysis and Machine Intelligence}, 
  title={Pedestrian Detection via Classification on Riemannian Manifolds}, 
  year={2008},
  volume={30},
  number={10},
  pages={1713-1727},
  doi={10.1109/TPAMI.2008.75}}

@inproceedings{10.5555/3045118.3045196,
author = {Huang, Zhiwu and Wang, Ruiping and Shan, Shiguang and Li, Xianqiu and Chen, Xilin},
title = {Log-euclidean metric learning on symmetric positive definite manifold with application to image set classification},
year = {2015},
publisher = {JMLR.org},
booktitle = {Proceedings of the 32nd International Conference on International Conference on Machine Learning - Volume 37},
pages = {720–729},
numpages = {10},
}

@InProceedings{Wang_2025_CVPR,
    author    = {Wang, Rui and Jin, Shaocheng and Chen, Ziheng and Luo, Xiaoqing and Wu, Xiao-Jun},
    title     = {Learning to Normalize on the SPD Manifold under Bures-Wasserstein Geometry},
    booktitle = {Proceedings of the IEEE/CVF Conference on Computer Vision and Pattern Recognition (CVPR)},
    month     = {June},
    year      = {2025},
    pages     = {8289-8298}
}

@inproceedings{aaai.v37i6.25867,
author = {Chen, Ziheng and Xu, Tianyang and Wu, Xiao-Jun and Wang, Rui and Huang, Zhiwu and Kittler, Josef},
title = {Riemannian local mechanism for SPD neural networks},
year = {2023},
publisher = {AAAI Press},
doi = {10.1609/aaai.v37i6.25867},
booktitle = {Proceedings of the Thirty-Seventh AAAI Conference on Artificial Intelligence and Thirty-Fifth Conference on Innovative Applications of Artificial Intelligence and Thirteenth Symposium on Educational Advances in Artificial Intelligence},
articleno = {798},
numpages = {9},
}

@inproceedings{v33i01.33018746,
author = {Liu, Hong and Li, Jie and Wu, Yongjian and Ji, Rongrong},
title = {Learning neural bag-of-matrix-summarization with riemannian network},
year = {2019},
publisher = {AAAI Press},
doi = {10.1609/aaai.v33i01.33018746},
booktitle = {Proceedings of the Thirty-Third AAAI Conference on Artificial Intelligence and Thirty-First Innovative Applications of Artificial Intelligence Conference and Ninth AAAI Symposium on Educational Advances in Artificial Intelligence},
articleno = {1073},
numpages = {8},
}

@inproceedings{ICLR2024_b1656d20,
 author = {Nguyen, Xuan Son and Yang, Yang and Histace, Aymeric},
 booktitle = {International Conference on Learning Representations},
 editor = {B. Kim and Y. Yue and S. Chaudhuri and K. Fragkiadaki and M. Khan and Y. Sun},
 pages = {40448--40478},
 title = {Matrix Manifold Neural Networks++},
 volume = {2024},
 year = {2024}
}

@inproceedings{jang2023geometrically,
title={Geometrically regularized autoencoders for non-Euclidean data},
author={Cheongjae Jang and Yonghyeon Lee and Yung-Kyun Noh and Frank C. Park},
booktitle={The Eleventh International Conference on Learning Representations },
year={2023},
}

@InProceedings{pmlr-v202-bonet23a,
  title = 	 {Sliced-{W}asserstein on Symmetric Positive Definite Matrices for {M}/{EEG} Signals},
  author =       {Bonet, Cl\'{e}ment and Mal\'{e}zieux, Beno\^{\i}t and Rakotomamonjy, Alain and Drumetz, Lucas and Moreau, Thomas and Kowalski, Matthieu and Courty, Nicolas},
  booktitle = 	 {Proceedings of the 40th International Conference on Machine Learning},
  pages = 	 {2777--2805},
  year = 	 {2023},
  editor = 	 {Krause, Andreas and Brunskill, Emma and Cho, Kyunghyun and Engelhardt, Barbara and Sabato, Sivan and Scarlett, Jonathan},
  volume = 	 {202},
  series = 	 {Proceedings of Machine Learning Research},
  month = 	 {23--29 Jul},
  publisher =    {PMLR},
}

@ARTICLE{6479703,
  author={Guo, Kai and Ishwar, Prakash and Konrad, Janusz},
  journal={IEEE Transactions on Image Processing}, 
  title={Action Recognition From Video Using Feature Covariance Matrices}, 
  year={2013},
  volume={22},
  number={6},
  pages={2479-2494},
  doi={10.1109/TIP.2013.2252622}}

@dataset{Poldrack2016phenomics,
  author = {Bilder, R AND Poldrack, R AND Cannon, T AND London, E AND Freimer, N AND Congdon, E AND Karlsgodt, K AND Sabb, F},
  title = {"UCLA Consortium for Neuropsychiatric Phenomics LA5c Study"},
  year = {2020},
  doi = {10.18112/openneuro.ds000030.v1.0.0},
  publisher = {OpenNeuro}
}

@article{Gorgolewski2017preprocessedCNP,
  author = {Gorgolewski, Krzysztof J. and Durnez, Joke and Poldrack, Russell A.},
  title = {Preprocessed {Consortium} for {Neuropsychiatric Phenomics} dataset},
  journal = {F1000Research},
  volume = {6},
  pages = {1262},
  year = {2017},
  doi = {10.12688/f1000research.11964.2},
}

@article{ADHD2002012,
  author  = {{The ADHD-200 Consortium}},
  title   = {The ADHD-200 Consortium: A Model to Advance the Translational Potential of Neuroimaging in Clinical Neuroscience},
  journal = {Frontiers in Systems Neuroscience},
  year    = {2012},
  volume  = {6},
  pages   = {62},
  doi     = {10.3389/fnsys.2012.00062}
}

@article{Bellec2017ADHD200,
  author  = {Bellec, Pierre and Chu, Carlton and Chouinard-Decorte, Fran{\c{c}}ois and Benhajali, Yassine and Margulies, Daniel S. and Craddock, R. Cameron},
  title   = {The Neuro Bureau ADHD-200 Preprocessed Repository},
  journal = {NeuroImage},
  year    = {2017},
  volume  = {144},
  number  = {Part B},
  pages   = {275--286},
  doi     = {10.1016/j.neuroimage.2016.06.034}
}

@article{DiMartino2014abide,
  author = {Di Martino, A. and Yan, C.-G. and Li, Q. and Denio, E. and Castellanos, F. X. and Alaerts, K. and Anderson, J. S. and Assaf, M. and Bookheimer, S. Y. and Dapretto, M. and Deen, B. and Delmonte, S. and Dinstein, I. and Ertl-Wagner, B. and Fair, D. A. and Gallagher, L. and Kennedy, D. P. and Keown, C. L. and Keysers, C. and Lainhart, J. E. and Lord, C. and Luna, B. and Menon, V. and Minshew, N. J. and Monk, C. S. and Mueller, S. and M{\"u}ller, R.-A. and Nebel, M. B. and Nigg, J. T. and O'Hearn, K. and Pelphrey, K. A. and Peltier, S. J. and Rudie, J. D. and Sunaert, S. and Thioux, M. and Tyszka, J. M. and Uddin, L. Q. and Verhoeven, J. S. and Wenderoth, N. and Wiggins, J. L. and Mostofsky, S. H. and Milham, M. P.},
  title = {The autism brain imaging data exchange: towards a large-scale evaluation of the intrinsic brain architecture in autism},
  journal = {Molecular Psychiatry},
  volume = {19},
  number = {6},
  pages = {659--667},
  year = {2014},
  doi = {10.1038/mp.2013.78},
}

@article{Yarkoni2011neurosynth,
  author = {Yarkoni, Tal and Poldrack, Russell A. and Nichols, Thomas E. and Van Essen, David C. and Wager, Tor D.},
  title = {Large-scale automated synthesis of human functional neuroimaging data},
  journal = {Nature Methods},
  volume = {8},
  number = {8},
  pages = {665--670},
  year = {2011},
  doi = {10.1038/nmeth.1635},
}

@article{Salo2023nimare,
  author = {Salo, Taylor and Yarkoni, Tal and Nichols, Thomas E. and Poline, Jean-Baptiste and Bilgel, Murat and Bottenhorn, Katherine L. and Eickhoff, Simon B. and Jarecka, Dorota and Kent, James D. and Kimbler, Adam and Nielson, Dylan M. and Oudyk, Kendra M. and Peraza, Julio A. and Pérez, Alexandre and Reeders, Puck C. and Yanes, Julio A. and Laird, Angela R.},
  title = {{NiMARE}: {N}euroimaging {M}eta-{A}nalysis {R}esearch {E}nvironment},
  journal = {Aperture Neuro},
  volume = {3},
  pages = {1--32},
  year = {2023},
  doi = {10.52294/001c.87681},
}

@article{Chopra2024tcp,
  author = {Chopra, Sidhant and Cocuzza, Carrisa V. and Lawhead, Connor and Ricard, Jocelyn A. and Labache, Lo{\"i}c and Patrick, Lauren M. and Kumar, Poornima and Rubenstein, Arielle and Moses, Julia and Chen, Lia and Blankenbaker, Crystal and Gillis, Bryce and Germine, Laura T. and Harpaz-Rotem, Ilan and Yeo, B. T. Thomas and Baker, Justin T. and Holmes, Avram J.},
  title = {The {Transdiagnostic} {Connectome} {Project}: an open dataset for studying brain-behavior relationships in psychiatry},
  journal = {Scientific Data},
  volume = {12},
  number = {1},
  pages = {923},
  year = {2025},
  doi = {10.1038/s41597-025-04895-z},
  note = {OpenNeuro \texttt{ds005237}}
}

@article{Glasser2016MMP,
  author = {Glasser, Matthew F. and Coalson, Timothy S. and Robinson, Emma C. and Hacker, Carl D. and Harwell, John and Yacoub, Essa and Ugurbil, Kamil and Andersson, Jesper and Beckmann, Christian F. and Jenkinson, Mark and Smith, Stephen M. and Van Essen, David C.},
  title = {A multi-modal parcellation of human cerebral cortex},
  journal = {Nature},
  volume = {536},
  number = {7615},
  pages = {171--178},
  year = {2016},
  doi = {10.1038/nature18933}
}

@article{SalimiKhorshidi2014icafix,
  author = {Salimi-Khorshidi, Gholamreza and Douaud, Gwena{\"e}lle and Beckmann, Christian F. and Glasser, Matthew F. and Griffanti, Ludovica and Smith, Stephen M.},
  title = {Automatic denoising of functional {MRI} data: combining independent component analysis and hierarchical fusion of classifiers},
  journal = {NeuroImage},
  volume = {90},
  pages = {449--468},
  year = {2014},
  doi = {10.1016/j.neuroimage.2013.11.046}
}

@article{Sadeghi2022catd, 
    author = {Sadeghi, Neda and Fors, Payton Q. and Eisner, Lillian and Taigman, Jeremy and Qi, Karen and Gorham, Lisa S. and Camp, Christopher C. and O'Callaghan, Georgia and Rodriguez, Diana and McGuire, Jerry and Garth, Erin M. and Engel, Chana and Davis, Mollie and Towbin, Kenneth E. and Stringaris, Argyris and Nielson, Dylan M.}, 
    title = {Mood and Behaviors of Adolescents With Depression in a Longitudinal Study Before and During the {COVID-19} Pandemic}, 
    journal = {Journal of the American Academy of Child {\&} Adolescent Psychiatry}, 
    volume = {61}, 
    number = {11}, 
    pages = {1341--1350}, 
    year = {2022}, 
    doi = {10.1016/j.jaac.2022.04.004}, 
    note = {OpenNeuro \texttt{ds004627}} 
}

@article{Camp2024catd, 
    author = {Camp, Chris C. and Noble, Stephanie and Scheinost, Dustin and Stringaris, Argyris and Nielson, Dylan M.}, 
    title = {Test-Retest Reliability of Functional Connectivity in Adolescents With Depression}, 
    journal = {Biological Psychiatry: Cognitive Neuroscience and Neuroimaging}, 
    volume = {9}, 
    number = {1}, 
    pages = {21--29}, 
    year = {2024}, 
    doi = {10.1016/j.bpsc.2023.09.002} 
}

@article{Esteban2019fmriprep, 
    author = {Esteban, Oscar and Markiewicz, Christopher J. and Blair, Ross W. and Moodie, Craig A. and Isik, A. Ilkay and Erramuzpe, Asier and Kent, James D. and Goncalves, Mathias and DuPre, Elizabeth and Snyder, Madeleine and Oya, Hiroyuki and Ghosh, Satrajit S. and Wright, Jessey and Durnez, Joke and Poldrack, Russell A. and Gorgolewski, Krzysztof J.}, 
    title = {{fMRIPrep}: a robust preprocessing pipeline for functional {MRI}}, 
    journal = {Nature Methods}, 
    volume = {16}, 
    number = {1}, 
    pages = {111--116}, 
    year = {2019}, 
    doi = {10.1038/s41592-018-0235-4} 
}

@article{Heinsfeld2018ABIDE,
    author  = {Heinsfeld, Anibal S\'{o}lon and Franco, Alexandre Rosa and Craddock, R. Cameron and Buchweitz, Augusto and Meneguzzi, Felipe},
    title   = {Identification of autism spectrum disorder using deep learning and the {ABIDE} dataset},
    journal = {NeuroImage: Clinical},
    volume  = {17},
    pages   = {16--23},
    year    = {2018},
    doi     = {10.1016/j.nicl.2017.08.017}
}

@article{Zeng2018ebiomed,
    author  = {Zeng, Ling-Li and Wang, Huaning and Hu, Panpan and Yang, Bo and Pu, Weidan and Shen, Hui and Chen, Xingui and Liu, Zhening and Yin, Hong and Tan, Qingrong and Wang, Kai and Hu, Dewen},
    title   = {Multi-Site Diagnostic Classification of Schizophrenia Using Discriminant Deep Learning with Functional Connectivity {MRI}},
    journal = {EBioMedicine},
    volume  = {30},
    pages   = {74--85},
    year    = {2018},
    doi     = {10.1016/j.ebiom.2018.03.017}
}

@article{Saponaro2022harmonization,
    author  = {Saponaro, Sara and Giuliano, Alessia and Bellotti, Roberto and Lombardi, Angela and Tangaro, Sabina and Oliva, Piernicola and Calderoni, Sara and Retico, Alessandra},
    title   = {Multi-site harmonization of {MRI} data uncovers machine-learning discrimination capability in barely separable populations: An example from the {ABIDE} dataset},
    journal = {NeuroImage: Clinical},
    volume  = {35},
    pages   = {103082},
    year    = {2022},
    doi     = {10.1016/j.nicl.2022.103082}
}

@article{Qiu2024ASTNet,
    author  = {Qiu, Bo and Wang, Qianqian and Li, Xizhi and Li, Wenyang and Shao, Wei and Wang, Mingliang},
    title   = {Adaptive spatial-temporal neural network for {ADHD} identification using functional {fMRI}},
    journal = {Frontiers in Neuroscience},
    volume  = {18},
    pages   = {1394234},
    year    = {2024},
    doi     = {10.3389/fnins.2024.1394234}
}

%%%%%%%%%%%%%%%%%%%%%%%%%%%%%%%%%%%%%%%%%%%%%%%%%%%%%%%%%%%%
\newpage
\appendix

\section{Related work}\label{Related_work}
\paragraph{Functional data analysis} Functional data analysis studies data whose basic observational units are functions (e.g., curves over time or space) and, more broadly, other complex objects such as images and shapes. In functional data analysis, each subject in a sample is represented by one or multiple functions observed over a continuum (such as time, wavelength, or spatial location). Let $\{\pmb{y}_i: i=1, \ldots, n\}$ be a set of $n$ independent $p$-dimensional sample functions: $\pmb{y}_i(t)=(y_i^1(t), \ldots, y_i^p(t))^{\top}\in\mathbb{R}^p$ for any $t\in [t_0, t_1]$; the superscript $\top$ is the transpose operator. Each sample function $\pmb{y}_i$ is a random realization of a $p$-dimensional random function $\vec{Y}=(Y^1, \ldots, Y^p)^{\top}$. For $d=1, \ldots, p$, the $n$ one-dimensional sample functions $\{y^d_1, \ldots, y^d_n\}$ are independent realizations of the component random function $Y^d=Y^d(t)=Y^d(t, \omega)$, defined on a probability space $(\Omega, \mathcal{F}, \Pr)$ and taking values in $\mathcal{H}([t_0, t_1], \mathbb{R})$. Here, $\mathcal{H}([t_0, t_1], \mathbb{R})$ is the separable Hilbert space of all square-integrable measurable functions that are defined on $[t_0, t_1]$ and taking values in $\mathbb{R}$. That is, $Y^d$ is a measurable map from $\Omega$ to $\mathcal{H}([t_0, t_1], \mathbb{R})$. %Alternatively, we can view the function value $y^d_i(t)$ as a realization of the random variable $Y^d(t, \cdot)$, a mapping from $(\Omega, \mathcal{F})$ to $(\mathbb{R}, \mathcal{B}_{\mathbb{R}})$, where $\mathcal{B}_{\mathbb{R}}$ is the Borel $\sigma$-algebra of $\mathbb{R}$. 

Functional regression is commonly classified by whether the response and the covariates are functional or finite-dimensional (vector/scalar). This yields three main settings: (i) functional responses with functional covariates \cite{Dette2024304, Cai20221435}, (ii) vector (including scalar) responses with functional covariates \cite{Hall200770, Yeon20232599, Balasubramanian2025973}, and (iii) functional responses with vector covariates \cite{Patilea01102016}. Among these, the most extensively studied case is (ii), particularly scalar-on-function regression, where the response is a scalar and the predictor is a function. For an overview of statistical tools in functional data analysis, see \cite{annurev-statistics-041715-033624}. Beyond regression, clustering is a core task in functional data analysis. \citet{Zhang2023TKDD} provided a comprehensive review of clustering methods for functional data and observed that approaches for clustering multi-dimensional functional data generally follow one of two paradigms: (1) first applying multivariate functional principal component analysis \cite{83c3d63f-3b1e-372a} and then clustering the resulting score vectors using standard multivariate clustering methods; or (2) defining an appropriate (dis)similarity measure for multi-dimensional functional data and clustering directly using distance- or similarity-based algorithms.

More recently, functional neural networks have been proposed to capture nonlinear relationships that may be difficult to represent with classical functional data analysis models \cite{1007599, 10618600.2022.2097914, pmlr-v202-heinrichs23a}. \citet{singh2025NeurIPS} provided the first proof that functional neural network has the universal approximation property. A key distinction of these architectures is the inclusion of functional layers, in which the learnable weights are themselves functions rather than scalar parameters, enabling the architectures operate directly on functional inputs. Although one can treat the sequence of discrete evaluations of each $y^d_i(t)$ over a grid  as a time series, enabling the use of recurrent neural networks or Transformer-based models, such approaches do not naturally encode shape (morphological) characteristics or derivative-based (first-order) information of the underlying functions. By contrast, functional neural networks offer a principled framework for learning representations that respect the functional nature of the data rather than relying solely on discrete measurements.

\paragraph{Riemannian neural networks} Deep neural networks for SPD matrices must respect the fact that the data live on a Riemannian manifold. Over the past decade, two main modeling paradigms have emerged: tangent-space learning and intrinsic manifold deep networks, with more recent work extending these ideas to attention, normalization, and alternative geometries. In the tangent-space learning paradigm, we map an SPD matrix to a vector space using a Riemannian logarithm map, then apply standard neural architectures on the mapped data \cite{10.1002/mrm.20965, 4479482, 10.5555/3045118.3045196}. This approach is computationally attractive  because it avoids repeated eigen/SVD constraints inside the network while keeping a principled geometric link to the manifold. In the intrinsic deep learning paradigm, every operation preserves positive definiteness. The canonical example is SPDNet \cite{10.5555/3298483.3298534}, which introduced a stack of SPD-preserving layers: bilinear mapping (BiMap) for dimension reduction on SPD, eigenvalue rectification (ReEig) as a nonlinearity, and log-eigenvalue (LogEig) to move to a Euclidean space only near the output for standard classifiers. This architecture family has been extended in multiple directions: regularization on manifolds \cite{jang2023geometrically}, Riemannian batch normalization on the SPD manifold \cite{NEURIPS2019_6e69ebbf, Wang_2025_CVPR}, a multi-channel mechanism to capture multi-scale features \cite{aaai.v37i6.25867}, and replacing LogEig with a learnable SPD encoder \cite{v33i01.33018746}. Another emerging direction is to broaden the layer toolbox (e.g., convolutional layers) to make SPD networks more closely aligned with standard deep learning building blocks, while still ensuring valid manifold backpropagation \cite{ICLR2024_b1656d20}.

To the best of our knowledge, there is currently no deep neural network architecture specifically designed to model SPD-valued trajectories \(\{\tX_i\in\mathcal{H}_g([t_0, t_1]; \mathcal{S}^m_+)\}_{i=1}^n\), where each observation $\tX_i$ is a function of time, and its value $\tX_i(t)$ is an SPD matrix. Such SPD-valued functional data arise naturally when second-order statistics evolve continuously over time, and they are increasingly encountered in modern data-rich scientific domains, most notably neuroscience. A representative example is magnetoencephalography (MEG) or electroencephalography (EEG) analysis. Let \(\pmb{x}_i(t)\in \mathbb{R}^m\) denote the multichannel sensor signal of subject $i$ at time $t$. Instead of analyzing raw signals, it is common to compute time-resolved covariance or cross-spectral matrices $\tX_i(t)$ over short, possibly overlapping, temporal windows. As the window slides continuously, this procedure yields a smooth trajectory of SPD matrices over time, capturing the dynamic evolution of brain networks during cognitive tasks, motor imagery, or resting-state activity \cite{pmlr-v202-bonet23a}. In this work, we develop \pkg{MatFAE} to bridge this methodological gap by unifying recent advances in Riemannian deep learning and functional data analysis for end-to-end modeling of SPD-valued trajectories.

\section{Properties of the two Riemannian manifolds }\label{Properties_of_the_two_metrics}

\subsection{The Log-Cholesky metric}
Let \(\mathcal{L}^m\) denote the set of \(m\times m\) lower triangular matrices, and \(\mathcal{L}^m_+\subset\mathcal{L}^m\) the subset of lower triangular matrices with positive diagonal elements. Both \(\mathcal{L}^m_+\) and \(\mathcal{S}^m_+\)  are smooth manifolds  and therefore have differential geometry properties. The tangent space \(\mathcal{T}_{\mathbf{L}}\mathcal{L}^m_+\) of \(\mathcal{L}^m_+\) at any point \(\mathbf{L}\in\mathcal{L}^m_+\) is \(\mathcal{L}^m\). For any \(\mathbf{X}\in\mathcal{S}^m_+\), there is a unique lower triangular matrix with positive diagonal elements \(\mathbf{L}=\mathfrak{L}(\mathbf{X})\in\mathcal{L}^m_+\) that satisfies \(\mathbf{X}=\mathbf{L}\mathbf{L}^{\top}\). Furthermore, the map \(\mathfrak{L}: \mathcal{S}^m_+\rightarrow\mathcal{L}^m_+\) is a diffeomorphism \citep[Proposition 2]{doi:10.1137/18M1221084}. The differential \(D_{\mathbf{X}}\mathfrak{L}: \mathcal{T}_{\mathbf{X}}\mathcal{S}^m_+\mapsto\mathcal{T}_{\mathfrak{L}(\mathbf{X})}\mathcal{L}^m_+\) of \(\mathfrak{L}\) at \(\mathbf{X}\) is given by
\[
(D_{\mathbf{X}}\mathfrak{L})(\mathbf{U})=\mathfrak{L}(\mathbf{X})(\mathfrak{L}(\mathbf{X})^{-1}\mathbf{U}\mathfrak{L}(\mathbf{X})^{-\top})_{\frac{1}{2}}, ~~\mathbf{U}\in\mathcal{T}_{\mathbf{X}}\mathcal{S}^m_+,
\]
where \((\mathbf{V})_{\frac{1}{2}}=\lfloor\mathbf{V}\rfloor+\frac{1}{2}\text{diag}(\mathbf{V})\) is the lower triangular part of \(\mathbf{V}\) with the diagonal elements halved, and \(\text{diag}(\mathbf{V})\) is the matrix of the main diagonal elements of \(\mathbf{V}\). Let \(\mathfrak{L}^{-1}: \mathcal{L}^m_+\mapsto\mathcal{S}^m_+\) denote the inverse of the map \(\mathfrak{L}\): \(\mathfrak{L}^{-1}(\mathbf{L})=\mathbf{L}\mathbf{L}^{\top}\). The differential \(D_{\mathbf{L}}\mathfrak{L}^{-1}: \mathcal{T}_{\mathbf{L}}\mathcal{L}^m_+\mapsto\mathcal{T}_{\mathbf{L}\mathbf{L}^{\top}}\mathcal{S}^m_+\) of \(\mathfrak{L}^{-1}\) at \(\mathbf{L}\) is given by
\[
(D_{\mathbf{L}}\mathfrak{L}^{-1})(\mathbf{U})=\mathbf{L}\mathbf{U}^{\top}+\mathbf{U}\mathbf{L}^{\top}, ~~\mathbf{U}\in\mathcal{T}_{\mathbf{L}}\mathcal{L}^m_+.
\]
Because \(\mathfrak{L}\) is a diffeomorphism, the Riemannian metric for \(\mathcal{S}^m_+\) can be induced (via pullback) from the Riemannian metric for \(\mathcal{L}^m_+\). 

For the manifold \(\mathcal{L}^m_+\), the Riemannian metric at a point \(\mathbf{L}\in\mathcal{L}^m_+\) is
\[\langle\mathbf{U}, \mathbf{V}\rangle_{\mathbf{L}}^{\mathcal{L}^m_+}=\langle \lfloor\mathbf{U}\rfloor, \lfloor\mathbf{V}\rfloor \rangle _{F} +\langle \text{diag}(\mathbf{L})^{-1}\text{diag}(\mathbf{U}), \text{diag}(\mathbf{L})^{-1}\text{diag}(\mathbf{V}) \rangle _{F}, ~~\mathbf{U}, \mathbf{V}\in \mathcal{T}_{\mathbf{L}}\mathcal{L}^m_+,
\]
where \(\langle \cdot, \cdot\rangle_F\) is the Frobenius inner product, with induced norm \(\|\cdot\|_F\). The Riemannian exponential map at \(\mathbf{L}\)  is 
\[\text{Exp}_{\mathbf{L}}(\mathbf{U})=\lfloor\mathbf{L}\rfloor+\lfloor\mathbf{U}\rfloor+\text{diag}(\mathbf{L})\exp(\text{diag}(\mathbf{U})\text{diag}(\mathbf{L})^{-1}), ~~\mathbf{U}\in\mathcal{T}_{\mathbf{L}}\mathcal{L}^m_+,
\]
and the Riemannian logarithmic map at \(\mathbf{L}\)  is 
\[
\text{Log}_{\mathbf{L}}(\mathbf{K})=\lfloor\mathbf{K}\rfloor-\lfloor\mathbf{L}\rfloor+\text{diag}(\mathbf{L})\log(\text{diag}(\mathbf{L})^{-1}\text{diag}(\mathbf{K})), ~~\mathbf{K}\in\mathcal{L}^m_+.
\]
The geodesic distance between two \(\mathbf{L}\) and \(\mathbf{K}\in\mathcal{L}^m_+\) is
\[
d_{\mathcal{L}^m_+}(\mathbf{L}, \mathbf{K})=\sqrt{\|\lfloor\mathbf{L}\rfloor-\lfloor\mathbf{K}\rfloor\|_F^2+\|\log(\text{diag}(\mathbf{L}))-\log(\text{diag}(\mathbf{K}))\|_F^2}.
\]

\paragraph{Riemannian geometry} Given the Riemannian metric \(\langle\cdot, \cdot\rangle_{\cdot}^{\mathcal{L}^m_+}\) for \(\mathcal{L}^m_+\), the manifold map \(\mathfrak{L}: \mathcal{S}^m_+\mapsto\mathcal{L}^m_+\) induces the Riemannian metric for \(\mathcal{S}^m_+\), called the Log-Cholesky (LC) metric:
\begin{equation*}\label{LC_metric}
    \langle \mathbf{U}, \mathbf{V}\rangle_{\mathbf{X}}^{\text{LC}}=\langle (D_{\mathbf{X}}\mathfrak{L})(\mathbf{U}), (D_{\mathbf{X}}\mathfrak{L})(\mathbf{V})\rangle_{\mathfrak{L}(\mathbf{X})}^{\mathcal{L}^m_+}, ~~\mathbf{U}, \mathbf{V}\in \mathcal{T}_{\mathbf{X}}\mathcal{S}^m_+.
\end{equation*}
The Riemannian exponential map at \(\mathbf{X}\in\mathcal{S}^m_+\)  is
\begin{equation*}\label{LC_Exp}
    \text{Exp}^{\text{LC}}_{\mathbf{X}}(\mathbf{V})=[\text{Exp}_{\mathfrak{L}(\mathbf{X})}((D_{\mathbf{X}}\mathfrak{L})(\mathbf{V}))][\text{Exp}_{\mathfrak{L}(\mathbf{X})}((D_{\mathbf{X}}\mathfrak{L})(\mathbf{V}))]^{\top}, ~~\mathbf{V}\in\mathcal{T}_{\mathbf{X}}\mathcal{S}^m_+.
\end{equation*}
The Riemannian logarithmic map at \(\mathbf{X}\) is
\begin{equation}\label{LC_Log}
    \text{Log}^{\text{LC}}_{\mathbf{X}}(\mathbf{Z})=(D_{\mathfrak{L}(\mathbf{X})}\mathfrak{L}^{-1})(\text{Log}_{\mathfrak{L}(\mathbf{X})}(\mathfrak{L}(\mathbf{Z}))), ~~\mathbf{Z}\in \mathcal{S}^m_+.
\end{equation}
When the reference point is the identity matrix \(\mathbf{I}\), the Riemannian logarithmic map of \(\mathbf{Z}\) is 
\begin{equation}\label{LC_Log_Iden}
    \text{Log}^{\text{LC}}_{\mathbf{I}}(\mathbf{Z})=\lfloor\mathfrak{L}(\mathbf{Z})\rfloor+\lfloor\mathfrak{L}(\mathbf{Z})\rfloor^{\top} +2\log(\text{diag}(\mathfrak{L}(\mathbf{Z}))).
\end{equation}
The map \(\mathfrak{L}\) is an isometry between \((\mathcal{S}^m_+, \langle\cdot, \cdot\rangle_{\cdot}^{\text{LC}})\) and \((\mathcal{L}^m_+, \langle\cdot, \cdot\rangle_{\cdot}^{\mathcal{L}^m_+})\), and therefore the geodesic distance between \(\mathbf{X}, \mathbf{Z}\in\mathcal{S}^m_+\) is simply \(d_{\text{LC}}(\mathbf{X}, \mathbf{Z})=d_{\mathcal{L}^m_+}(\mathfrak{L}(\mathbf{X}), \mathfrak{L}(\mathbf{Z}))\).

\paragraph{Fr\'{e}chet mean} Given a set of SPD matrices \(\{\mathbf{X}_i\in\mathcal{S}^m_+: i=1, \ldots, n\}\), under mild conditions, the Fr\'{e}chet mean \(\mathbb{E}_{\text{LC}}(\mathbf{X}_{1:n})\) exits and is unique: 
\[\mathbb{E}_{\text{LC}}(\mathbf{X}_{1:n})=\argmin_{\mathbf{X}\in\mathcal{S}^m_+}\sum_{i=1}^nd_{\text{LC}}(\mathbf{X}, \mathbf{X}_i)^2=\mathbf{L}_n\mathbf{L}_n^{\top},\] 
where
\begin{equation}\label{LC_Frechet_mean}
    \mathbf{L}_n=\frac{1}{n}\sum_{i=1}^n\lfloor\mathfrak{L}(\mathbf{X}_i)\rfloor+\exp(\frac{1}{n}\sum_{i=1}^n\log(\text{diag}(\mathfrak{L}(\mathbf{X}_i)))).
\end{equation}
The sectional curvature of \((\mathcal{S}^m_+, \langle\cdot, \cdot\rangle_{\cdot}^{\text{LC}})\) is constantly zero \citep[Proposition 8]{doi:10.1137/18M1221084}, and therefore \((\mathcal{S}^m_+, \langle\cdot, \cdot\rangle_{\cdot}^{\text{LC}})\) is a Cartan-Hadamard manifold.

\subsection{The Log-Euclidean metric}
Every real symmetric matrix \(\mathbf{U}\in\text{Sym}(m)\) admits an orthogonal eigen-decomposition:  \(\mathbf{U}=\mathbf{Q}\text{diag}(\lambda_1, \ldots, \lambda_m)\mathbf{Q}^{\top}\), and therefore taking the matrix exponential can be done by simply applying its scalar version to eigenvalues: \[\exp(\mathbf{U})=\mathbf{Q}\text{diag}(\exp(\lambda_1), \ldots, \exp(\lambda_m))\mathbf{Q}^{\top}.\] Furthermore, for an SPD matrix \(\mathbf{X}\in\mathcal{S}^m_+\subset\text{Sym}(m)\), the (principal) matrix logarithm exists, is unique, and also admits the eigen-mapping form: 
\[\log(\mathbf{X})=\mathbf{Q}\text{diag}(\log(\gamma_1), \ldots, \log(\gamma_m))\mathbf{Q}^{\top},\] where \(\mathbf{X}=\mathbf{Q}\text{diag}(\gamma_1, \ldots, \gamma_m)\mathbf{Q}^{\top}\) is the eigen-decomposition. Both \(\exp: \text{Sym}(m)\mapsto\mathcal{S}^m_+\) and its inverse \(\log: \mathcal{S}^m_+\mapsto\text{Sym}(m)\) are diffeomorphisms. Let \((D_{\mathbf{X}}\log)(\mathbf{U})\) be the differential (Fr\'{e}chet derivative) of the matrix logarithm at the reference point \(\mathbf{X}\), evaluated in the direction \(\mathbf{U}\), and \((D_{\mathbf{U}}\exp)(\mathbf{V})\) the differential of the matrix exponential. 

Let \(\mathbf{X}=\mathbf{Q}\mathbf{\Lambda}\mathbf{Q}^{\top}\) with \(\mathbf{\Lambda}=\text{diag}(\gamma_1, \ldots, \gamma_m)\). The differential \(D_{\mathbf{X}}\log: \mathcal{T}_{\mathbf{X}}\mathcal{S}^m_+\mapsto\mathcal{T}_{\log(\mathbf{X})}\text{Sym}(m)\) of the matrix logarithm at \(\mathbf{X}\) is given by
\begin{equation}\label{differential_derivative_log}  
(D_{\mathbf{X}}\log)(\mathbf{U})=\mathbf{Q}(\hat{\mathbf{\Lambda}}\odot\tilde{\mathbf{U}})\mathbf{Q}^{\top}, ~~\mathbf{U}\in\mathcal{T}_{\mathbf{X}}\mathcal{S}^m_+,
\end{equation}
where \(\odot\) is the Hadamard product,  \(\tilde{\mathbf{U}}=\mathbf{Q}^{\top}\mathbf{U}\mathbf{Q}\), and the Loewner matrix  \(\hat{\mathbf{\Lambda}}\) has diagonal entries \(\hat{\Lambda}_{i,i}=\gamma_i^{-1}\) and off-diagonal entries \(\hat{\Lambda}_{i,j}=\frac{\log(\gamma_i)-\log(\gamma_j)}{\gamma_i-\gamma_j}\). 

Let \(\mathbf{U}=\mathbf{Q}\mathbf{\Lambda}\mathbf{Q}^{\top}\) with \(\mathbf{\Lambda}=\text{diag}(\lambda_1, \ldots, \lambda_m)\). The differential \(D_{\mathbf{U}}\exp: \mathcal{T}_{\mathbf{U}}\text{Sym}(m)\mapsto\mathcal{T}_{\exp(\mathbf{U})}\mathcal{S}^m_+\) of matrix exponential map at \(\mathbf{U}\) is given by
\begin{equation}\label{differential_derivative_exp}  
(D_{\mathbf{U}}\exp)(\mathbf{K})=\mathbf{Q}(\hat{\mathbf{\Lambda}}\odot\tilde{\mathbf{K}})\mathbf{Q}^{\top}, ~~\mathbf{V}\in\mathcal{T}_{\mathbf{U}}\text{Sym}(m),
\end{equation}
where \(\tilde{\mathbf{K}}=\mathbf{Q}^{\top}\mathbf{K}\mathbf{Q}\), and the Loewner matrix \(\hat{\mathbf{\Lambda}}\)  has diagonal entries \(\hat{\Lambda}_{i,i}=\exp(\lambda_i)\) and off-diagonal entries \(\hat{\Lambda}_{i,j}=\frac{\exp(\lambda_i)-\exp(\lambda_j)}{\lambda_i-\lambda_j}\).

\paragraph{Riemannian geometry} The LE Riemannian metric for \(\mathcal{S}^m_+\) at \(\mathbf{X}\in\mathcal{S}^m_+\)  is
\begin{equation*}\label{LE_metric}
    \langle \mathbf{U}, \mathbf{V}\rangle_{\mathbf{X}}^{\text{LE}}=\langle (D_{\mathbf{X}}\log)(\mathbf{U}), (D_{\mathbf{X}}\log)(\mathbf{V})\rangle_F, ~~\mathbf{U}, \mathbf{V}\in\mathcal{T}_{\mathbf{X}}\mathcal{S}^m_+.
\end{equation*}
The Riemannian exponential map at \(\mathbf{X}\in\mathcal{S}^m_+\)  is
\begin{equation*}\label{LE_Exp}
    \text{Exp}^{\text{LE}}_{\mathbf{X}}(\mathbf{V})=\exp(\log(\mathbf{X})+(D_{\mathbf{X}}\log)(\mathbf{V})), ~~\mathbf{V}\in\mathcal{T}_{\mathbf{X}}\mathcal{S}^m_+.
\end{equation*}
The Riemannian logarithmic map at \(\mathbf{X}\) is
\begin{equation}\label{LE_Log}
    \text{Log}^{\text{LE}}_{\mathbf{X}}(\mathbf{Z})=(D_{\log(\mathbf{X})}\exp)(\log(\mathbf{Z})-\log(\mathbf{X})), ~~\mathbf{Z}\in \mathcal{S}^m_+.
\end{equation}
When the reference point is the identity matrix \(\mathbf{I}\), the matrix logarithm \(\log(\mathbf{I})\) is the zero matrix, and hence the differential \(D_{\log(\mathbf{X})}\exp\) is the identity linear map. Therefore, we have \(\text{Log}^{\text{LE}}_{\mathbf{I}}(\mathbf{Z})=\log(\mathbf{Z})\). The matrix exponential map \(\exp\) is an isometry between \((\mathcal{S}^m_+, \langle\cdot, \cdot\rangle_{\cdot}^{\text{LE}})\) and \((\text{Sym}(m), \langle\cdot, \cdot\rangle_F)\), and therefore the geodesic distance between \(\mathbf{X}, \mathbf{Z}\in\mathcal{S}^m_+\) is simply \(d_{\text{LE}}(\mathbf{X}, \mathbf{Z})=\|\log(\mathbf{X})- \log(\mathbf{Z})\|_F\).  

\paragraph{Fr\'{e}chet mean} Given a set of SPD matrices \(\{\mathbf{X}_i\in\mathcal{S}^m_+: i=1, \ldots, n\}\), the Fr\'{e}chet mean \(\mathbb{E}_{\text{LE}}(\mathbf{X}_{1:n})\) exits, is unique, and is simply the arithmetic mean in the domain of matrix logarithms: 
\begin{equation}\label{LE_Frechet_mean}
    \mathbb{E}_{\text{LE}}(\mathbf{X}_{1:n})=\argmin_{\mathbf{X}\in\mathcal{S}^m_+}\sum_{i=1}^nd_{\text{LE}}(\mathbf{X}, \mathbf{X}_i)^2=\exp(\frac{1}{n}\sum_{i=1}^n\log(\mathbf{X}_i)).
\end{equation}
Under the LE metric, \((\mathcal{S}^m_+, \langle\cdot, \cdot\rangle_{\cdot}^{\text{LE}})\) is globally isometric via the matrix logarithm to the Euclidean space \((\text{Sym}(m), \langle\cdot, \cdot\rangle_F)\), and hence \((\mathcal{S}^m_+, \langle\cdot, \cdot\rangle_{\cdot}^{\text{LE}})\) is a Cartan-Hadamard manifold. Moreover, the sectional curvature of \((\mathcal{S}^m_+, \langle\cdot, \cdot\rangle_{\cdot}^{\text{LE}})\) is constantly zero.

\section{Proof of Proposition \ref{not-invariant-equvariant}}\label{Properties_BiMap_layer}
\subsection{Congruence}
After conjugation of the input matrix \(\mathbf{X}\), the output of the multi-head BiMap layer is \(\{\mathbf{W}_1^h\mathbf{U}\mathbf{X}\mathbf{U}^{\top}(\mathbf{W}_1^h)^{\top}: h=1, \ldots, H\}\), where \(\mathbf{U}\in\text{O}(m)\) is an orthogonal matrix.  Each weight matrix \(\mathbf{W}_1^h\) is full row rank with orthonormal rows. Let \(\tilde{\mathbf{X}}_2\) denote the Fr\'{e}chet mean after congruence on the input \(\mathbf{X}\). If there exists an orthogonal matrix \(\mathbf{Q}\in\text{O}(m_1)\) that \(\mathbf{W}_1^h\mathbf{U}=\mathbf{Q}\mathbf{W}_1^h\) for all \(1\leq h\leq H\), then 
\begin{equation}\label{single_Q}
    \mathbf{W}_1^h\mathbf{U}\mathbf{X}\mathbf{U}^{\top}(\mathbf{W}_1^h)^{\top}=\mathbf{Q}\mathbf{W}_1^h\mathbf{X}(\mathbf{W}_1^h)^{\top}\mathbf{Q}^{\top}=\mathbf{Q}\mathbf{X}_1^h\mathbf{Q}^{\top}.
\end{equation}
Under the LE Riemainnian metric, we have \(\log(\mathbf{Q}\mathbf{X}_1^h\mathbf{Q}^{\top})=\mathbf{Q}\log(\mathbf{X}_1^h)\mathbf{Q}^{\top}\), and that the LE geodesic distance is invariant under orthogonal conjugation: 
\[d_{\text{LE}}(\mathbf{X}_2, \mathbf{X}_1^h)=d_{\text{LE}}(\mathbf{Q}\mathbf{X}_2\mathbf{Q}^{\top}, \mathbf{Q}\mathbf{X}_1^h\mathbf{Q}^{\top}).\]
Therefore, after matrix congruence, the LE Fr\'{e}chet mean is \(\tilde{\mathbf{X}}_2=\mathbf{Q}\mathbf{X}_2\mathbf{Q}^{\top}\).

Eq. (\ref{single_Q}) implies that conjugating the input by \(\mathbf{U}\) is equivalent to conjugating every head's output by \(\mathbf{Q}\). Let \(\text{Im}((\mathbf{W}_1^h)^{\top})\subset \mathbb{R}^{m_1}\) denote the subspace spanned by the rows of \(\mathbf{W}_1^h\). The relation \(\mathbf{W}_1^h\mathbf{U}=\mathbf{Q}\mathbf{W}_1^h\) forces \(\mathbf{U}~\text{Im}((\mathbf{W}_1^h)^{\top})=\text{Im}((\mathbf{W}_1^h)^{\top})\) and  \(\mathbf{W}_1^h\mathbf{U}(\mathbf{W}_1^h)^{\top}=\mathbf{Q}\); that is, each head's row-space is \(\mathbf{U}\)-invariant, and the ``compressed'' action of \(\mathbf{U}\) onto that subspace is the same \(\mathbf{Q}\) for all heads. With one head, such a \(\mathbf{Q}\) always exists. With multiple heads, asking for the same  \(\mathbf{Q}\) across all \(h\) is strong: either the subspaces \(\text{Im}((\mathbf{W}_1^h)^{\top})\) are all the same and  \(\mathbf{U}\)  acts on that subspace as  \(\mathbf{Q}\), or \(\mathbf{U}\)  acts trivially on every \(\text{Im}((\mathbf{W}_1^h)^{\top})\).

If each head rotates by its own \(\mathbf{Q}_h\in\text{O}(m_1)\):  \(\mathbf{W}_1^h\mathbf{U}=\mathbf{Q}_h\mathbf{W}_1^h\), assume additionally that the candidate Fr\'{e}chet mean \(\tilde{\mathbf{X}}_2\) commutes with every \(\mathbf{Q}_h\): \(\tilde{\mathbf{X}}_2\mathbf{Q}_h=\mathbf{Q}_h\tilde{\mathbf{X}}_2\). Then under the LE metric, we have
\begin{align*}
    \sum_{h=1}^H\|\log(\tilde{\mathbf{X}}_2) -\log(\mathbf{Q}_h\mathbf{X}_1^h\mathbf{Q}_h^{\top})\|_F^2&=\sum_{h=1}^H\|\log(\tilde{\mathbf{X}}_2) -\mathbf{Q}_h\log(\mathbf{X}_1^h)\mathbf{Q}_h^{\top}\|_F^2\\
    &=\sum_{h=1}^H\|\log(\tilde{\mathbf{X}}_2) -\log(\mathbf{X}_1^h)\|_F^2.
\end{align*}
Therefore, the LE Fr\'{e}chet mean is \(\tilde{\mathbf{X}}_2=\mathbf{X}_2\), invariant under general congruence.

The commutation requirement \(\tilde{\mathbf{X}}_2\mathbf{Q}_h=\mathbf{Q}_h\tilde{\mathbf{X}}_2\) implies that \(\tilde{\mathbf{X}}_2\) is invariant under all the head-wise rotations: \(\tilde{\mathbf{X}}_2=\mathbf{Q}_h\tilde{\mathbf{X}}_2\mathbf{Q}_h^{\top}\), for every \(h\); that is, \(\tilde{\mathbf{X}}_2\) shares a common orthonormal eigenbasis with all \(\mathbf{Q}_h\) up to blocks: the space splits into invariant blocks for the group \(\langle\mathbf{Q}_h\rangle\), and on each block \(\tilde{\mathbf{X}}_2\) is a scalar multiple of the identity. In group-theory terms, \(\tilde{\mathbf{X}}_2\) lies in the centralizer of the subgroup generated by \(\{\mathbf{Q}_h: 1\leq h\leq H\}\).

For the LC Riemannian metric, because the LC geometry depends on Cholesky factors of the matrices \(\mathbf{X}_1^h=\mathbf{W}_1^h\mathbf{X}(\mathbf{W}_1^h)^{\top}\) (not on \(\mathbf{X}\) directly), the inner congruence by \(\mathbf{U}\in\text{GL}(m)\) before compression does not translate to a uniform congruence on the \(\mathbf{X}_1^h\). In particular, the LC Fr\'{e}chet mean is defined by mapping each SPD \(\mathbf{X}_1^h\) to its Cholesky factor \(\lfloor\mathfrak{L}(\mathbf{X}_1^h)\rfloor+\log(\text{diag}(\mathfrak{L}(\mathbf{X}_1^h)))\), averaging in that triangular space, then mapping back. This construction is not affine-invariant and not even orthogonally equivariant in general. Therefore, the LC Fr\'{e}chet mean changes in a way that is not a simple transform of \(\mathbf{X}_2\).

The affine-invariant (AI) Fr\'{e}chet mean has congruence equivariance: \[\mathbb{E}_{\text{AI}}(\mathbf{Q}\mathbf{X}_1\mathbf{Q}^{\top}, \ldots, \mathbf{Q}\mathbf{X}_n\mathbf{Q}^{\top})=\mathbf{Q}\mathbb{E}_{\text{AI}}(\mathbf{X}_1, \ldots, \mathbf{X}_n)\mathbf{Q}^{\top}.\]
Therefore, the AI Fr\'{e}chet mean is equivariant under matrix congruence, only if there exists a single \(\mathbf{Q}\in\text{GL}(m_1)\) that \(\mathbf{W}_1^h\mathbf{U}=\mathbf{Q}\mathbf{W}_1^h\) for all \(1\leq h\leq H\).

\subsection{Isometry}
A diffeomorphism \(\psi: \mathcal{S}^m_+\rightarrow\mathcal{S}^m_+\) is an isometry if \(d_g(\psi(\mathbf{X}), \psi(\mathbf{Z}))=d_g(\mathbf{X}, \mathbf{Z})\) for all \(\mathbf{X}, \mathbf{Z}\in\mathcal{S}^m_+\). Here, \(d_g\)  is the geodesic distance induced by the LC or LE Riemannian metric. The isometries of \(\mathcal{S}^m_+\) form a group under composition, denoted by \(I(\mathcal{S}^m_+)\). Let \(F: \mathcal{S}^m_+\rightarrow\mathcal{S}^{m_1}_+\) be the function defined by $\mathbf{X}\mapsto\mathbb{E}_g(\mathbf{W}_1^1\mathbf{X}(\mathbf{W}_1^1)^{\top}, \ldots, \mathbf{W}_1^H\mathbf{X}(\mathbf{W}_1^H)^{\top})$, where \(\mathbb{E}_g\) is either the LC or the LE Fr\'{e}chet mean.  

Given any two points \(\mathbf{X}, \mathbf{Z}\in\mathcal{S}^m_+\), the LE geodesic between \(F(\mathbf{X})\) and \(F(\mathbf{Z})\) is
\begin{align*}
    d_{\text{LE}}(F(\mathbf{X}), F(\mathbf{Z}))&=\|\log(F(\mathbf{X}))- \log(F(\mathbf{Z}))\|_F\\
    &=\|\frac{1}{H}\sum_{h=1}^H\log(\mathbf{W}_1^h\mathbf{X}(\mathbf{W}_1^h)^{\top})- \frac{1}{H}\sum_{h=1}^H\log(\mathbf{W}_1^h\mathbf{Z}(\mathbf{W}_1^h)^{\top})\|_F,
\end{align*}
and the LE geodesic between \(F(\psi(\mathbf{X}))\) and \(F(\psi(\mathbf{Z}))\) is
\begin{equation*}
    d_{\text{LE}}(F(\psi(\mathbf{X})), F(\psi(\mathbf{Z})))=\|\frac{1}{H}\sum_{h=1}^H\log(\mathbf{W}_1^h\psi(\mathbf{X})(\mathbf{W}_1^h)^{\top})- \frac{1}{H}\sum_{h=1}^H\log(\mathbf{W}_1^h\psi(\mathbf{Z})(\mathbf{W}_1^h)^{\top})\|_F,
\end{equation*}
Any LE-isometry is exactly a Euclidean isometry in the log domain: \(\psi(\mathbf{X})=\exp(\mathcal{O}(\log(\mathbf{X}))+\mathbf{U})\), where \(\mathcal{O}: \text{Sym}(m)\rightarrow\text{Sym}(m)\) is linear orthogonal w.r.t. the Frobenius inner product, and \(\mathbf{U}\in\text{Sym}(m)\) is a fixed symmetric matrix (a translation in the log domain). The crucial observation is that for semi-orthogonal \(\mathbf{W}_1^h\),  \(\log(\mathbf{W}_1^h\psi(\mathbf{X})(\mathbf{W}_1^h)^{\top})\neq \mathbf{W}_1^h\log(\psi(\mathbf{X}))(\mathbf{W}_1^h)^{\top}\) in general. Hence the orthogonal action \(\mathcal{O}\) and translation \(\mathbf{U}\) in the log domain do not factor out through the \(\mathbf{W}_1^h(\cdot)(\mathbf{W}_1^h)^{\top}\)
compression, so the two distances above generally differ. Therefore, the map \(F\)  is not invariant (nor equivariant) to the action of the LE-isometry group.

Likewise, any LC-isometry is exactly a Euclidean isometry in the LC chart domain: \(\psi(\mathbf{X})=\Phi^{-1}(\mathbf{R}\Phi(\mathbf{X})+\mathbf{u})\), where \(\mathbf{R}\in O(m(m+1)/2)\), \(\mathbf{u}\in\mathbb{R}^{m(m+1)/2}\), and \(\Phi\) is the log-Cholesky coordinate map for the LC metric. In general, \(\mathfrak{L}(\mathbf{W}_1^h\psi(\mathbf{X})(\mathbf{W}_1^h)^{\top})\neq \mathbf{W}_1^h\mathfrak{L}(\psi(\mathbf{X}))\), even for orthogonal \(\mathbf{W}_1^h\).  In fact, the Cholesky factor \(\mathfrak{L}(\mathbf{W}_1^h\psi(\mathbf{X})(\mathbf{W}_1^h)^{\top})\) depends nonlinearly on both \(\psi(\mathbf{X})\) and \(\mathbf{W}_1^h\) via a re-triangularization step. Consequently, the linear orthogonal action \(\mathbf{R}\) and translation \(\mathbf{u}\) in the LC chart do not commute with the compression \(\mathbf{W}_1^h\psi(\mathbf{X})(\mathbf{W}_1^h)^{\top}\). Therefore, \(F\) is not invariant (nor equivariant) under the action of the LC-isometry group.

\section{Proof of Propositions \ref{mapping_Lipschitz_LE}-\ref{without_activation}}

\subsection{Proof of Proposition \ref{mapping_Lipschitz_LE}}
Under the LE metric, we have \(F(\mathbf{X})=\exp(\frac{1}{H}\sum_{h=1}^H\log(\mathbf{W}_1^h\mathbf{X}(\mathbf{W}_1^h)^{\top}))\), and 
\begin{align*}
    d_{\text{LE}}(F(\mathbf{X}), F(\mathbf{Z}))&=\|\frac{1}{H}\sum_{h=1}^H\left[\log(\mathbf{W}_1^h\mathbf{X}(\mathbf{W}_1^h)^{\top})- \log(\mathbf{W}_1^h\mathbf{Z}(\mathbf{W}_1^h)^{\top})\right]\|_F\\
    &\leq \frac{1}{H}\sum_{h=1}^H\|\log(\mathbf{W}_1^h\mathbf{X}(\mathbf{W}_1^h)^{\top})- \log(\mathbf{W}_1^h\mathbf{Z}(\mathbf{W}_1^h)^{\top})\|_F.
\end{align*}
Let \(\gamma(t)=\exp((1-t)\log(\mathbf{Z})+t\log(\mathbf{X}))\) be the LE geodesic from \(\mathbf{Z}\) to \(\mathbf{X}\), and define \(f_h(t)=\log(\mathbf{W}_1^h\gamma(t)(\mathbf{W}_1^h)^{\top})\). The function \(f_h(t)\) is absolutely continuous, and therefore we have
\[
\|\log(\mathbf{W}_1^h\mathbf{X}(\mathbf{W}_1^h)^{\top})- \log(\mathbf{W}_1^h\mathbf{Z}(\mathbf{W}_1^h)^{\top})\|_F=\|f_h(1)- f_h(0)\|_F=\|\int_0^1 f_h'(t)dt\|_F\leq \int_0^1 \|f_h'(t)\|_Fdt.
\]
We below prove that 
\begin{equation}\label{upper_bound_LE}
    \|f_h'(t)\|_F\leq \kappa(\gamma(t))\|\log(\mathbf{X})-\log(\mathbf{Z})\|_F=\kappa(\gamma(t))d_{\text{LE}}(\mathbf{X}, \mathbf{Z}),
\end{equation}
where \(\kappa(\gamma(t))= \lambda_{\text{max}}(\gamma(t))/\lambda_{\text{min}}(\gamma(t))\) is the condition number of \(\gamma(t)\). When \(\mathbf{X}, \mathbf{Z}\in\mathcal{X}_{\alpha, \beta}\), we have \(\gamma(t)\in\mathcal{X}_{\alpha, \beta}\) and hence \(\kappa(\gamma(t))\leq \beta/\alpha\). Then it directly follows from Eq. (\ref{upper_bound_LE}) that \(F\) is Lipschitz on \(\mathcal{X}_{\alpha, \beta}\) with constant \(L\leq \beta/\alpha\).

Write \(\mathbf{U}(t)=\log(\gamma(t))=(1-t)\log(\mathbf{Z})+t\log(\mathbf{X})\) and \(\mathbf{V}=\log(\mathbf{X})-\log(\mathbf{Z})\). We have \(\gamma'(t)=(D_{\mathbf{U}(t)}\exp)(\mathbf{V})\), and the chain rule gives
\[
f_h'(t)=(D_{\mathbf{W}_1^h\gamma(t)(\mathbf{W}_1^h)^{\top}}\log)\left(\mathbf{W}_1^h(D_{\mathbf{U}(t)}\exp)(\mathbf{V})(\mathbf{W}_1^h)^{\top}\right).
\]
The differential \(D_{\mathbf{X}}\log: \mathcal{T}_{\mathbf{X}}\mathcal{S}^m_+\mapsto\mathcal{T}_{\log(\mathbf{X})}\text{Sym}(m)\) of the matrix logarithm at \(\mathbf{X}\) has the integral form
\[
(D_{\mathbf{X}}\log)(\mathbf{K})=\int_0^\infty (\mathbf{X}+s\mathbf{I})^{-1}\mathbf{K}(\mathbf{X}+s\mathbf{I})^{-1}ds, ~~\mathbf{K}\in\mathcal{T}_{\mathbf{X}}\mathcal{S}^m_+,
\]
Taking Frobenius norms and using \(\|(\mathbf{X}+s\mathbf{I})^{-1}\|_2=\frac{1}{\lambda_{\text{min}}(\mathbf{X})+s}\), we have
\[
\|(D_{\mathbf{X}}\log)(\mathbf{K})\|_F\leq \int_0^\infty \|(\mathbf{X}+s\mathbf{I})^{-1}\|_2^2\|\mathbf{K}\|_Fds=\|\mathbf{K}\|_F\int_0^\infty \frac{ds}{(\lambda_{\text{min}}(\mathbf{X})+s)^2}=\frac{\|\mathbf{K}\|_F}{\lambda_{\text{min}}(\mathbf{X})}.
\]
Therefore, we have
\[
\|f_h'(t)\|_F\leq\frac{\|\mathbf{W}_1^h(D_{\mathbf{U}(t)}\exp)(\mathbf{V})(\mathbf{W}_1^h)^{\top}\|_F}{\lambda_{\text{min}}(\mathbf{W}_1^h\gamma(t)(\mathbf{W}_1^h)^{\top})}\leq \frac{\|(D_{\mathbf{U}(t)}\exp)(\mathbf{V})\|_F}{\lambda_{\text{min}}(\mathbf{W}_1^h\gamma(t)(\mathbf{W}_1^h)^{\top})},
\]
where we have utilized the property that  \(\|\mathbf{X}\mathbf{Y}\mathbf{Z}\|_F\leq \|\mathbf{X}\|_2\|\mathbf{Y}\|_F\|\mathbf{Z}\|_2\) for any matrices \(\{\mathbf{X}, \mathbf{Y}, \mathbf{Z}\}\), and  \(\|\mathbf{X}\|_2=1\) when \(\mathbf{X}\) has orthonormal rows. Moreover, because \(\mathbf{W}_1^h\gamma(t)(\mathbf{W}_1^h)^{\top}\) is a compression of \(\gamma(t)\) by an isometry, eigenvalues interlace \citep{Fan1957ImbeddingCF}: \(\lambda_{\text{min}}(\mathbf{W}_1^h\gamma(t)(\mathbf{W}_1^h)^{\top})\geq \lambda_{\text{min}}(\gamma(t))\).

The differential \(D_{\mathbf{U}}\exp: \mathcal{T}_{\mathbf{U}}\text{Sym}(m)\mapsto\mathcal{T}_{\exp(\mathbf{U})}\mathcal{S}^m_+\) of matrix exponential map at \(\mathbf{U}\) has the integral form
\[
(D_{\mathbf{U}}\exp)(\mathbf{V})=\int_0^1 \exp((1-s)\mathbf{U})\mathbf{V}\exp(s\mathbf{U}) ds, ~~\mathbf{V}\in\mathcal{T}_{\mathbf{X}}\mathcal{S}^m_+.
\]
Taking Frobenius norms, we have
\[
\|(D_{\mathbf{U}}\exp)(\mathbf{V})\|_F\leq\int_0^1 \|\exp((1-s)\mathbf{U})\|_2\|\mathbf{V}\|_F\|\exp(s\mathbf{U})\|_2 ds=\|\mathbf{V}\|_F\|\exp(\mathbf{U})\|_2.
\]
When \(\mathbf{U}=\mathbf{U}(t)\), we have \(\|\exp(\mathbf{U}(t))\|_2=\|\gamma(t)\|_2=\lambda_{\text{max}}(\gamma(t))\), and substituting \(\mathbf{V}\) with \(\log(\mathbf{X})-\log(\mathbf{Z})\), we have
\[
\|f_h'(t)\|_F\leq \frac{\lambda_{\text{max}}(\gamma(t))\|\log(\mathbf{X})-\log(\mathbf{Z})\|_F}{\lambda_{\text{min}}(\mathbf{W}_1^h\gamma(t)(\mathbf{W}_1^h)^{\top})}\leq \frac{\lambda_{\text{max}}(\gamma(t))}{\lambda_{\text{min}}(\gamma(t))}\|\log(\mathbf{X})-\log(\mathbf{Z})\|_F.
\]

\subsection{Proof of Proposition \ref{mapping_non-expansive_LC}}
The log-Cholesky coordinate map  \(\Phi: \mathcal{S}^m_+\rightarrow\mathbb{R}^{m(m+1)/2}\), \(\Phi(\mathbf{X})=(\text{vnz}(\mathfrak{L}(\mathbf{X})), \log(\text{vd}(\mathfrak{L}(\mathbf{X}))))\) is a global coordinate map, where \(\text{vnz}(\mathfrak{L}(\mathbf{X}))\) is the vector of the non-zero entries of \(\lfloor\mathfrak{L}(\mathbf{X})\rfloor\), and \(\text{vd}(\mathfrak{L}(\mathbf{X}))\) is the vector of the  diagonal entries of \(\mathfrak{L}(\mathbf{X})\). The LC metric makes \(\mathcal{S}^m_+\) flat via the chart map \(\Phi\), and the LC geodesic distance is simply the Euclidean norm in the log-Cholesky coordinate: \(d_{\text{LC}}(\mathbf{X}, \mathbf{Z})=\|\Phi(\mathbf{X})-\Phi(\mathbf{Z})\|_2\). For any \(\mathbf{X}, \mathbf{Z}\in\mathcal{S}^m_+\), we have
\begin{align*}
    d_{\text{LC}}(F(\mathbf{X}), F(\mathbf{Z}))&=\|\frac{1}{H}\sum_{h=1}^H\left[\Phi(\mathbf{W}_1^h\mathbf{X}(\mathbf{W}_1^h)^{\top})- \Phi(\mathbf{W}_1^h\mathbf{Z}(\mathbf{W}_1^h)^{\top})\right]\|_2\\
    &\leq \frac{1}{H}\sum_{h=1}^H\|\Phi(\mathbf{W}_1^h\mathbf{X}(\mathbf{W}_1^h)^{\top})- \Phi(\mathbf{W}_1^h\mathbf{Z}(\mathbf{W}_1^h)^{\top})\|_2.
\end{align*}
In the proof below, we drop the indices and write \(\mathbf{W}\) for \(\mathbf{W}_1^h\). We have \(\mathbf{W}\mathbf{X}\mathbf{W}^{\top}=(\mathbf{W}\mathfrak{L}(\mathbf{X}))(\mathbf{W}\mathfrak{L}(\mathbf{X}))^{\top}\), and a thin QR decomposition:
\((\mathbf{W}\mathfrak{L}(\mathbf{X}))^{\top}=\mathbf{Q}_{\mathbf{X}}\mathbf{R}_{\mathbf{X}},\) where \(\mathbf{Q}_{\mathbf{X}}\in\mathbb{R}^{m\times m_1}\) has orthonormal columns, and \(\mathbf{R}_{\mathbf{X}}\in\mathbb{R}^{m_1\times m_1}\) is upper-triangular with positive diagonal. The Cholesky factor of \(\mathbf{W}\mathbf{X}\mathbf{W}^{\top}\) is exactly \(\mathbf{R}_{\mathbf{X}}^{\top}\): \(\mathfrak{L}(\mathbf{W}\mathbf{X}\mathbf{W}^{\top})=\mathbf{R}_{\mathbf{X}}^{\top}\). Therefore, we have 
\[
   \|\Phi(\mathbf{W}\mathbf{X}\mathbf{W}^{\top})- \Phi(\mathbf{W}\mathbf{Z}\mathbf{W}^{\top})\|_2^2=\|\lfloor \mathbf{R}_{\mathbf{X}}^{\top}\rfloor- \lfloor \mathbf{R}_{\mathbf{Z}}^{\top}\rfloor\|_F^2+\|\log(\text{diag}( \mathbf{R}_{\mathbf{X}}^{\top}))-\log(\text{diag}( \mathbf{R}_{\mathbf{Z}}^{\top}))\|_F^2.
\]

Fix any spectral band \(0<\alpha\leq \beta<\infty\), and consider the subset \(\mathcal{X}_{\alpha, \beta}=\{\mathbf{X}\in\mathcal{S}^m_+: \alpha\mathbf{I}\preceq \mathbf{X}\preceq \beta\mathbf{I} \}\). Then we have
\begin{align*}
    \|\Phi(\mathbf{W}\mathbf{X}\mathbf{W}^{\top})- \Phi(\mathbf{W}\mathbf{Z}\mathbf{W}^{\top})\|_2^2&\leq \|\lfloor \mathbf{R}_{\mathbf{X}}^{\top}\rfloor- \lfloor \mathbf{R}_{\mathbf{Z}}^{\top}\rfloor\|_F^2+\alpha^{-1}\|\text{diag}( \mathbf{R}_{\mathbf{X}}^{\top})-\text{diag}( \mathbf{R}_{\mathbf{Z}}^{\top})\|_F^2\\
    &\leq \max\{1, \alpha^{-1}\}\|\mathbf{R}_{\mathbf{X}}^{\top}-\mathbf{R}_{\mathbf{Z}}^{\top}\|_F^2.
\end{align*}

Define a line in the \(m\times m\) matrix space: \(\mathbf{Y}(t)=(1-t)\mathbf{Z}+t\mathbf{X}\), for \(0\leq t\leq 1\), and let \(\mathbf{A}(t)=\mathbf{W}\mathbf{Y}(t)\mathbf{W}^{\top}\). It can be readily proved that \(\mathbf{A}(t)\in\mathcal{X}_{\alpha, \beta}\), for any \(0\leq t\leq 1\). The derivative in \(t\) is \(\dot{\mathbf{A}}(t)=\mathbf{W}(\mathbf{X}-\mathbf{Z})\mathbf{W}^{\top}\). 

Let \(\mathbf{L}(t)\) be the Cholesky factor of \(\mathbf{A}(t)\): \(\mathfrak{L}(\mathbf{A}(t))=\mathbf{L}(t)\). In particular, \(\mathbf{L}(0)=\mathbf{R}_{\mathbf{Z}}^{\top}\) and \(\mathbf{L}(1)=\mathbf{R}_{\mathbf{X}}^{\top}\). Differentiate the Cholesky map \(\mathbf{A}(t)=\mathbf{L}(t)\mathbf{L}(t)^{\top}\): \(\dot{\mathbf{A}}(t)=\dot{\mathbf{L}}(t)\mathbf{L}(t)^{\top}+\mathbf{L}(t)\dot{\mathbf{L}}(t)^{\top}\), and then we have
\[
\mathbf{L}(t)^{-1}\dot{\mathbf{A}}(t)\mathbf{L}(t)^{-\top}=\mathbf{L}(t)^{-1}\dot{\mathbf{L}}(t)+\dot{\mathbf{L}}(t)^{\top}\mathbf{L}(t)^{-\top}=\mathbf{\Omega}+\mathbf{\Omega}^{\top},
\]
where \(\mathbf{\Omega}=\mathbf{L}(t)^{-1}\dot{\mathbf{L}}(t)\) is lower-triangular. Therefore, we have \(\|\mathbf{\Omega}\|_F\leq \|\mathbf{\Omega}+\mathbf{\Omega}^{\top}\|_F\leq\| \mathbf{L}(t)^{-1}\|_2^2\|\dot{\mathbf{A}}(t)\|_F.\) Finally \(\dot{\mathbf{L}}(t)=\mathbf{L}(t)\mathbf{\Omega}\) gives
\[
\|\dot{\mathbf{L}}(t)\|_F\leq\|\mathbf{L}(t)\|_2\|\mathbf{\Omega}\|_F\leq \|\mathbf{L}(t)\|_2\| \mathbf{L}(t)^{-1}\|_2^2\|\dot{\mathbf{A}}(t)\|_F.
\]
The squared singular values of \(\mathbf{L}(t)\) are the eigen values of \(\mathbf{A}(t)\), and therefore \(\|\mathbf{L}(t)\|_2\leq \sqrt{\beta}\) and \(\| \mathbf{L}(t)^{-1}\|_2\leq \frac{1}{\sqrt{\alpha}}\). Finally, we have \(\|\dot{\mathbf{L}}(t)\|_F\leq\frac{\sqrt{\beta}}{\alpha}\|\dot{\mathbf{A}}(t)\|_F.\)

Integrating along the path, we have
\[
\|\mathbf{R}_{\mathbf{X}}^{\top}- \mathbf{R}_{\mathbf{Z}}^{\top}\|_F\leq \int_0^1 \|\dot{\mathbf{L}}(t)\|_F dt\leq\frac{\sqrt{\beta}}{\alpha} \int_0^1 \|\dot{\mathbf{A}}(t)\|_F dt=\frac{\sqrt{\beta}}{\alpha} \|\mathbf{W}(\mathbf{X}-\mathbf{Z})\mathbf{W}^{\top}\|_F.
\]
Given that \(\mathbf{W}\mathbf{W}^{\top}=\mathbf{I}\), we have \(\|\mathbf{W}(\mathbf{X}-\mathbf{Z})\mathbf{W}^{\top}\|_F\leq \|\mathbf{W}\|_2\|\mathbf{X}-\mathbf{Z}\|_F\|\mathbf{W}^{\top}\|_2=\|\mathbf{X}-\mathbf{Z}\|_F\). Write \(\mathbf{X}-\mathbf{Z}=\mathfrak{L}(\mathbf{X})(\mathfrak{L}(\mathbf{X})-\mathfrak{L}(\mathbf{Z}))^{\top}+(\mathfrak{L}(\mathbf{X})-\mathfrak{L}(\mathbf{Z}))\mathfrak{L}(\mathbf{Z})^{\top}\), we have
\[
\|\mathbf{X}-\mathbf{Z}\|_F\leq (\|\mathfrak{L}(\mathbf{X})\|_2+\|\mathfrak{L}(\mathbf{Z})\|_2)\|\mathfrak{L}(\mathbf{X})-\mathfrak{L}(\mathbf{Z})\|_F\leq 2\sqrt{\beta}\|\mathfrak{L}(\mathbf{X})-\mathfrak{L}(\mathbf{Z})\|_F.
\]
Applying the mean-value theorem on the diagonal entries, we have
\begin{align*}
|[\mathfrak{L}(\mathbf{X})]_{i,i}-[\mathfrak{L}(\mathbf{Z})]_{i,i}|&\leq \max\{[\mathfrak{L}(\mathbf{X})]_{i,i}, [\mathfrak{L}(\mathbf{Z})]_{i,i}\}|\log([\mathfrak{L}(\mathbf{X})]_{i,i})-\log([\mathfrak{L}(\mathbf{Z})]_{i,i})|\\
&\leq \sqrt{\beta}|\log([\mathfrak{L}(\mathbf{X})]_{i,i})-\log([\mathfrak{L}(\mathbf{Z})]_{i,i})|,
\end{align*}
and therefore
\begin{align*}
\|\mathfrak{L}(\mathbf{X})-\mathfrak{L}(\mathbf{Z})\|_F^2 &\leq \|\lfloor\mathfrak{L}(\mathbf{X})\rfloor-\lfloor\mathfrak{L}(\mathbf{Z})\rfloor\|_F^2+\beta\|\log(\text{diag}(\mathfrak{L}(\mathbf{X})))-\log(\text{diag}(\mathfrak{L}(\mathbf{Z})))\|_F^2\\
    &\leq \beta\|\Phi(\mathbf{X})- \Phi(\mathbf{Z})\|_2^2
\end{align*}
Putting everything together, we have 
\[
\|\Phi(\mathbf{W}\mathbf{X}\mathbf{W}^{\top})- \Phi(\mathbf{W}\mathbf{Z}\mathbf{W}^{\top})\|_2\leq \max\{1, \alpha^{-1}\}\frac{2\beta^{3/2}}{\alpha}\|\Phi(\mathbf{X})- \Phi(\mathbf{Z})\|_2,
\]
for any \(\mathbf{W}\) of orthonormal rows.

\subsection{Proof of Proposition \ref{without_activation}}
Let \(\phi\) be either the matrix logarithm map \(\log\) or the log-Cholesky coordinate map \(\Phi\). The unified formula for the Fr\'{e}chet mean in the Pooling layer is \[F(\mathbf{X}; \{\mathbf{W}_1^h\}_{h=1}^H)=\phi^{-1}(\frac{1}{H}\sum_{h=1}^H \phi(\mathbf{W}_1^h\mathbf{X}(\mathbf{W}_1^h)^{\top})).\] 
Write \(\mathbf{Z}=F(\mathbf{X}; \{\mathbf{W}_1^h\}_{h=1}^H)\). If the function \(F\) is ``linear'', then there exists another weight matrix \(\mathbf{W}\in\mathbb{R}^{m_1\times m}\) with \(\mathbf{W}\mathbf{W}^{\top}=\mathbf{I}\) such that \(\mathbf{Z}=F(\mathbf{X}; \{\mathbf{W}_1^h\}_{h=1}^H)=\mathbf{W}\mathbf{X}\mathbf{W}^{\top}\). 

Let \(\lambda_1(\mathbf{X})\geq ...\geq\lambda_m(\mathbf{X})\) be the eigenvalues of \(\mathbf{X}\), and \(\mu_1(\mathbf{Z})\geq ...\geq\mu_{m_1}(\mathbf{Z})\) be the eigenvalues of \(\mathbf{Z}\). According to \citet{Fan1957ImbeddingCF}, if and only if the eigenvalues of \(\mathbf{Z}\) satisfy the interlacing inequalities: \(\lambda_j(\mathbf{X})\geq\mu_j(\mathbf{Z})\geq\lambda_{j+(m-m_1)}(\mathbf{X})\),
for any \(j=1, \ldots, m_1\), then there exists an \(\mathbf{W}\in\mathbb{R}^{m_1\times m}\) with \(\mathbf{W}\mathbf{W}^{\top}=\mathbf{I}\) such that \(\mathbf{Z}=\mathbf{W}\mathbf{X}\mathbf{W}^{\top}\).

Under the LE metric, with the  eigenvalue decomposition \(\mathbf{W}_1^h\mathbf{X}(\mathbf{W}_1^h)^{\top}=\mathbf{Q}_1^h\mathbf{\Lambda}_1^h(\mathbf{Q}_1^h)^{\top}\), the Fr\'{e}chet mean is
\[
\mathbf{Z}=\exp(\frac{1}{H}\sum_{j=1}^H \mathbf{Q}_1^h\log(\mathbf{\Lambda}_1^h)(\mathbf{Q}_1^h)^{\top}).
\]
Note that, in general, the heads \(\{\mathbf{W}_1^h\mathbf{X}(\mathbf{W}_1^h)^{\top}: h=1, \ldots, H\}\) do not have a common orthonormal eigenbasis. Therefore, although each compression \(\mathbf{W}_1^h\mathbf{X}(\mathbf{W}_1^h)^{\top}\) individually interlaces with  \(\mathbf{X}\), the eigenvalues of the average \(\mathbf{Z}\) in general do not satisfy the interlacing inequalities.

The key point is that \(\Omega:=\{\log(\mathbf{W}\mathbf{X}\mathbf{W}^{\top}): \mathbf{W}\mathbf{W}^{\top}=\mathbf{I}\}\) is not a convex set for \(m_1\geq2\), and hence taking average of \(\{\log(\mathbf{W}_1^h\mathbf{X}(\mathbf{W}_1^h)^{\top}): h=1, \ldots, H\}\) can leave the set \(\Omega\). Therefore, in general there is no \(\mathbf{W}\) with \(\mathbf{W}\mathbf{W}^{\top}=\mathbf{I}\) such that \(\log(\mathbf{W}\mathbf{X}\mathbf{W}^{\top})=\frac{1}{H}\sum_{h=1}^H \log(\mathbf{W}_1^h\mathbf{X}(\mathbf{W}_1^h)^{\top}))\). Since \(\exp\) and \(\log\) are inverse diffeomorphisms between \(\mathcal{S}^{m_1}_+\) and Sym(\(m_1\)), this implies there is no such    \(\mathbf{W}\) with \(F(\mathbf{X}; \{\mathbf{W}_1^h\}_{h=1}^H)=\mathbf{W}\mathbf{X}\mathbf{W}^{\top}\). Consequently, the spectrum of the average  \(\frac{1}{H}\sum_{j=1}^H \mathbf{Q}_1^h\log(\mathbf{\Lambda}_1^h)(\mathbf{Q}_1^h)^{\top}\) need not satisfy the Fan-Pall interlacing inequalities relative to \(\mathbf{X}\).

An identical argument holds for the LC metric by replacing \(\log\) with \(\Phi\) and noting that  the set \(\{\Phi(\mathbf{W}\mathbf{X}\mathbf{W}^{\top}): \mathbf{W}\mathbf{W}^{\top}=\mathbf{I}\}\) is likewise non-convex.

\section{Backpropagation in the matrix-to-vector module}\label{matrix_partial_derivative}
We absorb each BiMap layer into its succeeding operation, yielding two composite maps \(f^{ap}(\cdot)\) and \(f^{log}(\cdot)\). In particular, \(f^{ap}(\cdot)\) directly maps  \(\mathbf{X}(t)\) to \(\mathbf{X}_3(t)\): 
\[\mathbf{X}_3(t)=f^{ap}(\mathbf{X}(t); \{\mathbf{W}_1^h\}_{h=1}^H)=\phi^{-1}(\frac{\alpha}{H}\sum_{h=1}^H \phi(\mathbf{W}_1^h\mathbf{X}(t)(\mathbf{W}_1^h)^{\top})),\] then \(f^{log}(\cdot)\) directly maps \(\mathbf{X}_3(t)\) to \(\mathbf{Y}^h(t)\): 
\[\mathbf{Y}^h(t)=f^{log}(\mathbf{X}_3(t); \mathbf{W}_4^h)=\text{Log}_{\mathbf{I}}^g(\mathbf{W}_4^h\mathbf{X}_3(t)(\mathbf{W}_4^h)^{\top}), ~~h=1, \ldots, H.\]

Note that both \(f^{ap}(\cdot)\) and \(f^{log}(\cdot)\) act pointwise in time: they are applied independently at each \(t\) to the matrix  \(\mathbf{X}(t)\) or  \(\mathbf{X}_3(t)\), rather than being functionals of the entire trajectory \(\tX\in\mathcal{H}_g([t_0, t_1]; \mathcal{S}^m_+)\). Consequently, the derivatives with respect to the inputs, namely \(\frac{\partial f^{ap}}{\partial \mathbf{X}(t)}\) and \(\frac{\partial f^{log}}{\partial \mathbf{X}_3(t)}\), are the usual matrix Jacobians evaluated at time \(t\) (as in the static, single-matrix case), rather than functional/variational derivatives with respect to the path.

However, the functional layer in the encoder \(\mathcal{E}\), namely Eq. \eqref{functional_layer_encoder}, is a functional of the input vector-valued function. We write \(\pmb{z}=\int_{t_0}^{t_1}\pmb{W}^{(1)}(t) \pmb{y}(t)dt+\pmb{b}\) and \(\pmb{x}^{(1)}=a(\pmb{z})\). For a perturbation \(\pmb{y}\rightarrow\pmb{y}+\epsilon \pmb{\eta}\), the first variation is
\[
\delta\pmb{z}=\int_{t_0}^{t_1}\pmb{W}^{(1)}(t) \pmb{\eta}(t)dt, ~~~\delta\pmb{x}^{(1)}=\text{diag}(a'(\pmb{z}))\delta\pmb{z}=\int_{t_0}^{t_1}[\text{diag}(a'(\pmb{z}))\pmb{W}^{(1)}(t)] \pmb{\eta}(t)dt,
\]
where \(\text{diag}(a'(\pmb{z}))\) is the Jacobian of the element-wise activation \(a\) at \(\pmb{z}\). Hence, the functional derivative density with respect to the whole function \(\pmb{y}\) is the matrix kernel \(\frac{\partial\pmb{x}^{(1)}}{\partial\pmb{y}}(s)=\text{diag}(a'(\pmb{z}))\pmb{W}^{(1)}(s)\in\mathbb{R}^{q_1\times p}\) for \(s\in[t_0, t_1]\), where \(q_1\) is the dimension of \(\pmb{x}^{(1)}\). 

Recall that \(\pmb{y}(t)^{\top}=[\text{vl}(\mathbf{Y}^1(t))^{\top}, \ldots, \text{vl}(\mathbf{Y}^H(t))^{\top}]\) and vl is an isometry. Now split the  matrix kernel into \(H\) blocks: \(\text{diag}(a'(\pmb{z}))\pmb{W}^{(1)}(s)=[\mathbf{K}_1(s), \ldots, \mathbf{K}_H(s)]\), where \(\mathbf{K}_h(s)\in\mathbb{R}^{q_1\times m_2(m_2+1)/2}\) and \(p=H\times m_2(m_2+1)/2\). Then the functional derivative density w.r.t. the whole \(\mathbf{Y}^h\) is
\[
\frac{\partial x^{(1)}_r}{\partial \mathbf{Y}^h}(s)=\text{vl}^{-1}([\mathbf{K}_h(s)]_{r,:}^{\top}), \text{ for } r=1, \ldots, q_1.
\]

\subsection{Logarithm layer}
For each head \(1\leq h\leq H\), we have \(\mathbf{Y}^h(t)=\text{Log}_{\mathbf{I}}^g(\mathbf{W}_4^h\mathbf{X}_3(t)(\mathbf{W}_4^h)^{\top})\), where \((\mathbf{W}_4^h)^{\top}\in\text{St}(m_1, m_2)\). To simplify notation, we drop the head index \(h\) and write \(\mathbf{X}_4(t)=\mathbf{W}_4\mathbf{X}_3(t)\mathbf{W}_4^{\top}\).

\paragraph{LE metric} When \(g\) is the LE metric, the Riemannian logarithm at the identity is exactly the matrix logarithm: \(\text{Log}_{\mathbf{I}}^g(\mathbf{X}_4(t))=\log(\mathbf{X}_4(t))\). The  Fr\'{e}chet  derivative of log at  \(\mathbf{X}_4(t)\) applied to a direction \(\mathbf{U}\) is
\[
(D_{\mathbf{X}_4(t)}\log)(\mathbf{U})=\int_0^\infty [\mathbf{X}_4(t)+s\mathbf{I}]^{-1}\mathbf{U}[\mathbf{X}_4(t)+s\mathbf{I}]^{-1}ds,
\]
which is self-adjoint w.r.t. the Frobenius inner product. If \(\mathbf{U}\) is symmetric (resp. skew), then \((D_{\mathbf{X}_4(t)}\log)(\mathbf{U})\) is symmetric (resp. skew). The spectral form for the Fr\'{e}chet  derivative \((D_{\mathbf{X}_4(t)}\log)(\mathbf{U})\) is given in Eq. \eqref{differential_derivative_log}. Then, by the chain rule \(D_{\mathbf{W}_4}\mathbf{Y}(t)=D_{\mathbf{X}_4(t)}\log\circ D_{\mathbf{W}_4}\mathbf{X}_4(t)\), the directional derivative with respect to \(\mathbf{W}_4\) in direction \(\Delta \mathbf{W}\) is
\[
(D_{\mathbf{W}_4}\mathbf{Y}(t))(\Delta \mathbf{W})=(D_{\mathbf{X}_4(t)}\log)(\Delta \mathbf{W}\mathbf{X}_3(t)\mathbf{W}_4^{\top}+\mathbf{W}_4\mathbf{X}_3(t)\Delta \mathbf{W}^{\top}).
\]

Let \(\mathbf{V}(t)\) be an ``upstream gradient'', and write \(f (t)=\langle\mathbf{V}(t), \mathbf{Y}(t) \rangle_F\). The linear mapping \((D_{\mathbf{X}_4(t)}\log)(\cdot)\) is self-adjoint, and therefore we have
\begin{align*}
   df(t)=\langle\mathbf{V}(t), d\mathbf{Y}(t) \rangle_F &=\langle\mathbf{V}(t), (D_{\mathbf{X}_4(t)}\log)(d\mathbf{X}_4(t)) \rangle_F\\
   &=\langle\mathbf{V}_s(t), (D_{\mathbf{X}_4(t)}\log)(d\mathbf{X}_4(t)) \rangle_F=\langle(D_{\mathbf{X}_4(t)}\log)(\mathbf{V}_s(t)), d\mathbf{X}_4(t) \rangle_F ,
\end{align*}
where \(d\mathbf{X}_4(t)=(d\mathbf{W}_4)\mathbf{X}_3(t)\mathbf{W}_4^{\top}+\mathbf{W}_4\mathbf{X}_3(t)(d\mathbf{W}_4)^{\top}\), and \(\mathbf{V}_s(t)=\frac{1}{2}(\mathbf{V}(t)+\mathbf{V}(t)^{\top})\). Then we obtain
\begin{align*}
   df(t)&=\text{tr}((D_{\mathbf{X}_4(t)}\log)(\mathbf{V}_s(t))(d\mathbf{W}_4)\mathbf{X}_3(t)\mathbf{W}_4^{\top})+\text{tr}((D_{\mathbf{X}_4(t)}\log)(\mathbf{V}_s(t))\mathbf{W}_4\mathbf{X}_3(t)(d\mathbf{W}_4)^{\top})\\
    &=\text{tr}(\mathbf{X}_3(t)\mathbf{W}_4^{\top}(D_{\mathbf{X}_4(t)}\log)(\mathbf{V}_s(t))(d\mathbf{W}_4))+\text{tr}(((D_{\mathbf{X}_4(t)}\log)(\mathbf{V}_s(t))\mathbf{W}_4\mathbf{X}_3(t))^{\top}(d\mathbf{W}_4))\\
    &=\langle2(D_{\mathbf{X}_4(t)}\log)(\mathbf{V}_s(t))\mathbf{W}_4\mathbf{X}_3(t), d\mathbf{W}_4\rangle_F .
\end{align*}
Then the \textbf{Euclidean} gradient w.r.t. \(\mathbf{W}_4\) is 
\begin{equation}
\nabla _{\mathbf{W}_4} f(t)=2(D_{\mathbf{X}_4(t)}\log)(\mathbf{V}_s(t))\mathbf{W}_4\mathbf{X}_3(t).
\end{equation}
Then we can project \(\nabla _{\mathbf{W}_4} f(t)\) onto the tangent space to get the Riemannian gradient.

When taking directional derivative with respect to \(\mathbf{X}_3(t)\), we have
\begin{align*}
df(t)=\langle\mathbf{V}(t), (D_{\mathbf{X}_4(t)}\log)(d\mathbf{X}_4(t)) \rangle_F&=\langle(D_{\mathbf{X}_4(t)}\log)(\mathbf{V}_s(t)), d\mathbf{X}_4(t) \rangle_F\\
&=\langle(D_{\mathbf{X}_4(t)}\log)(\mathbf{V}_s(t)), \mathbf{W}_4d\mathbf{X}_3(t)\mathbf{W}_4^{\top} \rangle_F\\
&=\langle\mathbf{W}_4^{\top}(D_{\mathbf{X}_4(t)}\log)(\mathbf{V}_s(t))\mathbf{W}_4, d\mathbf{X}_3(t) \rangle_F.
\end{align*}
The Euclidean gradient w.r.t. the input \(\mathbf{X}_3(t)\) is 
\begin{equation}
\nabla _{\mathbf{X}_3(t)} f(t)=\mathbf{W}_4^{\top}(D_{\mathbf{X}_4(t)}\log)(\mathbf{V}_s(t))\mathbf{W}_4.
\end{equation}

\paragraph{LC metric}  Let \(\mathbf{L}(t)=\mathcal{L}(\mathbf{X}_4(t))\) denote the Cholesky factor of \(\mathbf{X}_4(t)=\mathbf{W}_4\mathbf{X}_3(t)\mathbf{W}_4^{\top}\). Then \(\mathbf{Y}(t)=\text{Log}_{\mathbf{I}}^{\text{LC}}(\mathbf{X}_4(t))=2\log(\text{diag}(\mathbf{L}(t)))+\lfloor\mathbf{L}(t)\rfloor+\lfloor\mathbf{L}(t)\rfloor^{\top}\), and hence
\[
d\mathbf{Y}(t)=2\text{diag}(\frac{d L_{1,1}(t)}{L_{1,1}(t)}, \cdots, \frac{d L_{m_2,m_2}(t)}{L_{m_2,m_2}(t)})+\lfloor d\mathbf{L}(t)\rfloor+\lfloor d\mathbf{L}(t)\rfloor^{\top}.
\]
Let \(\mathbf{V}(t)\) be an ``upstream gradient'', and write \(f (t)=\langle\mathbf{V}(t), \mathbf{Y}(t) \rangle_F\). Then the differential is
\begin{align*}
       df(t)=\langle\mathbf{V}(t), d\mathbf{Y}(t) \rangle_F &= 2\sum_{i=1}^{m_2}V_{i,i}(t)\frac{d L_{i,i}(t)}{L_{i,i}(t)}+\sum_{i>j}[V_{i,j}(t)+V_{j,i}(t)]d L_{i,j}(t)\\
       &= \langle\mathbf{H}(t), d\mathbf{L}(t) \rangle_F,
\end{align*}
where \(\mathbf{H}(t)\) is a lower-triangular matrix: \(H_{i,i}(t)=\frac{2V_{i,i}(t)}{L_{i,i}(t)}\), \(H_{i, j}(t)=V_{i,j}(t)+V_{j,i}(t)\) (\(i>j\)), and \(H_{i, j}(t)=0\) (\(i<j\)).

The differential of the Cholesky factor satisfies \(d\mathbf{L}(t)=\mathbf{L}(t)\Psi(\mathbf{L}(t)^{-1}d\mathbf{X}_4(t)\mathbf{L}(t)^{-\top})\), where \(\Psi\) is the lower-symmetrizer (keep strictly lower part and take half of the diagonal; zero upper part). Since  \(\Psi\) is self-adjoint, the adjoint mapping gives 
\begin{align*}
    \langle\mathbf{H}(t), d\mathbf{L}(t) \rangle_F&=\langle\mathbf{L}(t)^{\top}\mathbf{H}(t), \Psi(\mathbf{L}(t)^{-1}d\mathbf{X}_4(t)\mathbf{L}(t)^{-\top})\rangle_F\\
    &=\langle\Psi(\mathbf{L}(t)^{\top}\mathbf{H}(t)), \mathbf{L}(t)^{-1}d\mathbf{X}_4(t)\mathbf{L}(t)^{-\top}\rangle_F\\
    &=\langle\mathbf{L}(t)^{-\top}\Psi(\mathbf{L}(t)^{\top}\mathbf{H}(t))^{\top}\mathbf{L}(t)^{-1}, d\mathbf{X}_4(t)\rangle_F
\end{align*}
Because \(d\mathbf{X}_4(t)\) is symmetric, we have \(\langle\mathbf{H}(t), d\mathbf{L}(t) \rangle_F=\langle\mathbf{U}(t), d\mathbf{X}_4(t) \rangle_F\), where \(\mathbf{U}(t)=\frac{1}{2}\mathbf{L}(t)^{-\top}(\Psi(\mathbf{L}(t)^{\top}\mathbf{H}(t))+\Psi(\mathbf{L}(t)^{\top}\mathbf{H}(t))^{\top})\mathbf{L}(t)^{-1}\).

When taking directional derivative with respect to \(\mathbf{W}_4\), we have
\begin{align*}
df(t)=\langle\mathbf{U}(t), d\mathbf{X}_4(t) \rangle_F &=\langle\mathbf{U}(t), (d\mathbf{W}_4)\mathbf{X}_3(t)\mathbf{W}_4^{\top}+\mathbf{W}_4\mathbf{X}_3(t)(d\mathbf{W}_4)^{\top} \rangle_F\\
&=\langle2\mathbf{U}(t)\mathbf{W}_4\mathbf{X}_3(t), d\mathbf{W}_4 \rangle_F .
\end{align*}
Therefore, the \textbf{Euclidean} gradient w.r.t. \(\mathbf{W}_4\) is 
\begin{equation}
    \nabla _{\mathbf{W}_4} f(t)=2\mathbf{U}(t)\mathbf{W}_4\mathbf{X}_3(t).
\end{equation}
When taking directional derivative with respect to \(\mathbf{X}_3(t)\), we have
\begin{align*}
df(t)=\langle\mathbf{U}(t), d\mathbf{X}_4(t) \rangle_F =\langle\mathbf{U}(t), \mathbf{W}_4d\mathbf{X}_3(t)\mathbf{W}_4^{\top} \rangle_F=\langle\mathbf{W}_4^{\top}\mathbf{U}(t)\mathbf{W}_4, d\mathbf{X}_3(t) \rangle_F.
\end{align*}
Therefore, the Euclidean gradient w.r.t. \(\mathbf{X}_3(t)\) is 
\begin{equation}
    \nabla _{\mathbf{X}_3(t)} f(t)=\mathbf{W}_4^{\top}\mathbf{U}(t)\mathbf{W}_4.
\end{equation}

\subsection{Activated-pooling layer}
For the geodesic-shrinkage activation, we have
\[\mathbf{X}_3(t)=\phi^{-1}(\frac{\alpha}{H}\sum_{h=1}^H \phi(\mathbf{W}_1^h\mathbf{X}(t)(\mathbf{W}_1^h)^{\top})),\] 
where \(\phi\) is the principal matrix logarithm log for the LE metric, and the log-Cholesky coordinate map \(\Phi\) for the LC metric. Let \(\mathbf{A}(t)=\frac{\alpha}{H}\sum_{h=1}^H \phi(\mathbf{W}_1^h\mathbf{X}(t)(\mathbf{W}_1^h)^{\top})\) and \(\mathbf{X}_1^h(t)=\mathbf{W}_1^h\mathbf{X}(t)(\mathbf{W}_1^h)^{\top}\). By the chain rule, for any \(h\), we have
\[
(D_{\mathbf{W}_1^h}\mathbf{X}_3(t))(\Delta \mathbf{W})=(D_{\mathbf{A}(t)}\phi^{-1})(\frac{\alpha}{H}(D_{\mathbf{X}_1^h(t)}\phi)(\Delta \mathbf{W}\mathbf{X}(t)(\mathbf{W}_1^h)^{\top}+\mathbf{W}_1^h\mathbf{X}(t)\Delta \mathbf{W}^{\top})).
\]

\paragraph{LE metric} Under the LE metric, we have \[(D_{\mathbf{W}_1^h}\mathbf{X}_3(t))(\Delta \mathbf{W})=(D_{\mathbf{A}(t)}\exp)(\frac{\alpha}{H}(D_{\mathbf{X}_1^h(t)}\log)(\Delta \mathbf{W}\mathbf{X}(t)(\mathbf{W}_1^h)^{\top}+\mathbf{W}_1^h\mathbf{X}(t)\Delta \mathbf{W}^{\top})).\]  The spectral form for the Fr\'{e}chet  derivative \((D_{\mathbf{X}}\log)(\mathbf{U})\) is given in Eq. \eqref{differential_derivative_log}, and for \((D_{\mathbf{U}}\exp)(\mathbf{K})\) given in Eq. \eqref{differential_derivative_exp}. 

Let \(\mathbf{V}(t)\) be an ``upstream gradient'', and write \(f (t)=\langle\mathbf{V}(t), \mathbf{X}_3(t) \rangle_F\). The mapping  \((D_{\mathbf{U}}\exp)(\cdot)\) is self-adjoint, and therefore we have 
\[
df(t)=\langle\mathbf{V}(t), d\mathbf{X}_3(t) \rangle_F=\langle\mathbf{V}(t), (D_{\mathbf{A}(t)}\exp)(d\mathbf{A}(t)) \rangle_F=\langle(D_{\mathbf{A}(t)}\exp)(\mathbf{V}_s(t)), d\mathbf{A}(t) \rangle_F,
\]
where again \(\mathbf{V}_s(t)=\frac{1}{2}(\mathbf{V}(t)+\mathbf{V}(t)^{\top})\). Then we obtain
\begin{align*}
df(t)=\langle(D_{\mathbf{A}(t)}\exp)(\mathbf{V}_s(t)),\frac{\alpha}{H}(D_{\mathbf{X}_1^h(t)}\log)(d\mathbf{X}_1^h) \rangle_F=\frac{\alpha}{H}\langle(D_{\mathbf{X}_1^h(t)}\log)((D_{\mathbf{A}(t)}\exp)(\mathbf{V}_s(t))),d\mathbf{X}_1^h \rangle_F.
\end{align*}
Finally, we obtain \(df(t)=\langle\frac{2\alpha}{H}(D_{\mathbf{X}_1^h(t)}\log)((D_{\mathbf{A}(t)}\exp)(\mathbf{V}_s(t)))\mathbf{W}_1^h\mathbf{X}(t), d\mathbf{W}_1^h\rangle_F\), and therefore the \textbf{Euclidean} gradient w.r.t. \(\mathbf{W}_1^h\) is 
\begin{equation}
    \nabla _{\mathbf{W}_1^h} f(t)=\frac{2\alpha}{H}(D_{\mathbf{X}_1^h(t)}\log)((D_{\mathbf{A}(t)}\exp)(\mathbf{V}_s(t)))\mathbf{W}_1^h\mathbf{X}(t).
\end{equation}

\paragraph{LC metric} Under the LC metric, for a perturbation \(\Delta \mathbf{W}\) in \(\mathbf{W}_1^h\), we have \(d\mathbf{X}_1^h(t)=\Delta \mathbf{W}\mathbf{X}(t)(\mathbf{W}_1^h)^{\top}+\mathbf{W}_1^h\mathbf{X}(t)\Delta \mathbf{W}^{\top}\), \(d\mathbf{A}(t)=\frac{\alpha}{H}\sum_{h=1}^H(D_{\mathbf{X}_1^h(t)}\Phi)(d\mathbf{X}_1^h(t))\) and \(d\mathbf{X}_3(t)=(D_{\mathbf{A}(t)}\Phi^{-1})(d\mathbf{A}(t))\).
Let \(\mathbf{V}(t)\) be an ``upstream gradient'', and write \(f (t)=\langle\mathbf{V}(t), \mathbf{X}_3(t) \rangle_F\). Let \((D_{\mathbf{A}(t)}\Phi^{-1})^*\) be the adjoint operator of \((D_{\mathbf{A}(t)}\Phi^{-1})\), and \((D_{\mathbf{X}_1^h(t)}\Phi)^*\) the adjoint operator of \((D_{\mathbf{X}_1^h(t)}\Phi)\). Then we have 
\begin{align*}
df(t)&=\frac{\alpha}{H}\langle(D_{\mathbf{X}_1^h(t)}\Phi)^*((D_{\mathbf{A}(t)}\Phi^{-1})^*(\mathbf{V}_s(t))),d\mathbf{X}_1^h \rangle_F\\
&=\langle\frac{2\alpha}{H}(D_{\mathbf{X}_1^h(t)}\Phi)^*((D_{\mathbf{A}(t)}\Phi^{-1})^*(\mathbf{V}_s(t)))\mathbf{W}_1^h\mathbf{X}(t), d\mathbf{W}_1^h\rangle_F,
\end{align*}
where \(\mathbf{V}_s(t)=\frac{1}{2}(\mathbf{V}(t)+\mathbf{V}(t)^{\top})\). Then the \textbf{Euclidean} gradient w.r.t. \(\mathbf{W}_1^h\) is 
\begin{equation}
    \nabla _{\mathbf{W}_1^h} f(t)=\frac{2\alpha}{H}(D_{\mathbf{X}_1^h(t)}\Phi)^*((D_{\mathbf{A}(t)}\Phi^{-1})^*(\mathbf{V}_s(t)))\mathbf{W}_1^h\mathbf{X}(t).
\end{equation} 
We below give the explicit adjoints \((D_{\mathbf{A}(t)}\Phi^{-1})^*\) and \((D_{\mathbf{X}_1^h(t)}\Phi)^*\) for the log-Cholesky coordinates.

Let \(\mathbf{L}(t)=\mathcal{L}(\mathbf{X}_3(t))\) denote the Cholesky factor of \(\mathbf{X}_3(t)\). Define \(\mathbf{B}(t)=\mathbf{L}(t)^{\top}\mathbf{V}_s(t)\); The adjoint \((D_{\mathbf{A}(t)}\Phi^{-1})^*(\mathbf{V}_s(t))\in\mathbb{R}^{m_1(m_1+1)/2}\) is the vector of the lower-diagonal entries of the following matrix:
\[
2\text{diag}(\mathbf{L}(t))\text{diag}(\mathbf{B}(t))+\lfloor\mathbf{B}(t)\rfloor+\lfloor\mathbf{B}(t)^{\top}\rfloor.
\]

Let \(\mathbf{L}^h(t)=\mathcal{L}(\mathbf{X}_1^h(t))\) denote the Cholesky factor of \(\mathbf{X}_1^h(t)\) for \(h=1, \ldots, H\). Given a coordinate covector \(\pmb{g}(t)\in\mathbb{R}^{m_1(m_1+1)/2}\), build a lower-triangular \(\mathbf{H}^h(t)\) by
\[
[\mathbf{H}^h(t)]_{i,i}=\frac{g_{\kappa(i,i)}(t)}{[\mathbf{L}^h(t)]_{i,i}}, ~~[\mathbf{H}^h(t)]_{i,j}=g_{\kappa(i,j)}(t) ~(i>j),~~[\mathbf{H}^h(t)]_{i,j}=0 ~(i<j),
\]
where \(\kappa\) is the indexing function. Let \(\Psi\) be the lower-symmetrizer. Then
\[
(D_{\mathbf{X}_1^h(t)}\Phi)^*(\pmb{g}(t))=\frac{1}{2}\mathbf{L}^h(t)^{-\top}(\Psi(\mathbf{L}^h(t)^{\top}\mathbf{H}^h(t))+\Psi(\mathbf{L}^h(t)^{\top}\mathbf{H}^h(t))^{\top})\mathbf{L}^h(t)^{-1}.
\]

\section{Congruence matrix regularization}\label{Congruence matrix regularization}
Without additional regularization on the congruence weight matrices \(\{\mathbf{W}_1^h\}_{h=1}^H\) and \(\{\mathbf{W}_4^h\}_{h=1}^H\), different heads are likely to focus on the same dominant patterns in the functional data. To encourage the heads to learn complementary features, we introduce a row-space decorrelation penalty. 

For the matrices \(\{\mathbf{W}_1^h\}_{h=1}^H\), a natural choice is to penalize the overlap between their row spaces. Specifically, for each pair \(1 \le h < k \le H\), we consider
\[
\|\mathbf{W}_1^h(\mathbf{W}_1^k)^{\top}\|_F^2,
\]
which quantifies the degree of alignment between the row spaces of \(\mathbf{W}_1^h\) and \(\mathbf{W}_1^k\).  Summing over all pairs gives the total pairwise-overlap penalty:
\[
\mathcal{L}_{\mathrm{pair}}
=
\sum_{1\le h<k\le H}
\|\mathbf{W}_1^h(\mathbf{W}_1^k)^{\top}\|_F^2.
\]
To reduce the computational cost, we exploit the fact that each matrix \(\mathbf{W}_1^h\) is constrained to be full row rank with orthonormal rows, that is, \(\mathbf{W}_1^h(\mathbf{W}_1^h)^{\top} = \mathbf{I}_{m_1}.\)
For each head \(h\), we define
\[
\mathbf{P}_h = (\mathbf{W}_1^h)^{\top} \mathbf{W}_1^h \in \mathbb{R}^{m\times m},
\]
which is the orthogonal projector onto the row space of \(\mathbf{W}_1^h\).  Using \(\|\mathbf{W}_1^h(\mathbf{W}_1^k)^{\top}\|_F^2 = \operatorname{tr}(\mathbf{P}_h \mathbf{P}_k)\), we can rewrite the total pairwise-overlap penalty as
\[
\mathcal{L}_{\mathrm{pair}} = \sum_{1\le h<k\le H}\operatorname{tr}(\mathbf{P}_h\mathbf{P}_k)
= \frac{1}{2} \left( \| \sum_{h=1}^H \mathbf{P}_h \|_F^2 - \sum_{h=1}^H \|\mathbf{P}_h\|_F^2 \right).
\]
Moreover, for each projector \(\mathbf{P}_h\),
\[
\|\mathbf{P}_h\|_F^2 = \operatorname{tr}(\mathbf{P}_h^2) = \operatorname{tr}(\mathbf{P}_h) = m_1.
\]
Hence,
\[
\mathcal{L}_{\mathrm{pair}} = \frac{1}{2} \left(\| \sum_{h=1}^H \mathbf{P}_h \|_F^2 - H m_1 \right).
\]
Therefore, minimizing \(\mathcal{L}_{\mathrm{pair}}\) is equivalent to minimizing \(\| \sum_{h=1}^H \mathbf{P}_h \|_F^2\). 

When \(H m_1 > m\), the row spaces of all heads cannot be mutually orthogonal. In this overcomplete regime, a more appropriate objective is to distribute the row spaces as uniformly as possible, which corresponds to encouraging the sum of the projection matrices to be close to an isotropic target:
\[
\sum_{h=1}^H \mathbf{P}_h \approx \frac{H m_1}{m}\mathbf{I}_m.
\]
Accordingly, we define the following diversity regularizer:
\[
\mathcal{L}_{\mathrm{div}} = \lambda
\left\|\sum_{h=1}^H (\mathbf{W}_1^h)^{\top}\mathbf{W}_1^h - \frac{H m_1}{m}\mathbf{I}_m \right\|_F^2,
\]
where \(\lambda>0\) is a tuning parameter. The same regularization is also applied to \(\{\mathbf{W}_4^h\}_{h=1}^H\):
\[
\mathcal{L}_{\mathrm{div}} = \lambda
\left\|\sum_{h=1}^H (\mathbf{W}_4^h)^{\top}\mathbf{W}_4^h - \frac{H m_2}{m_1}\mathbf{I}_{m_1} \right\|_F^2.
\]

\section{Supplementary experimental details and results}\label{Supplementary experimental}

\subsection{Details on the clustering benchmark methods and their configuration}\label{Supplementary_algorithm}
\begin{description}
    \item[TSRVF] In \citep{8786184}, trajectory dissimilarity is measured in the Transported Square-Root Vector Field (TSRVF) space, with temporal reparameterization used to factor out inter-subject differences in execution rate. The authors adopted a particular Riemannian metric on the SPD manifold, which provides closed-form expressions for the key geometric operations they need (e.g., parallel transport). When the matrix dimension $m$ is large (e.g., $m > 100$), efficient comparison of covariance trajectories becomes computationally impractical. Therefore, \citet{8786184} apply dimension reduction before trajectory comparison. We adopt the same approach here and set the reduced dimension to 8. The other hyperparameter is the number of optimization iterations which is set to 30.
    
    \item[GeoAtt] The network feeds each SPD matrix $\mathbf{X}(t)$ into a manifold-aware CNN to produce a lower-dimensional SPD matrix $\mathbf{Z}(t)$, and the sequence $\{\mathbf{Z}(t)\}$ is then processed by a manifold-aware GRU. We take the hidden state from the GRU as the latent embedding of the whole trajectory $\{\mathbf{X}(t)\}$. The hyperparameters and their values are:   epochs${}=100$, latent dimension${}=8$, Adam learning rate${}=0.001$ (Euclidean parameters), Stiefel manifold learning rate${}=0.005$ (BiMap weights), and batch size${}=16$.
    
    \item[SPDNet] SPDNet learns a latent vector embedding \(\mathbf{x}(t)\) for each input SPD matrix \(\mathbf{X}(t)\). Given a sequence of SPD matrices \(\{\mathbf{X}(t_1), \mathbf{X}(t_2), \mathbf{X}(t_3), \ldots\}\), we average the corresponding latent embeddings \(\{\mathbf{x}(t_1), \mathbf{x}(t_2), \mathbf{x}(t_3), \ldots\}\), and use this mean vector as the latent representation of the entire SPD matrix-valued function. The hyperparameter settings are as follows:   training epochs${}=100$, latent dimension${}=8$, Adam learning rate${}=0.001$ (Euclidean parameters), Stiefel manifold learning rate${}=0.005$ (BiMap weights), and batch size${}=16$. The intermediate BiMap layer dimensions are determined automatically by $m_1 = \max(4, \lfloor m/2\rfloor)$ and $m_2 = \max(3, \lfloor m_1/2\rfloor)$.
    
    \item[PGA] The authors in \citep{1318725} introduced principal geodesic analysis (PGA), which is essentially a nonlinear generalization of PCA to manifold-valued data. In practice, PGA is implemented by mapping the manifold-valued data to the tangent space at the intrinsic mean via the Logarithm map and then performing standard PCA on the resulting tangent vectors. In our approach, we perform standard functional PCA in the tangent space, and then apply k-means on the principal component score vectors. We adopt the log-Euclidean metric and retain 10 principal components.
    
    \item[PDM] We compute the pairwise distances between functions using the affine-invariant metric and then apply k-medoids on the resulting distance matrix. The number of random initializations for the clustering method is 10.
    
    \item[DC] DC is the  standard MLP autoencoder for deep clustering (DC). To adapt it to functional data, we make the input layer contain \(m(m+1)/2\) nodes, where each node receives a T-dimensional input vector (corresponding to one $X_{ij}(t)$ observed over T time points). The hyperparameter settings are as follows: training epochs${}=100$, latent dimension${}=8$, Adam learning rate${}=0.001$, and batch size${}=16$. The encoder consists of 1D convolutional layers with channel sizes $\mathrm{tri} \to 64 \to 32$, followed by a fully connected layer that produces the latent representation.
    
    \item[MatFAE] The algorthm repository documentation provides a clear explanation of all hyperparameters and their default values.
\end{description}

For all four deep learning models, including GeoAtt, SPDNet, DC, and MatFAE, we used Bayesian optimization implemented in Optuna for hyperparameter tuning. In all the experiments below, MatFAE training and competitive methods computations were performed on a 2.5 GHz Intel Xeon Gold 6548Y+ CPU (32 cores) and an NVIDIA L40S GPU with 46 GB of VRAM.

\subsection{Details on the classification benchmark methods and their configuration}\label{classfication_method_config}
Shared protocol for all SPD-native methods (SPD-SRU, BW-norm, MatFAE-C): internal 80/20 split with early stopping on validation balanced accuracy (patience 25), class-weighted cross-entropy loss, Adam optimizer on Euclidean weights combined with a manual Stiefel-manifold step (learning rate $5\times 10^{-3}$) on BiMap/orthogonal weights, gradient clipping at 5.0, and an SPD floor of $\varepsilon = 10^{-4}$. STAGIN, which does not operate on the SPD manifold, uses the same early-stopping and loss protocol but standard Euclidean Adam throughout (no Stiefel step or SPD floor). All architecture designs and default hyperparameters (learning rate, weight decay, batch size, dropout, epoch budget) otherwise follow each method's original publication, as detailed below.
\begin{description}
    \item[SPD-SRU] Each frame is reduced to an 8$\times$8 SPD matrix via BiMap, then passed through one SPD-SRU recurrent cell with five time scales $\alpha=[0.01,0.25,0.5,0.9,0.99]$ (multi-scale log-Euclidean-weighted running SPD means, orthogonal recurrence, learnable scale/output/blend gates), followed by LogEig$\to$vech$\to$head. Batch size 40, weight decay 0, $\le$200 epochs (early-stopped), dropout 0.1.
    \item[STAGIN] Per-frame correlation graphs are thresholded to a top-30$\%$ binary adjacency and passed through a 4-layer GIN (sum aggregation, hidden dimension 128) with per-layer SERO readouts summed, sinusoidal temporal positional encoding, one-head temporal self-attention, and a summed CLS token over time; a linear head produces the final prediction. Learning rate $5\times 10^{-4}$, weight decay $10^{-5}$, batch size 3, 100 epochs, dropout 0.5. 
    \item[BW-norm] Standard SPDNet-style BiMap layer $\to$ BW batch normalization (running BW barycenter, momentum 0.1, $\varepsilon = 10^{-5}$, geodesic running mean; centering in the BW log-chart, scaling by BW variance, with a learnable SPD bias) $\to$ ReEig $\to$ LogEig $\to$ linear head; per-frame outputs are averaged before the final head. Learning rate $2.5\times 10^{-3}$, weight decay $5\times 10^{-2}$, batch size 30, $\le$200 epochs, dropout 0.1.
    \item[MatFAE-C] We retain the \pkg{MatFAE} encoder backbone, remove the decoder, and append a linear classification head to the latent representation \(\pmb{x}=\mathcal{E}(\mathcal{F}(\mathbf{X}))\). The objective combines class-weighted cross-entropy with the same regularization terms used in the unsupervised model, namely congruence matrix regularization, the orthogonality penalty on functional weights, and the roughness penalty on functional weights; class weights are computed within each training fold using \texttt{sklearn.utils.class\_weight.compute\_class\_weight}. Evaluation uses a class-stratified outer five-fold split. Euclidean parameters use Adam with learning rate \(10^{-3}\) and weight decay \(10^{-5}\), batch size \(16\), latent dimension \(16\), and dropout \(0.1\) before the classification head. Unless otherwise stated, all backbone hyperparameters \((H, m_1, m_2, p_1, p_2, \alpha, \alpha_{\min}, \mathrm{basis~}K)\), encoder and MLP widths, and loss weights are set to their default values. %The folds are not site-stratified and no ComBat harmonisation is applied; therefore, the reported results should be interpreted as within-distribution classification performance rather than evidence of out-of-site generalisation.
\end{description}

\subsection{Details on the simulation scenarios and additional experimental results}\label{Supplementary_synthetic}
Each synthetic dataset consists of 100 functional trajectories, each defined on the manifold \(\mathcal{S}^{48}_+\) . The matrix dimension is determined by the 48 cortical regions in the Harvard-Oxford atlas. Below, let \(q\) denote the number of time points at which the function \(\mathbf{X}(t)\) is evaluated over the interval  \([t_0, t_1]\), and $k$ the true number of clusters.

The benchmark is organized as an ablation ladder, where each synthetic dataset isolates a single type of structural signal. Consequently, failure on a given rung suggests that the method does not possess the feature-extraction mechanism required to capture that specific structure.

\begin{description}
    \item[A \((q=40, k=2)\)]: The data generation process is as follows. First, we define two well-separated SPD states, A and B. For Group 0, each trajectory starts in state A and switches to state B halfway through, producing an A\(\to\)B pattern. For Group 1, each trajectory starts in state B and switches to state A halfway through, producing a B\(\to\)A pattern. To introduce between-subject variability, we add random jitter of up to \(\pm 3\) windows to the transition point, together with tangent-space noise.  \emph{Rung: temporal ordering.}
    \item[B \((q=40, k=2)\)]: We define three SPD states, A, B, and C. Every trajectory begins and ends in state A. The two groups differ in the intermediate state they visit: Group 0 passes through state B, while Group 1 passes through state C. \emph{Rung: static identity.}
    \item[C \((q=120, k=3)\)]: We follow the HMM-based dFC construction of \citep{doi:10.1073/pnas.1705120114}. In particular, we consider three brain states and assign each group a different Markov transition matrix, such that the corresponding stationary dwell-time ratios are 75/15/10, 50/30/20, and 25/35/40, respectively. \emph{Rung: dwell time.}
    \item[D \((q=60, k=2)\)]: Both groups switch between the same two states, A and B, following piecewise-constant trajectories. Group 0 has a switching period of approximately \(q/4\), while Group 1 has a shorter switching period of approximately \(q/12\). As a result, both groups share the same visited-state set and stationary proportions, but differ in how frequently they transition between states. \emph{Rung: transition frequency.}
    \item[E \((q=30, k=2)\)]: Each trajectory is generated by sampling a symmetric-matrix-valued Gaussian process in the tangent space with squared-exponential kernel
    \[
    \kappa_\ell(t,t')=\exp\!\left(-\frac{(t-t')^2}{2\ell^2}\right),
    \]
    and then mapping it to the SPD manifold through \(\mathbf{X}(t)=\exp(\mathbf{Y}(t))\). Group 0 uses \(\ell=12\), while Group 1 uses \(\ell=2\). The cluster means are asymptotically the same, so the groups are distinguished only by trajectory smoothness. \emph{Rung: smoothness.}
    \item[F \((q=20, k=3)\)] For each time point, we sample \(\mathbf{X}(t)\sim \mathcal{W}_m(\mathbf{V}, v_c)/v_c\), with common scale matrix \(\mathbf{V}=\mathbf{I}\) and group-specific degrees of freedom \(v_c\in\{150, 75, 50\}\). Because the mean is the same for all groups, the groups differ only by their concentration, as reflected in quantities such as the log-determinant and trace. \emph{Rung: Wishart concentration.}
    \item[G \((q=30, k=2)\)]: We generate group-specific neural source signals and convolve them with the canonical double-gamma hemodynamic response function (HRF) \citep{GLOVER1999416}. We then construct the functional connectivity trajectory \(\mathbf{X}(t)\) by applying sliding-window Pearson correlation to the simulated signals \citep{10.1093/cercor/bhs352}. \emph{Rung: real HRF-blurred functional connectivity.}
    \item[H \((q=30, k=2)\)]: Following \citep{doi:10.1137/050637996, Pennec200641}, we consider two groups that share the same endpoints, \(\mathbf{X}_0\) and \(\mathbf{X}_1\). The geodesic connecting these two points on the SPD manifold is given by
    \[ \gamma(t)=\mathbf{X}_0^{1/2} \exp(t\log(\mathbf{X}_0^{-1/2}\mathbf{X}_1\mathbf{X}_0^{-1/2})) \mathbf{X}_0^{1/2}. \]
    For Group 0, the trajectory follows the geodesic under the identity warp: \(\tau \mapsto \tau\). For Group 1, the trajectory follows the same geodesic under a triangle-bounce warp: \(\tau \mapsto 2\tau\) for \(\tau \leq 1/2\), and \(\tau \mapsto 2(1-\tau)\) for \(\tau > 1/2\). Since the two warps have the same mean (\(1/2\)) and variance (\(1/12\)), the Jensen gap is zero. Thus, the groups are distinguished only by temporal progression along the geodesic. \emph{Rung: trajectory direction.}
    \item[I \((q=60, k=2)\)]: We generate the data using a two-level hierarchical HMM \citep{Fine199841}. The outer state, taking values in \(\{O_1, O_2\}\), evolves slowly over time, while the inner state, taking values in \(\{I_1, I_2, I_3\}\), evolves conditional on the current outer state. The two groups differ in the coupling structure between the outer and inner states. \emph{Rung: multi-scale dynamics.}

    \item[J \((q=50, k=2)\):] The two groups are anchored to meta-analytic brain maps from the Neurosynth database \citep{Yarkoni2011neurosynth}, accessed through \pkg{NiMARE} \citep{Salo2023nimare}. We selected six cognitive terms and divided them into two concept families: an attentional-control family (\emph{attention}, \emph{executive}, \emph{inhibition}) and a memory family (\emph{memory}, \emph{recall}, \emph{encoding}). For each term \(t\), we identified all Neurosynth studies whose abstract-level TF-IDF loading for that term exceeded \(10^{-3}\). We then computed a multilevel kernel density analysis map by placing a 10\,mm spherical kernel at each reported peak coordinate and summing the resulting kernels across studies. Each term-specific density map was parcellated into the 48 Harvard-Oxford cortical regions, producing a regional score vector \(\mathbf{s}(t) \in \mathbb{R}^{48}\). This vector was mean-centred and unit-normalised to obtain \(\mathbf{u}(t)\), and then converted into an SPD anchor matrix using \(\mathbf{X}(t) = \epsilon \mathbf{I} + \mathbf{u}(t) \mathbf{u}(t)^{\top},\) where \(\epsilon\) was chosen so that \(\mathbf{X}(t)\) had condition number 7.5. For each synthetic trajectory, the sequence was divided into three approximately equal temporal segments. Each group cycled through the three Neurosynth-derived anchors from its own concept family, with a random \(\pm 3\)-window jitter applied to each segment boundary. Additional log-Euclidean tangent-space perturbations were added at both the subject and window levels. 

\end{description}

\begin{table}[!ht]
\centering
\caption{ARI scores for the ten simulation scenarios.  The table reports the mean (top row) and standard deviation (bottom row) of the scores over 100 repetitions.}
\label{data_synthetic_ARI}
\begin{tabular}{l|ccccccc}
\toprule
Scenario (\textit{rung}) & TSRVF& GeoAtt & SPDNet & PGA & DC & PDM & MatFAE \\
\midrule
A (\textit{temporal ordering}) &  0.999&  -0.001&  -0.000&  -0.001&  0.990&  -0.000& \textbf{1.000} \\
  &  0.009&  0.011&  0.013&  0.012&  0.101&  0.012&  0.000 \\
B (\textit{static identity}) &  0.010&  0.028&  0.043&  \textbf{0.999}&  0.007&  0.961& \textbf{0.980} \\
  &  0.029&  0.059&  0.070&  0.007&  0.028&  0.119&  0.009 \\
C (\textit{dwell time})&  0.295&  0.437&  0.429&  0.439&  0.434&  0.440& \textbf{0.541} \\
  &  0.069&  0.082&  0.065&  0.069&  0.083&  0.069&  0.032 \\
D (\textit{transition frequency})&  0.279&  -0.002&  0.000&  -0.000&  0.272&  0.001& \textbf{0.286} \\
  &  0.062&  0.011&  0.014&  0.013&  0.107&  0.011&  0.052 \\
E (\textit{smoothness})&  0.064&  0.061&  0.011&  0.063&  0.012&  0.000& \textbf{0.808} \\
  &  0.062&  0.043&  0.027&  0.052&  0.016&  0.000&  0.078 \\
F (\textit{Wishart concentration})&  0.077&  0.999&  0.439&  0.672&  0.128&  0.535& \textbf{1.000} \\
  &  0.045&  0.005&  0.080&  0.072&  0.062&  0.075&  0.000 \\
G (\textit{real HRF})&  -0.001&  0.000&  0.004& \textbf{1.000}&  0.981&  0.170&  0.960 \\
  &  0.012&  0.015&  0.020&  0.000&  0.115&  0.124&  0.018 \\
H (\textit{trajectory direction})& \textbf{1.000}&  0.192&  0.038&  0.012& \textbf{1.000}&  0.001& \textbf{1.000} \\
  &  0.000&  0.092&  0.042&  0.028&  0.000&  0.013&  0.000 \\
I (\textit{multi-scale dynamics}) & \textbf{0.097}&  0.034&  0.035&  0.033&  0.046&  0.034&  0.085 \\
  &  0.072&  0.046&  0.043&  0.043&  0.053&  0.044&  0.064 \\
J (\textit{Neurosynth}) &  0.000&  0.282&  0.791&  \textbf{1.000}&  0.784&  \textbf{1.000}&  \textbf{1.000}\\
  &  0.013&  0.267&  0.121&  0.000&  0.343&  0.000&  0.000\\
\bottomrule
\end{tabular}
\end{table}

Table~\ref{data_synthetic_ARI} shows a pattern broadly consistent with the ARI results. MatFAE achieves the highest mean ARI in seven scenarios (A, C, D, E, F, H and J), with especially large advantages in the more challenging settings C and E. The gain in the smoothness scenario E is particularly striking: MatFAE reaches 0.808, whereas all competing methods remain near zero. It also performs best in F and ties for perfect recovery in H and J.

The baseline methods again show more specialized behavior. TSRVF and DC perform very well in the temporally driven scenarios A and H, while PGA is strongest in the static identity and real-HRF scenarios B, J and G. Scenario I remains the most difficult overall, with all methods achieving low ARI values; unlike the AMI results, TSRVF attains the highest mean score there, although the margin over MatFAE is small. Overall, the ARI results reinforce the conclusion that MatFAE is the most robust method across the benchmark, with its clearest advantage appearing in the harder scenarios.

\subsection{Details on the real datasets and additional experimental results}\label{Supplementary_real}

All six resting-state fMRI cohorts were obtained from publicly available preprocessed releases. We retained the preprocessing provided by each consortium, including motion correction, T1 coregistration, spatial normalization to MNI space, and, where applicable, upstream confound regression. To improve comparability across datasets, we then applied a common post-hoc processing pipeline where the required data were available.

Unless stated otherwise, the post-hoc pipeline included nuisance regression using the full Friston-24 motion model, 5-6 aCompCor components, and volume-level scrubbing for frames with framewise displacement greater than \(0.5\,\mathrm{mm}\). Scrubbing was implemented either as one-hot spike regressors or through sample-mask exclusion, depending on the confound information provided with each release. We then applied temporal band-pass filtering at \(0.01\)-\(0.1\,\mathrm{Hz}\), linear and quadratic detrending, and per-ROI \(z\)-scoring.

When native NIfTI images were available, we parcellated the data using the 100-parcel, 7-network Schaefer atlas in MNI \(2\,\mathrm{mm}\) space. When only pre-extracted ROI time series were provided, we used the atlas distributed with the corresponding release. Time-varying functional connectivity was estimated using Ledoit-Wolf shrinkage covariance matrices computed within \(30\,\mathrm{s}\) sliding windows with a \(4\,\mathrm{s}\) step. This produced one SPD-valued covariance trajectory for each subject.

Dataset-specific deviations from this common pipeline are described below. Here, \(n\) denotes the number of retained subjects, \(q\) the number of sliding windows, \(m\) the final ROI dimension after applying a common non-zero-variance mask, and \(k\) the number of diagnostic classes.

\begin{description}
    \item[CNP \((n=257,\ q=57,\ m=100,\ k=4)\) \citep{Poldrack2016phenomics}]: The Consortium for Neuropsychiatric Phenomics (CNP) dataset was obtained from OpenNeuro under accession number \texttt{ds000030}. The cohort comprises healthy controls \((n=120)\), individuals with schizophrenia \((n=48)\), bipolar disorder \((n=49)\), and ADHD \((n=40)\). Each participant has one resting-state fMRI run with TR \(=2\,\mathrm{s}\) and 152 volumes. We used the fMRIPrep-v0.4.4 preprocessed release \citep{Gorgolewski2017preprocessedCNP}, which provides motion-corrected BOLD images spatially normalized to the MNI152NLin2009cAsym template (the asymmetric 2009c nonlinear MNI152 standard space) at 2\,mm resolution. The release also includes a 24-column confound TSV file containing 6 motion parameters, 6 aCompCor components, 6 tCompCor components, white-matter signal, global signal, framewise displacement, and DVARS. 
  
    We applied a standard preprocessing pipeline consisting of Friston-24 motion regression, 6 aCompCor components, FD \(>0.5\,\mathrm{mm}\) spike regression, and temporal band-pass filtering. The data were then parcellated in MNI space using the 100-parcel Schaefer atlas. Ledoit-Wolf covariance matrices were computed over 30\,s sliding windows, corresponding to 15 TRs, with a step size of 2 TRs. Four subjects were excluded because more than 50\% of their volumes exceeded the FD threshold.

    \item[COBRE \((n=144,\ q=68,\ m=100,\ k=2)\)]: The COBRE (RRID:SCR\_010482) dataset was obtained from the International Neuroimaging Data-sharing Initiative. The cohort includes individuals with schizophrenia ($n = 70$) and healthy controls ($n = 74$), giving a total sample size of 144 participants. Each resting-state scan was acquired with TR \(=2\,\mathrm{s}\) and contains 150 volumes. The data were preprocessed upstream by Bellec and colleagues using the NIAK pipeline (v0.17). This preprocessing produced BOLD images in MNI152NLin2009a space at 6 mm isotropic resolution, together with per-subject confound TSV files. These confound files include 6 motion parameters, framewise displacement, a scrub flag, 6 slow-drift discrete cosine transform basis functions, white-matter and ventricle mean signals, and 5 aCompCor components. 
    
    Importantly, these confound variables are provided with the data but were not regressed out in the released BOLD time series. We therefore applied our common nuisance-regression and filtering pipeline, including Friston-24 motion regressors, 5 aCompCor components, white-matter and ventricle signals, one-hot spike regressors for volumes marked with scrub = 1, and temporal band-pass filtering. The resulting time series were parcellated in MNI space using the 100-parcel Schaefer atlas. Time-varying functional connectivity was then estimated by computing Ledoit-Wolf covariance matrices within 30 s sliding windows, corresponding to 15 TRs, with a step size of 2 TRs. 

    \item[ADHD-200 \((n=138,\ q=57,\ m=100,\ k=2)\) \citep{ADHD2002012,Bellec2017ADHD200}]: The ADHD-200 dataset is a multi-site cohort released by the ADHD-200 Consortium and preprocessed by the Preprocessed Connectomes Project. We used the CPAC-preprocessed version distributed at \texttt{s3://fcp-indi/data/Projects/ADHD200/Outputs/cpac/}, with the configuration \texttt{pc10.linear1.wm0.global0.motion1.quadratic1.gm0.compcor1.csf0}.  The cohort includes typically developing controls \((n=88)\) and individuals with ADHD \((n=50)\). Data were acquired across 10 sites, with site-specific TR values of \(1.5\), \(1.96\), \(2.0\), or \(2.5\,\mathrm{s}\).  In this preprocessing stream, 6 motion parameters, 5 aCompCor components, and linear and quadratic trends were already regressed from the BOLD time series. ROI time series were then extracted using the 100-parcel Schaefer atlas.
    
    Because only the extracted ROI time series are available, rather than the raw BOLD images or confound TSV files, our post-hoc preprocessing was limited to temporal band-pass filtering using the site-specific TR. We then estimated time-varying functional connectivity by computing Ledoit-Wolf covariance matrices within sliding windows. Additional Friston-24 motion regression or volume scrubbing could not be applied retrospectively. Subjects were excluded if their scan duration satisfied \(q \cdot \mathrm{TR} < 240\,\mathrm{s}\), or if more than 5\% of parcels had zero variance. This led to the exclusion of one outlier, \texttt{sub-0015038}, and allowed the full 100-ROI Schaefer atlas to be retained across the cohort.

    \item[ABIDE-I \((n=846,\ q=57,\ m=101,\ k=2)\)]: The Autism Brain Imaging Data Exchange I (ABIDE I) dataset \citep{DiMartino2014abide} is a multi-site resting-state fMRI cohort comprising healthy controls \((n=455)\) and individuals with autism spectrum disorder (ASD; \(n=391\)). Data were acquired across 17 sites, with TR values ranging from \(1.5\) to \(3.0\,\mathrm{s}\). We used the CPAC-preprocessed \texttt{nofilt\_noglobal} release from the Preprocessed Connectomes Project, distributed at \texttt{s3://fcp-indi/data/Projects/ABID\_Initiative/Outputs/cpac/nofilt\_noglobal/}. In this preprocessing stream, 24 Friston motion parameters, 5 aCompCor components, and linear and quadratic trends were regressed from the BOLD data. As indicated by the \texttt{nofilt} label, temporal band-pass filtering was not applied upstream; volume scrubbing was also not performed.
    
    Because Schaefer-atlas time series are not provided in the PCP release for ABIDE I, we used the shipped Harvard-Oxford ROI time series directly. Our post-hoc preprocessing therefore consisted of temporal band-pass filtering using the site-specific TR, followed by estimation of time-varying functional connectivity using Ledoit-Wolf covariance matrices within sliding windows. After applying a common non-zero-variance mask across all subjects, 101 of the 111 Harvard-Oxford parcels were retained.

    \item[TCP \((n=241,\ q=80,\ m=100,\ k=2)\) \citep{Chopra2024tcp}]: The Transdiagnostic Connectome Project (TCP) dataset was obtained from OpenNeuro under accession number \texttt{ds005237}. The cohort includes healthy general-population controls \((n=92)\) and transdiagnostic patients \((n=149)\), aged 18-70 years, recruited at two acquisition sites: Yale University and McLean Hospital. Each participant has four resting-state fMRI runs, acquired with anterior–posterior and posterior–anterior phase-encoding directions (\texttt{task-restAP\_run-01/02} and \texttt{task-restPA\_run-01/02}). Each run was acquired with TR \(=0.8\,\mathrm{s}\), multiband factor 8, and 488 volumes. The OpenNeuro release provides time series after the consortium's HCP-style minimal preprocessing pipeline, including motion and distortion correction, T1 coregistration, spatial normalization to MNI space, CIFTI grayordinate projection, ICA-FIX automated noise classification and regression \citep{SalimiKhorshidi2014icafix}, global signal regression, and \(1/2000\,\mathrm{Hz}\) high-pass detrending. The released time series are parcellated using the 488-region CAB-NP atlas, comprising 360 Glasser cortical parcels \citep{Glasser2016MMP} and 128 subcortical CIFTI grayordinate parcels. 
    
    We retained the first anterior-posterior resting-state run \((\texttt{task-restAP\_run-01})\) for each subject. Because the released data were already denoised and provided as parcellated time series, our post-hoc preprocessing was limited to temporal band-pass filtering at \(0.01\)-\(0.1\,\mathrm{Hz}\), followed by estimation of time-varying functional connectivity using Ledoit-Wolf covariance matrices. Covariance matrices were computed within 30\,s sliding windows, corresponding to 38 TRs, with a step size of 5 TRs. To improve comparability with CNP, COBRE, and ADHD-200, we mapped the 360 Glasser cortical parcels to the 100-parcel Schaefer atlas using a row-stochastic voxel-overlap mapping computed in MNI152NLin2009cAsym \(2\,\mathrm{mm}\) space; the 128 subcortical parcels were not used. Subjects were excluded if more than 50\% of volumes exceeded the framewise-displacement threshold of \(0.5\,\mathrm{mm}\), based on the released \texttt{motion\_FD/} files.
    
    \item[CAT-D \((n=118,\ q=49,\ m=100,\ k=2)\) \citep{Sadeghi2022catd,Camp2024catd}]: The NIMH Characterization and Treatment of Adolescent Depression (CAT-D) dataset was obtained from OpenNeuro under accession number \texttt{ds004627}. The cohort includes healthy volunteers \((n=50)\) and adolescents with major depressive disorder \((n=68)\), all scanned at a single site on a GE Discovery MR750 3T scanner. Each participant has a multi-echo resting-state fMRI run with TR \(=2.5\,\mathrm{s}\) and three echoes, with TE approximately 14.4, 28.8, and 43.2\,\(\mathrm{ms}\). We retained only the second echo for parity with the other real datasets and restricted the analysis to the baseline session (\texttt{ses-v1}), so that each subject contributed a single resting-state trajectory. 
    
    Unlike the other datasets, CAT-D does not have a publicly available upstream-preprocessed release. We therefore preprocessed the data using fMRIPrep v23.2.0 \citep{Esteban2019fmriprep}, which produced motion-corrected BOLD images spatially normalized to the MNI152NLin2009cAsym template at \(2\,\mathrm{mm}\) resolution, together with per-subject confound TSV files. We then applied our post-hoc preprocessing pipeline, including Friston-24 motion regression, 6 anatomical aCompCor components, white-matter and CSF mean signals, volume censoring based on FD \(>0.5\,\mathrm{mm}\) and standardized DVARS \(>1.5\), and temporal band-pass filtering at \(0.01\)-\(0.1\,\mathrm{Hz}\). The resulting time series were parcellated in MNI space using the 100-parcel Schaefer atlas. Time-varying functional connectivity was estimated by computing Ledoit-Wolf covariance matrices within 30 s sliding windows, corresponding to 12 TRs, with a step size of 2 TRs.
    
\end{description}

\begin{table}[!ht]
\centering
\caption{The ARI scores for the six real datasets.}
\label{data_real_ARI}
\begin{tabular}{l|ccccccc}
\toprule
Dataset & TSRVF & GeoAtt & SPDNet & PGA & DC & PDM & MatFAE\\
\midrule
CNP & 0.057 & 0.017 & 0.015 & 0.039 & 0.018 & 0.001 & \textbf{0.128}\\
COBRE & 0.087 & 0.071 & 0.071 & 0.056 & 0.018 & 0.056 & \textbf{0.169}\\
ADHD-200 & 0.037 & 0.005 & 0.006 & 0.040 & 0.075 & 0.078 & \textbf{0.109}\\
ABIDE-I & 0.014 & 0.001 & 0.001 & 0.000 & 0.005 & 0.001 & \textbf{0.038}\\
TCP & 0.036 & 0.002 & 0.013 & 0.028 & 0.005 & 0.025 & \textbf{0.053}\\
CAT-D & 0.026 & 0.031 & 0.007 & 0.056 & 0.000 & 0.039 & \textbf{0.085}\\
\bottomrule
\end{tabular}
\end{table}

\pkg{MatFAE} achieves the highest ARI on all six cohorts, indicating a consistent advantage over the competing methods. The absolute ARI values, however, remain modest, which reflects the difficulty of unsupervised diagnostic discovery from resting-state fMRI.  Several patterns are worth noting. First, the deep-learning baselines, especially GeoAtt and SPDNet, perform close to chance on several cohorts, including CNP, ABIDE-I, TCP, and CAT-D. This suggests that parameter-heavy SPD encoders may be vulnerable to weak disease signal, limited sample size, site effects, and preprocessing variability. Second, the strongest competing baselines differ across cohorts: TSRVF is the strongest baseline on CNP and COBRE, PDM and DC are strongest on ADHD-200, and PGA is strongest on CAT-D. Nevertheless, \pkg{MatFAE} remains consistently ahead of these cohort-specific competitors, with the largest margins observed on COBRE and CNP. This suggests that the combination of geometric layers and functional feature extraction provides a more stable representation than relying only on temporal alignment, tangent-space summaries, or static geometric structure. Read together with the simulation study, the real-data results support the same conclusion. Individual baselines are competitive in specific regimes, such as TSRVF and DC for temporal ordering, PGA for static identity or HRF-blurred connectivity, and GeoAtt for Wishart concentration. By contrast, \pkg{MatFAE} performs strongly across multiple regimes and is particularly effective on dwell time, transition frequency, and smoothness, which are relevant to resting-state dynamic connectivity. The real fMRI results therefore show that \pkg{MatFAE} is more discriminative than the tested alternatives, while the low absolute ARI values also highlight that unsupervised recovery of DSM diagnostic labels from resting-state dynamic connectivity remains a challenging open problem.

\subsection{Statistical significance tests and ablation study}\label{Statistical significance tests and ablation study}
We test the statistical significance for the reported improvements on both simulated and real data. For each synthetic rung, we test whether MatFAE (clustering) and MatFAE-C (classification) perform significantly above chance. Specifically, we apply a one-sided one-sample $t$-test to the scores across random seeds, using a chance level of 0 for adjusted ARI and 0.5 for balanced accuracy on the binary-class rungs (0.33 for the three-class rung). We report both the $p$-value and effect size (Cohen's $d$). The clustering analysis uses 100 seeds per rung, while the classification analysis uses 20 seeds (5$\times$3 stratified CV). For the six real datasets, we apply the same procedure: For clustering, we set the number of clusters to the known number and test whether the mean ARI across 20 training seeds exceeds the chance level of 0. For classification, we test whether the mean balanced accuracy across the 15 outer test folds of the $5\times3$ stratified cross-validation exceeds the chance level of $1/K$, where $K$ is the number of classes. This corresponds to 0.25 for the four-class CNP dataset and 0.5 for the remaining binary datasets. The test results are given in Table \ref{statistical_test}.
\begin{table}[!ht]
\centering
\caption{Statistical significance of performance on the synthetic rungs and real datasets.}
\label{statistical_test}
\begin{tabular}{l|cccc}
\toprule
Scenario (\textit{rung}) & $p$ (MatFAE) & $d$ & $p$ (MatFAE-C) & $d$\\
\midrule
A (\textit{temporal ordering}) &$<$0.001  &ceiling  &$<$0.001  & 63.8\\
B (\textit{static identity}) &$<$0.001  &62.8  &$<$0.001  & 8.3\\
C (\textit{dwell time}) &$<$0.001  &16.9  &$<$0.001  & 4.5\\
D (\textit{transition frequency}) &$<$0.001  &	5.1  &$<$0.001  & 4.9\\
E (\textit{smoothness}) &$<$0.001  &10.4  &$<$0.001  & 6.6\\
F (\textit{Wishart concentration}) &$<$0.001  &ceiling  &$<$0.001  & 20.6\\
G (\textit{real HRF}) &$<$0.001  &10.8  &$<$0.001  & 7.3\\
H (\textit{trajectory direction}) &$<$0.001  &ceiling  &$<$0.001  & 50.9\\
I (\textit{multi-scale dynamics}) &0.003  &1.1  &$<$0.001  & 19.3\\
J (\textit{Neurosynth}) &$<$0.001  &ceiling  &$<$0.001  & 14.9\\
\hline
CNP & $<$0.001 & 12.7 & $<$0.001 & 4.2 \\
COBRE & $<$0.001 & 11.9 & $<$0.001 & 5.1 \\
ADHD-200 & $<$0.001 & 5.8 & 0.008 & 1.8 \\
ABIDE-I & $<$0.001 & 17.3 & 0.005 & 2.0 \\
TCP & $<$0.001 & 3.3 & 0.011 & 1.6 \\
CAT-D & $<$0.001 & 12.0 & $<$0.001 & 2.5\\
\bottomrule
\end{tabular}
\end{table}

We conduct a component-wise ablation, removing one module at a time. Each model variant is evaluated on three representative simulation rungs (E, G, and J) and the mean ARI over 100 independent repetitions is reported.
\begin{table}[!ht]
\centering
\caption{Component-wise ablation results for MatFAE on three simulation rungs. Each row indicates a component removed or simplified relative to the full model. Values are mean ARIs over 100 independent repetitions.}
\label{ablation_study}
\begin{tabular}{l|ccc}
\toprule
Module removed & E (\textit{smoothness}) & G (\textit{real HRF}) & J (\textit{Neurosynth})\\
\midrule
 -- multi-head ($H\rightarrow1$) & 0.031 & 0.007 & 0.136\\
 -- functional layers & 0.021 & 0.818 & 0.680\\
 -- module $\mathcal{F}$ (direct half-vectorization) & 0.016 & 0.030 & 0.120\\
 -- module $\mathcal{F}$ (Riemannian logarithm vectorization) &0.025 &-0.001 &0.770\\
 -- second BiMap layer  & 0.674 & 0.866 & 0.939\\
 -- geodesic shrinkage ($\alpha\rightarrow1$) & 0.624 & 0.804 & 0.960\\
 \hline
 Full MatFAE  & 0.808  & 0.960  & 1.000\\
\bottomrule
\end{tabular}
\end{table}
Across the three rungs, the most pronounced and recurring performance degradation occurs when the multi-head BiMap is reduced to a single head, the matrix-to-vector module $\mathcal{F}$ is replaced, or the functional layers are removed. These results indicate that all three components are critical to the performance of MatFAE.

We conduct a one-factor-at-a-time sensitivity analysis on the same three simulation rungs: E, G, and J. Each hyperparameter is varied while all others are held fixed. Table~\ref{sensitivity_study} reports the mean ARI over 100 independent repetitions.
\begin{table}[!ht]
\centering
\caption{One-factor-at-a-time hyperparameter sensitivity analysis on three simulation rungs. The number of heads ($H$), latent dimension ($s$), basis size ($K$), intermediate SPD dimension ($m_1$), and geodesic shrinkage parameter ($\alpha$) are varied individually while the remaining hyperparameters are held fixed. Values are mean ARIs over 100 independent repetitions.}
\label{sensitivity_study}
\begin{tabular}{l|ccc}
\toprule
Hyper-parameter & E (\textit{smoothness}) & G (\textit{real HRF}) & J (\textit{Neurosynth})\\
\midrule
$H$ \{1, 2, 4, 8, 16\} &\{.03, .11, .14, \textbf{.81}, .79\} &\{.01, .33, .51, .96, \textbf{.98}\} &\{.14, .28, .77, \textbf{1.00}, 1.00\} \\
$s$ \{2, 4, 8, 16, 32\} &\{.64, \textbf{.79}, .78, .76, .78\} &\{.81, .89, \textbf{.95}, .93, .90\} & \{.96, .99, \textbf{1.00}, 1.00, .96\}\\
$K$ \{6, 10, 15, 20, 25\} &\{.29, .40, \textbf{.82}, .79, .79\} &\{.64, .87, .93, .95, \textbf{.96}\} &\{.84, .94, \textbf{1.00}, 1.00, 1.00\} \\
$m_1$ \{16, 24, 32, 48\} &\{.68, .72, \textbf{.81}, .80\} &\{.85, .88, \textbf{.96}, .96\} &\{.96, \textbf{1.00}, 1.00, 1.00\} \\
$\alpha$  \{.1, .25, .5, .75, .9\} &\{.54, .60, \textbf{.80}, .79, .62\} &\{.78, .84, \textbf{.94}, .86, .84\} &\{.94, .96, \textbf{1.00}, .99, 1.00\} \\
\bottomrule
\end{tabular}
\end{table}

The results reveal the following trends:
\begin{description}
    \item[Number of heads H:] Performance generally improves as additional heads capture distinct localized temporal patterns, before plateauing once the relevant temporal regimes are adequately represented.
    \item[Latent dimension \(s\):] Performance is relatively stable over a moderate range of latent dimensions. Small values can restrict representational capacity, whereas larger values provide no consistent benefit and may introduce redundant features.
    \item[Basis size \(K\):] Performance improves until the basis is sufficiently expressive, after which additional basis functions yield little or no benefit. Thus, \(K\) acts primarily as a minimum capacity requirement rather than following a larger-is-better relationship.
    \item[Intermediate SPD dimension \(m_1\):] Moderate values generally provide sufficient representational capacity. Increasing \(m_1\) beyond this range produces only marginal improvements, indicating limited sensitivity once an adequate dimension is reached.
    \item[Geodesic shrinkage \(\alpha\):] Intermediate values generally perform best. Because the pooling operation already introduces nonlinearity, geodesic shrinkage primarily acts as a regularizer; excessive or insufficient shrinkage can reduce performance.
\end{description}

\subsection{Interpretability of \pkg{MatFAE}}\label{COBRE_interpretability}

Compared with existing complex deep learning models used in neuroscience, \pkg{MatFAE} combines a lightweight architecture with intrinsic interpretability. In particular, the temporal profile of the functional weights \(\pmb{W}^{(1)}(t)\) in the encoder functional layer indicates which slices of the SPD trajectory contribute most strongly to the latent representations. We illustrate this interpretability using the COBRE dataset by comparing statistics of \(\pmb{y}(t)\) with the temporal profile of \(\pmb{W}^{(1)}(t)\) learned by \pkg{MatFAE} under the configuration reported in Appendix~\ref{Supplementary_real}.

\paragraph{Profile and peak window} 
Let \(g(t)=\|\pmb{W}^{(1)}(t)\|_F\) denote the Frobenius norm of the functional weight at time index \(t\), and let \(e(t)=\sqrt{\sum_{k=1}^{K}B_k(t)^2}\) denote the pointwise \(\ell_2\) envelope of the B-spline basis. Because clamped open-uniform B-splines satisfy \(e(0)=e(T-1)=1\), while the interior values are below unity, the raw norm \(g(t)\) contains a deterministic boundary effect induced by the basis itself. To isolate the data-driven temporal preference of the encoder, we therefore use the basis-corrected profile \(\tilde{g}(t)=g(t)/e(t)\). The peak window \([t_1,t_2]\) is selected as the contiguous interior interval of width \(\lfloor 0.20\,T\rfloor\) centred at \(\arg\max_t \tilde{g}(t)\), with two end-windows excluded on each side.

\begin{figure}[!htp]
    \centering
    \includegraphics[width=\linewidth]{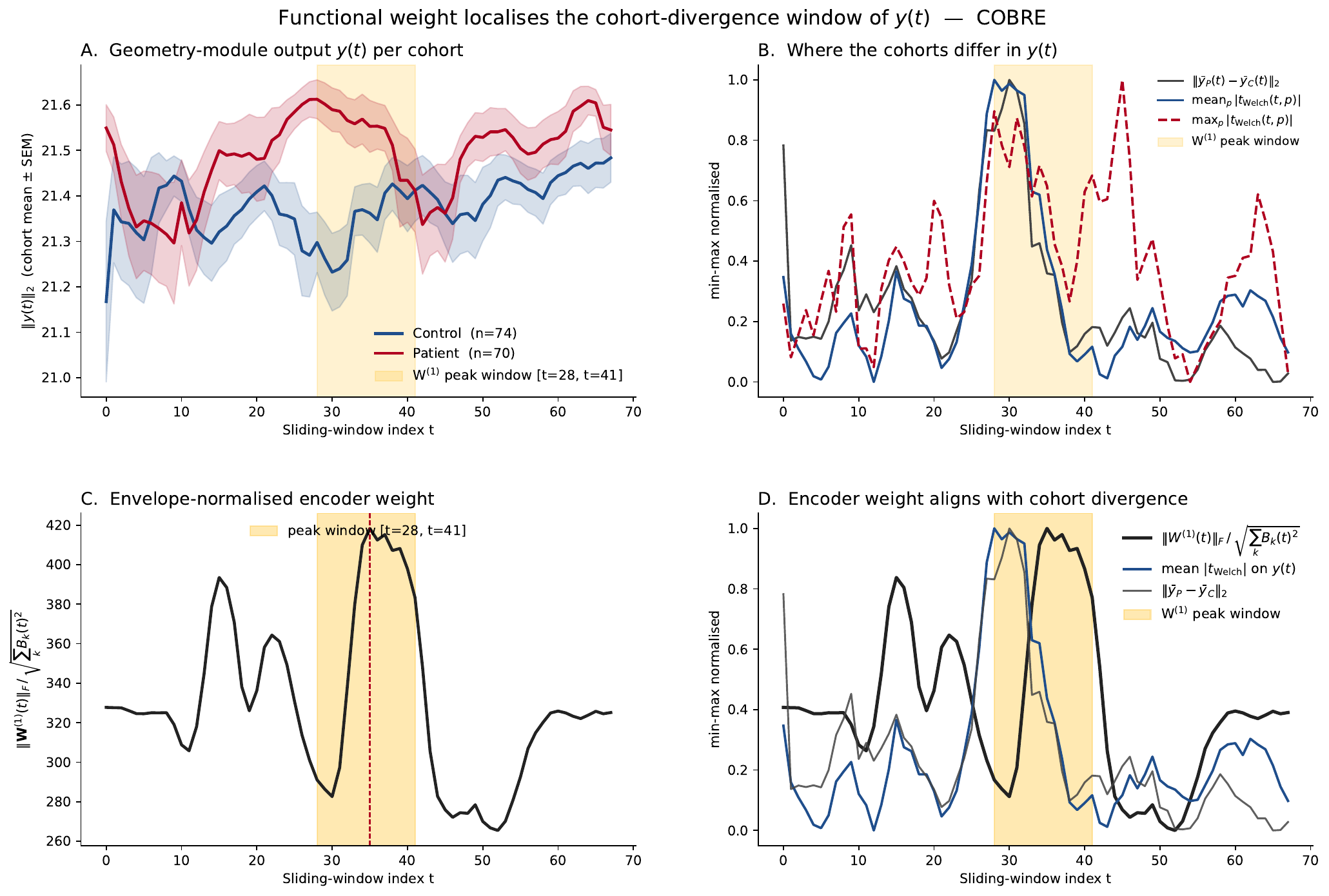}
    \caption{\textbf{The encoder functional weight localises the cohort-divergence window of \(\pmb{y}(t)\).} \textbf{(A)} Cohort-mean \(\|\pmb{y}(t)\|_2\) for healthy controls (\(n=74\)) and schizophrenia patients (\(n=70\)), with within-cohort SEM bands. \textbf{(B)} Two pointwise cohort-divergence summaries on \(\pmb{y}(t)\), namely \(\|\bar{\pmb{y}}_P(t)-\bar{\pmb{y}}_C(t)\|_2\) and \(\mathrm{mean}_p\,|t_{\mathrm{Welch}}(t,p)|\), both min-max normalised. \textbf{(C)} Basis-corrected encoder weight profile \(\tilde{g}(t)=\|\pmb{W}^{(1)}(t)\|_F/\sqrt{\sum_k B_k(t)^2}\), with the selected peak window shaded. \textbf{(D)} Min-max overlay of (B) and (C). The maximum of \(\tilde{g}\) at \(t=35\) lies \(5\)-\(7\) windows away from the cohort-divergence maxima at \(t=28\) for \(\mathrm{mean}_p\,|t_{\mathrm{Welch}}(t,p)|\) and \(t=30\) for \(\|\bar{\pmb{y}}_P(t)-\bar{\pmb{y}}_C(t)\|_2\). All three peaks fall within the encoder peak window \([t_1,t_2]=[28,41]\).}
    \label{fig:cobre_interp_story}
\end{figure}

\paragraph{Cohort divergence in \(\pmb{y}(t)\)} 
For each subject \(i\), we compute \(\pmb{y}_i(t)\in\mathbb{R}^{p}\) and define two pointwise cohort-divergence summaries: the \(\ell_2\) norm of the cohort-mean difference, \(\|\bar{\pmb{y}}_P(t)-\bar{\pmb{y}}_C(t)\|_2\), and the mean absolute Welch two-sample \(t\)-statistic across the \(p\) coordinates, \(\mathrm{mean}_p\,|t_{\mathrm{Welch}}(t,p)|\). These two quantities identify the time indices at which the patient and control cohorts are most separable in \(\pmb{y}(t)\). \autoref{fig:cobre_interp_story} supports the design claim that \(\pmb{W}^{(1)}(t)\) concentrates on the same temporal interval in which the geometry-module output \(\pmb{y}(t)\) shows the strongest cohort separation. Using the basis-corrected profile is important because clamped B-spline endpoints inflate \(g(t)\) at \(t\in\{0,T-1\}\) by a factor of approximately \(1/0.68\) for purely geometric reasons; these endpoint spikes therefore do not reflect learned temporal content.

\begin{figure}[!htp]
    \centering
    \includegraphics[width=0.85\linewidth]{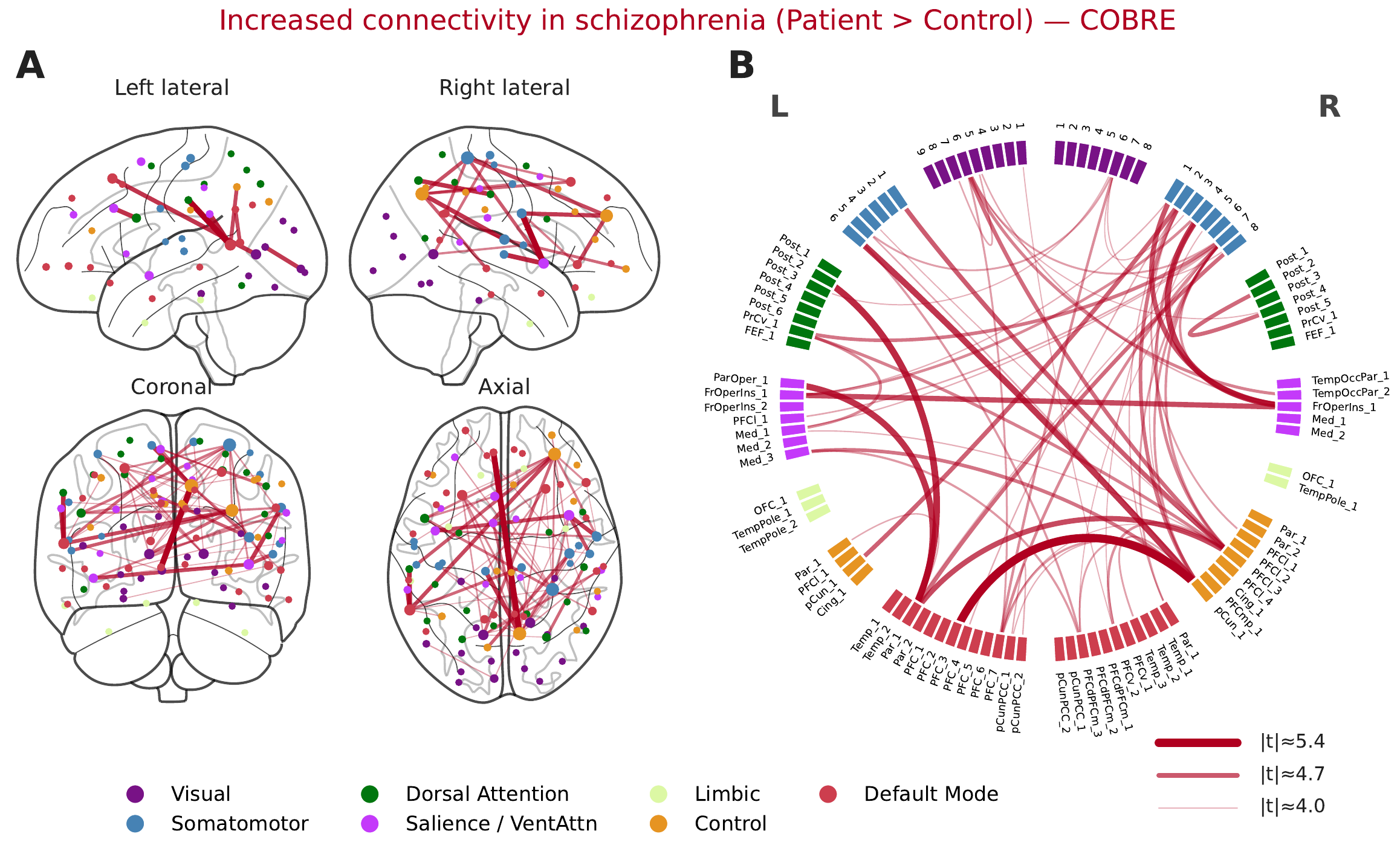}
    \caption{\textbf{Increased connectivity in schizophrenia on COBRE.} Top-60 edges by Welch \(|t|\) where the schizophrenia cohort (\(n=70\)) has higher Fisher-\(z\) correlation than the healthy-control cohort (\(n=74\)). \textbf{(A)} Multi-view glass brain, including left lateral, right lateral, coronal, and axial views, with parcel centroids in MNI 2 mm space. Nodes are coloured by Yeo-7 network and sized by the total Welch \(|t|\) at each node. \textbf{(B)} Chord plot with all 100 Schaefer-100 parcels on the rim, grouped by Yeo-7 network. The right hemisphere fills the right half of the circle clockwise from 12 o'clock, while the left hemisphere fills the left half counter-clockwise. Both Visual arcs are located at 12 o'clock and both Default-Mode arcs at 6 o'clock, with a cyclic gap at the top and an internal hemisphere gap at the bottom. Edge thickness and opacity scale linearly across the top-60 Welch \(|t|\) range, with the weakest edges drawn first so that the strongest edges appear on top.}
    \label{fig:cobre_increased}
\end{figure}

\paragraph{Connectivity contrast restricted to the peak window} 
We next provide a static functional-connectivity visualisation of the contrast that \pkg{MatFAE} is expected to capture. \autoref{fig:cobre_increased} shows the edges for which the schizophrenia cohort has higher mean Fisher-\(z\) correlation than the healthy-control cohort. This visualisation identifies the spatial structure of the static signal and helps explain why COBRE is the cohort on which \pkg{MatFAE} achieves its largest gain. To construct this static visualisation, each subject's SPD trajectory \(\{\tX_i(t)\}_{t=1}^{q}\), with \(q=68\) sliding-window covariance matrices and \(m=100\) Schaefer-100 7-network parcels, is collapsed to a single SPD matrix by the Log-Euclidean Fr\'{e}chet mean,
\begin{equation}\label{cobre_lefrechet}
\bar{\mathbf{X}}_i \;=\; \mathrm{Exp}^{\mathrm{LE}}_{\mathbf{I}}\!\Big(\frac{1}{q}\sum_{t=1}^{q} \mathrm{Log}^{\mathrm{LE}}_{\mathbf{I}} (\tX_i(t))\Big)
\;=\; \exp\!\Big(\frac{1}{q}\sum_{t=1}^{q} \log\, (\tX_i(t))\Big).
\end{equation}
The static covariance matrix \(\bar{\mathbf{X}}_i\) is then converted to a correlation matrix \(\mathbf{R}_i\) and Fisher-\(z\) transformed entry-wise. For each off-diagonal edge \((j,k)\), a Welch two-sample \(t\)-statistic is computed between the patient (\(n=70\)) and control (\(n=74\)) cohorts using \(\{ \mathrm{atanh}([\mathbf{R}_i]_{j,k}) \}_i\). The top \(60\) edges with the largest Welch \(|t|\) in each direction are retained. Edges are displayed in panel~A on a multi-view glass brain, using parcel centroids in MNI 2\,mm space, and in panel~B on a 100-node chord plot grouped by Yeo-7 network. Within each hemisphere, networks are ordered as \(\text{Vis} \to \text{SomMot} \to \text{DorsAttn} \to \text{SalVentAttn} \to \text{Limbic} \to \text{Cont} \to \text{Default}\), so that both \emph{Vis} arcs are placed at 12 o'clock and both \emph{Default} arcs at 6 o'clock. The right hemisphere fills the right half of the circle clockwise from 12 o'clock, while the left hemisphere fills the left half counter-clockwise, mirroring the standard radiological axial view. Edge thickness and opacity scale linearly across the top-60 Welch \(|t|\) range, with weaker edges drawn first so that the strongest edges appear on top. %A single solid colour is used for each direction: red for increased connectivity and blue for decreased connectivity.

\begin{figure}[!htp]
    \centering
    \includegraphics[width=0.85\linewidth]{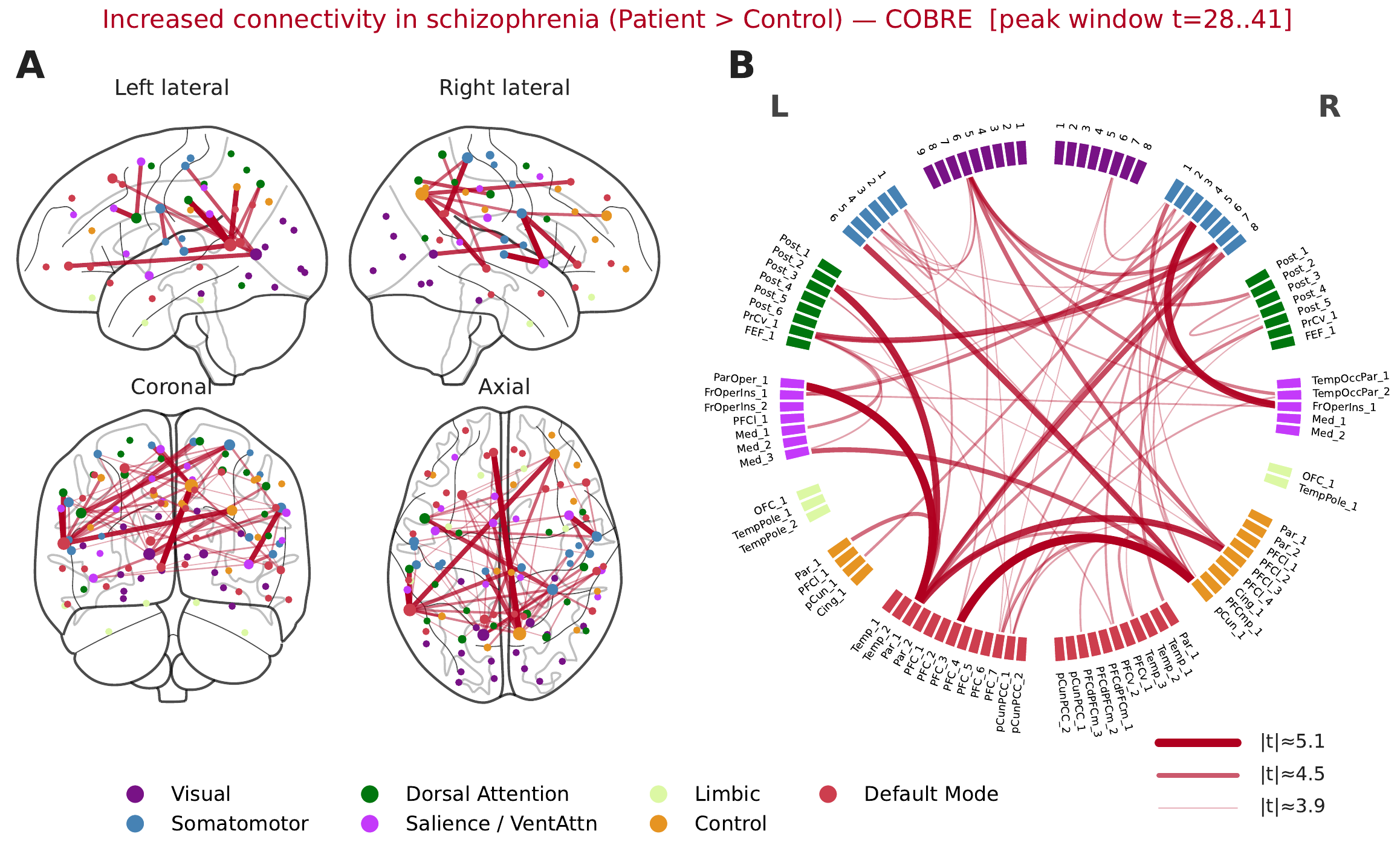}
    \caption{\textbf{Increased connectivity restricted to the encoder peak window.} Top-60 edges by Welch \(|t|\) where the schizophrenia cohort has higher Fisher-\(z\) correlation than the healthy-control cohort, computed from per-subject LE-Fr\'{e}chet means restricted to \(t\in[28,41]\), the window highlighted by \(\pmb{W}^{(1)}(t)\) in \autoref{fig:cobre_interp_story}. The same Default-Mode / Salience-Ventral-Attention / Frontoparietal-Control hyperconnectivity pattern is recovered, with retained $|t|$ values ranging from 3.9 to 5.1. This suggests that the encoder-highlighted window overlaps with the period in which the cohort-discriminative static signal is most concentrated.}
    \label{fig:cobre_interp_increased}
\end{figure}

To test whether the encoder-identified peak window contains the strongest static functional-connectivity contrast, we recompute the LE-Fréchet mean in Eq.~\eqref{cobre_lefrechet} using only \(t\in[t_1,t_2]=[28,41]\). We then repeat the Welch/Fisher-\(z\) edge-level test and retain the top 60 edges by \(|t|\) for which the patient cohort exhibits higher correlation than the control cohort. \autoref{fig:cobre_interp_increased} shows that the canonical schizophrenia hyperconnectivity pattern is preserved when the average is restricted to the encoder-flagged window. The contrast remains dominated by connections among the Default-Mode, Salience/Ventral-Attention, and Frontoparietal-Control networks, with retained \(|t|\in[3.9,5.1]\), compared with \([4.0,5.4]\) under all-window pooling. The modest reduction in the upper \(|t|\) bound is consistent with the peak-window analysis using only \(14/68\approx 21\%\) of the temporal data, while still preserving the main spatial structure of the contrast.

Taken together, Figures~\ref{fig:cobre_interp_story}-\ref{fig:cobre_interp_increased} support the interpretability claim. The functional weight profile \(\pmb{W}^{(1)}(t)\) acts as an internally derived temporal saliency map, and its peak window aligns with the interval in which the underlying clinical contrast is most pronounced.

\section{Limitations}

Individual-level diagnosis and patient/control classification from resting-state fMRI remain challenging because disease-related signals are often weak, distributed, and strongly confounded by scanner/site differences, preprocessing choices, parcellation schemes, and the unknown mental state of participants during scanning. Although we applied a common post-hoc processing pipeline whenever possible, the real datasets were not fully homogeneous in preprocessing or parcellation. Three cohorts used the Schaefer-100 atlas, whereas ABIDE-I used a 101-region Harvard-Oxford parcellation. Therefore, performance differences across cohorts may partly reflect differences in ROI definition, rather than only differences in disease phenotype or model behaviour.

Several additional sources of heterogeneity may affect the estimated dynamic functional connectivity trajectories. ADHD-200 and ABIDE-I are multi-site cohorts with TR values ranging from \(1.5\) to \(3.0\,\mathrm{s}\). We used site-specific band-pass filtering, but did not explicitly harmonise scanner, acquisition, or site effects; residual site-related variability may therefore remain. Upstream preprocessing also differed across consortia, including fMRIPrep for CNP, NIAK for COBRE, and CPAC for ADHD-200 and ABIDE-I. In particular, for ADHD-200 and ABIDE-I, only pre-extracted ROI time series were available, and the raw BOLD images and confound files were not redistributed. Consequently, retrospective Friston-24 motion regression and frame-wise scrubbing could not be applied, making nuisance-control procedures less directly comparable across datasets.

Finally, the two BiMap layers in the matrix-to-vector module contain $H(m_1m+m_2m_1)+1$ trainable entries, independent of the time-series length. The geometry-aware decoder introduces two additional matrices, $\mathbf{M}_1$ and $\mathbf{M}_2$, containing $p_2p_1+mp_2$ trainable entries. All functional weights in FAEclust are represented using basis expansions, so only their expansion coefficients are trainable. The main parameter bottleneck is $\mathcal{W}^{(2)}(t)$ in the decoder. With $K$ basis functions, the matrix contains $O(Kp_2^4)$ trainable coefficients. This limits scalability to very high-dimensional parcellations, such as Schaefer-400 or Schaefer-1000, and motivates future work on more efficient geometric layers, low-rank approximations, or sparse SPD representations.

%%%%%%%%%%%%%%%%%%%%%%%%%%%%%%%%%%%%%%%%%%%%%%%%%%%%%%%%%%%%

\end{document}